\documentclass[reqno]{amsart}

\numberwithin{equation}{section}
\numberwithin{figure}{section}
\numberwithin{table}{section}

\usepackage{caption}
\usepackage[table]{xcolor}
\usepackage{boldline} 
\usepackage{arydshln}

\usepackage{multirow}

\usepackage{colortbl}

\usepackage{amssymb}
\usepackage{amscd}
\usepackage{mathtools}

\usepackage{lipsum}
\usepackage{color}

\theoremstyle{definition}

\newtheorem{theorem}{Theorem}[section]
\newtheorem{definition}[theorem]{Definition}
\newtheorem{remarks}[theorem]{Remark}

\definecolor{mylightgray}{rgb}{0.9, 0.9, 0.9}

\newcommand{\mymu}{\tau}

\newcommand{\myK}{I_1\cdot \dots \cdot I_N }

\newcommand{\staralgebra}{\raisebox{-0.1em}{*}-algebra}
\newcommand\myconj[1]{ {#1}^{\raisebox{0.05em}{$*$} }_T }

\newcommand\myperp[1]{ {#1}^{\raisebox{0.05em}{\scalebox{0.7}{$\perp$}} }_T }

\newcommand\myQ[1]{Q_{T, \,#1}}

\begin{document}

\title[General Data Analytics: A 
Semisimple Paradigm over T-Algebra]
{
General Data Analytics \\
with
Applications to  
Visual Information Analysis: \\ 
A Provable
Backward-Compatible  
Semisimple
Paradigm over T-Algebra 
} 

\author{Liang Liao
and Stephen John Maybank
}

\begin{abstract}  
We consider a novel backward-compatible paradigm of general data analytics over a recently-reported semisimple algebra called t-algebra. We study the generalized matrix framework over the t-algebra by representing the elements of t-algebra by fixed-sized multi-way arrays of complex numbers and the algebraic modules over the t-algebra by a finite number of direct-product factors. Over the t-algebra, many algorithms are generalized in a straightforward manner using this new semisimple paradigm. To demonstrate the new paradigm's performance and its backward-compatibility, we generalize some canonical algorithms for visual pattern analysis. Experiments on public datasets show that the generalized algorithms compare favorably with their canonical counterparts.
\end{abstract}

\maketitle

\tableofcontents

\listoffigures

\section{Introduction}
\label{section:Introduction}

\subsection{Motivation}

In the big-data deluge era, the canonical matrix and tensor paradigm over an algebraically closed field plays an essential role in many areas, including but not limited to machine learning, computer vision, pattern analysis, and statistic inference. Under the canonical matrix and tensor paradigm, observed data are given in the form of high-order arrays of canonical scalars (i.e., real or complex numbers).
For example, an RGB image is a real number array of order three, two orders for the image's spatial measures, and a third for the image's spectral measures. An RGB image is also said to have three modes or three-way. A color video sequence of images is of order four, with three orders for spatial-spectral measures and the fourth-order chronological tempo.

Therefore, it is a natural question of whether there exists an extension of the field $\mathbb{C}$ over which a generalized matrix and tensor paradigm can be established and backward-compatible to the canonical paradigm over a field. Fortunately, the answer is yes, but one had to sacrifice at least one of the axioms of a field to obtain something extended.

\subsection{Background and Related Work}

The most well-known generalization of the field of complex numbers is probably the ring 
$M_{n}(\mathbb{C})$ of all $n\times n$ (where $n$ is a positive  integer) matrices over complex numbers under  the usual matrix addition and multiplication.

The field of complex numbers specializes the ring $M_{n}(\mathbb{C}) $ for $n = 1$.  However, 
when $n \geqslant 2$, the matrix ring $M_{n}(\mathbb{C})$ is not a field.
Two axioms of a field are sacrificed, 
(\romannumeral1) not all non-zero matrices are multiplicatively invertible, and (\romannumeral2) the multiplication is non-commutative.

Besides the matrix ring, some hypercomplex number systems also generalize complex numbers. Among these hypercomplex number systems, well-known is  Hamilton's $\mathbb{H}$
of quaternions, which, up to isomorphism, is a real division subring and subalgebra of $M_2(\mathbb{C})$ \cite{hamilton1848xi,history,hilgert2011structure}. However, the multiplication of quaternions is not commutative.

Most hypercomplex number systems, including Hamilton's quaternions, are all subalgebras of Clifford algebra, and the fruits of generating complex numbers to obtain something extended. However, Clifford algebra's hypercomplex number systems are not suitable for general data analytics partially because they are either non-commutative or incompatible with many canonical notions such as euclidean norms. These hypercomplex number systems so far only find narrow niches in geometry and geometry-related branches of physics and computer sciences \cite{hestenes2003reforming,ablamowicz2004lectures}.

To have a well-defined extension of the field $\mathbb{R}$ other than $\mathbb{C}$, Kilmer et al. 
proposed a tensorial model called ``t-product'' for characterizing the multi-way structures of 
higher-order data. In the ``t-product'' model, a circulant matrix representation is chosen for its 
formulation 
\cite{Kilmer2013Third-TProduct003-0005,Kilmer2011Factorization-TProduct001-0006,liaoliang-00016,Zhang2016A-00015}.  

In the ``t-product'' model, the generalized scalars are fixed-sized first-order arrays of real numbers. Equipped with a circular-convolution multiplication, a scalar multiplication, and an entry-wise addition, these generalized scalars form a finite-dimensional commutative unital real algebra $R$.

With the circulant matrix representation over the generalized scalars, many authors have studied and extended the ``t-product'' model. 
Gleich et al. \cite{Gleich2014The-0004} investigate the generalized eigenvalues and eigenvectors of matrices over the algebra $R$ and show how the standard power method for finding an eigenvector and the standard Arnoldi method for constructing an orthogonal basis for a Krylov subspace can both be generalized over $R$. Braman et al. 
\cite{Braman2010Third-TProduct002-0001} show that the vectors over $R$ form a free module.

Kilmer and Martin show that many properties and structures 
of canonical matrices and vectors
can be generalized. Their examples include transposition, orthogonality, and the singular value decomposition (SVD). The tensor SVD is used to compress tensors. A tensor-based method for image de-blurring is also described. Kilmer et al. \cite{Kilmer2013Third-TProduct003-0005} generalize the inner product of two vectors, suggesting a notion of the angle between two vectors with elements in $R$, and define a notion of orthogonality for two vectors. A generalization of the Gram-Schmidt method for generating an orthonormal set of vectors is also studied \cite{Kilmer2013Third-TProduct003-0005}.

Zhang et al. \cite{liaoliang-00016} use the tensor SVD to efficiently store video sequences and fill in missing entries in video sequences.
Zhang et al. \cite{Zhang2016A-00015} use a randomized version of the tensor SVD to produce low-rank approximations to matrices. Ren and Liao et al. \cite{Liao2017Hyperspectral} define a tensor version of principal component analysis and extract features from hyperspectral images. The features are classified using standard methods such as support vector machines and nearest neighbors. 
Liao et al. \cite{Liao2017Supervised-0009} generalize a sparse representation classifier to tensor data and apply the generalized classifier to image data such as numerals and faces. 
Chen et al. \cite{Chen2015Change-0002} use a four-dimensional HOSVD 
(Higher-Order Singular Value Decomposition), one generalization of the matrix singular value decomposition
over caonical scalars, to detect changes in a time sequence of hyperspectral images. The K-means clustering algorithm is used to classify the pixel values as changed or unchanged. 
Fan et al. \cite{Fan2018Spatial-0003} model a hyperspectral image as the sum of an ideal image, a sparse noise term, and a Gaussian noise term. A product of two low-rank tensors models the ideal image. The low-rank tensors are estimated by minimizing a penalty function obtained by adding the squared errors in a fit of the hyperspectral image to penalty terms for the sparse noise and the sizes of the two low-rank tensors. 
Lu et al. \cite{lu2019tensor,lu2016tensor} approximate a third-order tensor using the sum of a low-rank tensor and a sparse tensor. Under suitable conditions, the low-rank tensor and the sparse tensor are recovered exactly.

However, the formulation in circulant matrices is not straightforwardly compatible with the canonical formulation in standard matrices.  The elements of real algebra R so far remain as first-order arrays of real numbers. 
To represent and extend the existing theories via a straightforward compatible approach, Liao and Maybank et al. proposed a framework called ``t-matrix'' \cite{liao2020generalized, Liao2017Hyperspectral}
via modules over an algebra $C$. 
In the t-matrix framework, generalized scalars are represented by fixed-sized multi-way arrays of complex numbers.  These complex arrays can be added in the usual way, but there is no definition of multiplication satisfying the axioms of a field such as $\mathbb{R}$ or $\mathbb{C}$. 
However, multiplication based on multi-way circular convolution has many but not all of the properties of a field. Multi-way circular convolution differs from the multiplication in a field in that an infinite number of elements have no multiplicative inverse. These complex arrays form a finite-dimensional commutative algebra $C$ under the vector addition, scalar multiplication, and convolution-based multiplication. The elements of the algebra $C$ generalize complex numbers and are referred to as t-scalars.

The bijective map by the multi-way Fourier transform shows that the algebra $C$ of t-scalars under the convolution-based multiplication is isomorphic to an algebra of complex arrays of the same size in which the Hadamard product defines the multiplication.

In effect, the algebra, mapped by the Fourier transform, splits into 
a finite number of copies of $\mathbb{C}$. This splitting allows the construction of 
generalized algorithms for analyzing tensorial data without data unraveling. The so-called 
t-matrices with t-scalar entries have many properties of canonical real or complex matrices. 
In particular, t-matrices can be scaled with a real or complex number, added and multiplied. There are an additive identity and a multiplicative identity of the algebra $C$. 
The generalized rank of a t-matrix is defined by a nonnegative t-scalar, which 
generalizes the canonical rank of a real or complex matrix, and 
is a nonnegative element of a partially ordered set of self-conjugate t-scalars.  
A given t-matrix is invertible if and only if it is square and of full rank over $C$. 
The t-matrices include but are not limited to the generalizations of unitary matrices and Hermitian matrices.

\subsection{Contributions of This Work and Organization of This Article}
This article introduces semisimplicity, a concept in algebra and other algebraical disciplines, to general data analytics, with visual information analysis applications. Launching with a few postulates, one has a generalized paradigm over a semisimple algebra. Using the generalized paradigm,  data analytics can be more effective than with the canonical paradigm over real or complex numbers.

This article shows that the semisimple algebra $C$, called ``t-algebra'', generalizes the field $\mathbb{C}$ and can be represented as a direct product of a finite number of simple algebras all isomorphic to the field $\mathbb{C}$.  
The semisimplicity of the t-algebra $C$ allows
a straightforward backward-compatible generalization of many canonical linear or multilinear structures and algorithms over $\mathbb{C}$.  In the direct product representation of the t-algebra $C$, its idempotent elements play a critical role. Via the idempotent elements of the t-algebra $C$, many generalized algebraic notions, including but not limited to generalized scalars (called t-scalars), generalized rank, generalized norm, generalized orthogonality, are reducible to the corresponding canonical notions defined over the field $\mathbb{C}$. Analogous to their canonical counterparts, generalized matrices over $C$, called t-matrices, can be scaled, added, multiplied, conjugate transposed, and inverted or pseudo-inverted, in a way backward-compatible to their canonical counterparts defined over $\mathbb{C}$. The t-algebra $C$ and the t-matrix framework over it allow us to establish a generalized ``semisimple'' paradigm of data analytics, which is backward-compatible with the canonical paradigm over the field 
$\mathbb{C}$.

To demonstrate the ``semisimple'' paradigm on general visual information 
analysis, we propose spatial solutions for elevating lower-order visual information to 
higher-order and pooling higher-order information to lower-order. With the proposed spatial 
solutions, we adopt generalized algorithms to represent, approximate or analyze images data. 
Our experiments using the generalized algorithms on public datasets show a provable performance increase compared with the corresponding canonical algorithms' results. We also give 
principles on generalizing canonical algorithms and models, including but not limited to CNN 
(Convolutional Neural Network) for classifying visual patterns. 
Besides visual information, if appropriate topological information of each data point is known, the ``semisimple'' paradigm also applies to non-spatially-constrained data.

The remainder of this article is organized as follows. 
The generalized scalars, called t-scalars, their set called t-algebra, and the generalization of complex 
numbers are described in Section \ref{section:tensor-algebra}. 
The idempotent t-scalars and the 
semisimplicity and decomposability of the t-algebra are discussed in Section \ref{section:idem}. 
Generalized matrices with entries of t-scalars, the semisimplicity and decomposability of the modules, and generalized minimization over the t-algebra are discussed in Section 
\ref{section:generalized-matrix-over-talgebra}. In Section \ref{section:Applications-of-T-matrices}, we 
discuss and demonstrate the principles of applying the semisimple paradigm to generalized visual 
information analytics. In Section \ref{section:Experimental-Verifications}, we give provable experimental verifications on public datasets, where 
results by generalized algorithms compare favorably with the canonical counterparts.  We conclude this 
article in Section \ref{section:conclusions}. Finally, a brief discussion on adopting the proposed paradigm on supervised 
classification and neural network is given in an appendix.

\section{T-algebra and T-scalars}
\label{section:tensor-algebra}

This work is a continued effort to complete the t-scalar and t-matrix paradigm proposed by Liao and Maybank \cite{liao2020generalized}. The {\color{black}notations}, index protocols, symbols, and others in the existing work \cite{liao2020generalized} are followed as much as possible.

For example, all indices begin from $1$ rather than $0$. Different symbol subscripts other  than symbol fonts are used
for different data types
since there are many data types
rather than just canonical scalars, vectors, matrices, and tensors.
Interested readers are referred to \cite{liao2020generalized} for more details of these symbol subscripts. 
For the notations not consistent with those or not yet appearing in \cite{liao2020generalized}, we give their descriptions when necessary.

\subsection{T-algebra}

The t-algebra $C$ also referred to as the ring of t-scalars in \cite{liao2020generalized}, generalizes the field of complex numbers $\mathbb{C}$. It shows that many structures over $C$ are algebraically semisimple and can be defined as a direct product of a finite number of simple factors over $\mathbb{C}$. We discuss the semisimplicity and the decomposability of $C$ with more details later in Section \ref{section:idem}.

The genesis of the t-algebra and its elements, called t-scalars, are from the following several postulates.

\begin{definition}[\textbf{Multi-way array}] 	
\label{definition:001}
The generalized scalars, called t-scalars, are order-$N$ arrays of complex numbers belonging to the set $C \equiv \mathbb{C}^{I_1\times \cdots \times I_N}$.
\end{definition}

\begin{definition}[\textbf{Addition}]  
\label{definition:addition}
The addition of t-scalars is identified with the addition bestowed to linear space
namely, given any two t-scalars 
$X_\mathit{T},  Y_\mathit{T} \in C$, their addition $A_\mathit{T} \doteq X_\mathit{T} + Y_\mathit{T} \in C$ is given by the following complex-entry-wise addition.
\begin{equation}
(A_\mathit{T})_{i_1,\cdots,i_N} = (X_\mathit{T})_{i_1,\cdots,i_N} + (Y_\mathit{T})_{i_1,\cdots,i_N} \in \mathbb{C}, 
\;\forall (i_1,\cdots,i_N) \in [\scalebox{0.92}{$I_1$}] \times \cdots \times [\scalebox{0.92}{$I_N$}]
\label{equation:tscalar-addition}
\end{equation}
where $(X_\mathit{T})_{i_1,\cdots,i_N}$ denotes the $(i_1,\cdots,i_N)$-th complex entry of $X_\mathit{T}$ for all $X_\mathit{T} \in C$ and $[I_n] \doteq \{1,\cdots,I_n\}$ for all $n = 1,\cdots,N$.
\end{definition}

\begin{definition}[\textbf{Scalar multiplication}] 
\label{definition:scalar-multiplication}
For each a t-scalar $X_\mathit{T} \in C$ and each scalar $\lambda \in \mathbb{C}$, the scalar 
multiplication 
$Y_\mathit{T} \doteq \lambda \cdot X_\mathit{T} \in C$ is given by the following entry-wise complex multiplication.
\begin{equation}
(Y_\mathit{T})_{i_1,\cdots,i_N} =  \lambda \cdot (X_\mathit{T})_{i_1,\cdots,i_N} \; \in \; \mathbb{C} ,\; \forall\, i_1, \cdots, i_N \;\;.
\label{equation:scalar-multiplication}
\end{equation}
\end{definition}

\begin{definition}[\textbf{Convolutional multiplication}] 
\label{def:multiplication}
The convolutional multiplication of a pair of t-scalars is defined by $N$-way  circular convolution --- for each pair of two t-scalars $X_\mathit{T},  Y_\mathit{T} \in C$, the product $A_\mathit{T} \doteq X_\mathit{T} \circ Y_\mathit{T} \in C$ is given as follows.
\begin{equation}
\begin{matrix}
(A_\mathit{T})_{i_1,\cdots, i_N} = 
\scalebox{0.95}{$\sum\nolimits_{m_1 = 1}^{I_1}$} \cdots 
\scalebox{0.95}{$\sum\nolimits_{m_N = 1}^{I_N}$} 
(X_\mathit{T})_{m_1, \cdots, m_N} \cdot (Y_\mathit{T})_{m'_1, \cdots, m'_N} \;\;\in \mathbb{C} \;
\end{matrix}
\label{equation:tscalar-multiplication}
\end{equation}
where $m'_{n} = \operatorname{mod}(i_n - m_n, I_n) + 1$ for all $n \in [N] \doteq \{1,\cdots,N\}$.   
\end{definition}

The product of $p$  copies of t-scalar $X_\mathit{T}$ is also denoted by the shorthand notation $X_\mathit{T}^{p}$ where $p \geqslant 1$ is an integer.

\textbf{The zero t-scalar and the identity t-scalar.}\;
T-scalars, under the addition, form an abelian group. The additive identity, denoted by $Z_\mathit{T}$, 
is the array of zeros, namely $(Z_\mathit{T})_{i_1,\cdots,i_N} \equiv 0,\; \forall i_1,\cdots,i_N$. It is easy to verify that the t-scalar multiplication is associative, commutative, and distributive to the addition. The multiplicative identity, denoted by $E_\mathit{T}$, is a t-scalar whose inception entry, with the subscript indices $i_1 = \cdots = i_N = 1$, is equal to $1$ and all other entries equal to $0$.

\begin{remarks}[\textbf{T-algebra}]
The addition and scalar multiplication show that t-scalars form a linear space of dimension $K \doteq 
\myK$. Under the addition and t-scalar multiplication, 
$C$ is also a commutative ring. Then, by the definition of algebra, $C$ is a finite-dimensional commutative 
unital algebra over the field $\mathbb{C}$. However, $C$ is not a division algebra because not all non-zero 
t-scalars in $C$ are multiplicatively invertible. For example, all t-scalars with identical complex entries are not multiplicatively invertible. In other words, $C$ can not be a field or even a skew field.  On the other hand, the t-algebra $C$ generalizes the field $\mathbb{C}$ of complex numbers such that $C$ reduces to $\mathbb{C}$ when $I_1 = I_2 = \cdots  = I_N = 1$.  
\end{remarks}

An equivalence of Definition \ref{def:multiplication} is given by the Hadamard product via the Fourier transform. The equivalence is guaranteed by the convolution theorem \cite{bracewell1986fourier}. More precisely, the Fourier transform is an isomorphism of algebra $F: (C, +, \,\cdot, \circ) \rightarrow (C, +, \,\cdot, *)$ such that for all $X_\mathit{T}, Y_\mathit{T} \in C$, the following condition holds.       
\begin{equation}
\begin{aligned}
& F(X_\mathit{T} \circ Y_\mathit{T} ) = F(X_\mathit{T}) * F(Y_\mathit{T}) \,\in\, \mathbb{C}^{I_1\times \cdots \times I_N}  
\end{aligned}
\end{equation}
where $*$ denotes the Hadamard multiplication and is given by entry-wise multiplication of complex arrays 
$F(X_\mathit{T})$ and $F(X_\mathit{T})$ in $\mathbb{C}^{I_1\times \cdots \times I_N} $. More precisely, let $\tilde{X}_\mathit{T} 
\doteq F(X_\mathit{T})$, $\tilde{Y}_\mathit{T} \doteq F(Y_\mathit{T})$
and $\tilde{C}_\mathit{T} \doteq F(X_\mathit{T}) * F(Y_\mathit{T})$. Then, $(\tilde{C}_\mathit{T})_{i_1,\cdots,i_N} = 
(\tilde{X}_\mathit{T})_{i_1,\cdots,i_N} \cdot (\tilde{Y}_\mathit{T})_{i_1,\cdots,i_N} \in \mathbb{C} $ for all 
$i_1,\cdots,i_N$.

The Fourier transform is an isomorphism 
defined by the $N$-mode multiplication of tensors, which sends each element $X_\mathit{T} \in C$  to $\tilde{X}_\mathit{T} \in C$ as follows. 
\begin{equation}
\tilde{X}_\mathit{T} \doteq  
F(X_\mathit{T}) \doteq  X_\mathit{T} \times_1 W_\mathit{mat}^{(I_1)} \cdots \times_k W_\mathit{mat}^{(I_n)} 
\cdots \times_N W_\mathit{mat}^{(I_N)} \in 
C \equiv 
\mathbb{C}^{I_1\times \;\cdots\; \times I_N}
\label{equation:fourier-transform-tscalar}
\end{equation}
where $W_\mathit{mat}^{(I_n)} \in \mathbb{C}^{I_n\times I_n} $ denotes the $I_n\times I_n$ 
Fourier matrix  for all $n \in [N]$.

The $(m_1, m_2)$-th 
complex entry of the matrix $W_\mathit{mat}^{(I_n)}$ is given by
\begin{equation}
\left( \scalebox{0.92}{$W_\mathit{mat}^{(I_n)}$}  \right)_{m_1, m_2} = e^{-2\pi i
\cdot (m_1 - 1) \cdot (m_2 - 1) \cdot I_{n}^{-1}  } 
\in \mathbb{C}
\;,\;\; \text{for all}\; m_1, m_2 \;\;. 
\end{equation}

The inverse transform $F^{-1}: (C, +, \,\cdot, *) \rightarrow (C, +, \,\cdot, \circ)$
is given by the following $N$-mode multiplication for tensors as follows. 
\begin{equation}
X_\mathit{T} \doteq F^{-1}(\tilde{X}_\mathit{T}) =  \tilde{X}_\mathit{T} \times_1 
\left( \scalebox{0.92}{$W_\mathit{mat}^{(I_1)}$}   \right)^{-1} \cdots \times_n 
\left( \scalebox{0.92}{$W_\mathit{mat}^{(I_n)}$} \right)^{-1} 
\cdots  \times_N \left( \scalebox{0.92}{$W_\mathit{mat}^{(I_N)}$} \right)^{-1}
\in C \equiv \mathbb{C}^{I_1\times \cdots \times I_N}
\label{equation:inverse-fourier-transform-tscalar}
\end{equation}
where $\left(\scalebox{0.92}{$W_\mathit{mat}^{(I_n)}$} \right)^{-1}$ denotes  
the inverse of the matrix 
$W_\mathit{mat}^{(I_n)} $ for all $n \in [N]$.

By definition, the Fourier transform is a linear mapping, and
the following equality holds for all $X_\mathit{T} \in
C \equiv \mathbb{C}^{I_1\times \cdots \times I_N}$, 
\begin{equation}
\raisebox{0.1em}{$\|X_\mathit{T}\|_F$} =
K^{-{1}/{2}}   
\,\cdot\, 
\| \tilde{X}_\mathit{T} \|_F 
\end{equation}
where $K \doteq \myK$ and $\|\cdot\|_F $ denotes the 
canonical Frobenius norm for the normed linear space $C \equiv \mathbb{C}^{I_1\times \cdots \times I_N}\,$. 

\subsection{T-scalars: Generalization of Complex Numbers}
The t-algebra $C$ 
reduces to the field $\mathbb{C}$ when $I_1 = \cdots = I_N = 1$. 
Besides the fundamental operations addition and multiplication, one 
can generalize more notions of complex numbers over the t-algebra $C$.

\textbf{Conjugation}.\;
One of the generalizations is the notion of conjugation over $C$. The conjugation is an involutory antiautomorphism $\varphi$ of $C$ such that the following conditions hold  
\begin{equation}
\begin{aligned}
\varphi(E_\mathit{T}) = E_\mathit{T} \\
\varphi(\varphi(X_\mathit{T})) = X_\mathit{T} \\
\varphi(X_\mathit{T} \circ Y_\mathit{T}) = \varphi(Y_\mathit{T}) \circ \varphi(X_\mathit{T}) \\
\varphi(\alpha \cdot X_\mathit{T} \,+\, \beta \cdot Y_\mathit{T} ) = \bar{\alpha} \cdot \varphi(X_\mathit{T}) + \bar{\beta} \cdot 
{\color{black}\varphi(Y_\mathit{T})} \\
\end{aligned}  
\end{equation}
for all $\alpha, \beta \in \mathbb{C}$ and $X_\mathit{T}, Y_\mathit{T} \in C$.

Also, note that 
the antiautomorphism $\varphi$ is also automorphic since 
the t-algebra $C$ is commutative. In other words, the antiautomorphism  condition $\varphi(X_\mathit{T} \circ 
Y_\mathit{T}) = \varphi(Y_\mathit{T}) \circ \varphi(X_\mathit{T})$ is equivalent to the automorphism  condition  
$\varphi(X_\mathit{T} \circ Y_\mathit{T}) = \varphi(X_\mathit{T}) \circ \varphi(Y_\mathit{T})$.

Let the map  
$\varphi: C \rightarrow C,   X_\mathit{T} \mapsto \varphi(X_\mathit{T})$ be a homomorphism from $C$ to itself, such that 
\begin{equation}
\big(\raisebox{0.05em}{$\varphi(X_\mathit{T})$}  \big)_{i_1,\cdots, i_N} = \overline{(X_\mathit{T})_{m_1,\cdots,\,m_N}} \;\in\; \mathbb{C} 
\label{equation:conjugate}
\end{equation}
where $m_n \doteq \operatorname{mod}(1- i_n, I_n) + 1$ for all $n \in  [N]$.

The homomorphism $\varphi$ defined as in equation (\ref{equation:conjugate})
satisfies all the conditions of the notion of conjugation. 
When $I_1 = \cdots =I_N = 1$, the conjugation $\varphi$ over $C$ reduces to the conjugation over complex numbers.

To comply with the standard notation of \staralgebra, we use the notation 
$ \myconj{X} 
\doteq \varphi(X_\mathit{T}) $ for the conjugate of a t-scalar $X_\mathit{T} \in C$.~~\footnote{\,The original notation of the conjugate of a t-scalar $X_\mathit{T}$
in \cite{liao2020generalized} is $\operatorname{conj}(X_\mathit{T})$. }

\textbf{Self-conjugate}.\;
The conjugate of a t-scalar can be used for characterizing a particular type of t-scalars  --- a 
t-scalar $X_\mathit{T}$ is called self-conjugate if $\myconj{X} = X_\mathit{T} $. 
It is immediately verified that $Z_\mathit{T}$ and $E_\mathit{T}$ are both self-conjugate. A necessary and 
sufficient condition for a t-scalar $X_\mathit{T}$ to be self-conjugate is that the Fourier transform $F(X_\mathit{T})$ 
is a real array \cite{liao2020generalized}.

Let the set of self-conjugate t-scalars be $C^\mathit{sc} = \{X_\mathit{T} \in C \,|\, \myconj{X} = X_\mathit{T}  \}$. 
The set $C^\mathit{sc}$ under the addition, scalar multiplication, and t-scalar multiplication is a 
subalgebra of $C$. The t-algebra $C$ is a free complex algebra of dimension $K \doteq 
\myK$. On the other hand, the 
subalgebra $C^\mathit{sc}$ is a free real algebra of the same dimension $K$
over $\mathbb{R}$ since the subalgebra $C^\mathit{sc}$ is 
isomorphic to the algebra $\scalebox{0.95}{$F(C^\mathit{sc})$} \doteq \{F(X_\mathit{T}) \,|\,  X_\mathit{T} \in 
C^\mathit{sc}\}$. 
That $C^\mathit{sc}$ is a $K$-dimensional real algebra does not necessarily mean 
that all self-conjugate t-scalar must be of real arrays. 
The subalgebra $C^\mathit{sc}$ contains real arrays 
if the parameters $I_1,\cdots,I_N$ are in the set $\{1, 2\}$.

When $I_1 = \cdots = I_N = 1$, the subalgebra $C^\mathit{sc}$ reduces to the field $\mathbb{R}$ of real numbers. As a generalization of real numbers, $C^\mathit{sc}$ helps establish many fundamental notions over $C$.

\textbf{Real part and imaginary part}.\;
Each t-scalar $X_\mathit{T}$ is representable by two unique self-conjugate t-scalars. More precisely, the 
equality $X_\mathit{T} \equiv \frac{X_\mathit{T} + \myconj{X} }{2} + 
i \cdot \frac{X_\mathit{T} - \myconj{X} }{2i} $ holds for all $X_\mathit{T} \in C$.

The self-conjugate t-scalars 
$\frac{X_\mathit{T} + \myconj{X}  }{2}$  and 
$\frac{X_\mathit{T} - \myconj{X}   }{2i}$ 
are respectively called the real part and the imaginary part of $X_\mathit{T}$. 
Let $\mathit{Re}(X_\mathit{T}) \doteq \frac{X_\mathit{T} + \myconj{X} }{2}$ and $\mathit{Im}(X_\mathit{T}) \doteq \frac{X_\mathit{T} - \myconj{X} }{2i}$ for all $X_\mathit{T} \in C$. 
Then, the following equations hold for all t-scalars $X_\mathit{T}, Y_\mathit{T} \in C$, 
\begin{equation}
\resizebox{0.9\textwidth}{!}{$
\begin{matrix}
\myconj{X} =  \mathit{Re}(X_\mathit{T}) - {\color{black}i \cdot \mathit{Im}(X_\mathit{T})}  	\vspace{0.3em}\\
\myconj{X} \circ X_\mathit{T} =  \mathit{Re}(X_\mathit{T})^{2} + \mathit{Im}(X_\mathit{T})^{2}  	\vspace{0.3em}\\
X_\mathit{T} + Y_\mathit{T} = \big( \scalebox{0.9}{$\mathit{Re}(X_\mathit{T}) + \mathit{Re}(Y_\mathit{T})$}  \big) + i\cdot \big(
 \scalebox{0.9}{$\mathit{Im}(X_\mathit{T}) + \mathit{Im}(Y_\mathit{T})$}   \big)   \vspace{0.3em}\\
X_\mathit{T} \circ Y_\mathit{T} = \big( \scalebox{0.9}{$\mathit{Re}(X_\mathit{T}) \circ \mathit{Re}(Y_\mathit{T}) - \mathit{Im}(X_\mathit{T}) \circ \mathit{Im}(Y_\mathit{T})$}   \big) 
+ i\cdot \big( \scalebox{0.9}{$\mathit{Im}(X_\mathit{T}) \circ \mathit{Re}(Y_\mathit{T}) + \mathit{Re}(X_\mathit{T}) \circ \mathit{Im}(Y_\mathit{T})$}
\big)\;.  
\end{matrix}
\label{equation:complex-like-identities}
$}
\end{equation}

\textbf{Nonnegative t-scalar}.\;
Over the subalgebra $C^\mathit{sc}$, one can generalize the notion of nonnegative real numbers  --- a t-scalar  $Y_\mathit{T} \in C^\mathit{sc}$ is said nonnegative if and only if there exists a t-scalar $X_\mathit{T} \in C$ such that the condition
$Y_\mathit{T} = \myconj{X} \circ X_\mathit{T} $ holds.
It is easy to verify that both $Z_\mathit{T}$ and $E_\mathit{T}$ are nonnegative. Furthermore, 
any t-scalar in the form of $\mathit{Re}(X_\mathit{T})^{2}  + \mathit{Im}(X_\mathit{T})^{2} $ is nonnegative.
A t-scalar $X_\mathit{T} \in C^\mathit{sc}$ is nonnegative iff its Fourier transform $F(X_\mathit{T})$ only contains nonnegative real entries \cite{liao2020generalized}.

Let the set of nonnegative t-scalars be   
$S^\mathit{nonneg} \doteq \{Y_\mathit{T} \,|\, Y_\mathit{T} = \myconj{X} \circ X_\mathit{T}, X_\mathit{T} \in C \}$. 
The set   
$S^\mathit{nonneg}$ is a commutative submonoid of $C^\mathit{sc}$  under the t-scalar addition and the t-scalar multiplication. When 
$I_1 = \cdots = I_N = 1$, the monoid $S^\mathit{nonneg}$ reduces to the monoid of nonnegative real numbers. \footnote{A monoid is a set equipped with an associative binary operation and an identity element \cite{hungerford1980texts02}. }

\textbf{Partial order}.\;
The field $\mathbb{R}$ is a totally ordered set under the usual binary relation ``$\leqslant$''. To be a 
well-behaved generation of the field $\mathbb{R}$, the algebra $C^\mathit{sc}$ needs to be ordered under a binary relation ``$\le $'',  defined for at least some pairs of its elements. The notion of $S^\mathit{nonneg}$ can help define such the relation ``$\le$''. More formally, the binary relation `$\le$'' on $C^\mathit{sc}$ 
defines a proper subset of the Cartesian product $C^\mathit{sc} \times C^\mathit{sc}$ such that $X_\mathit{T} \le Y_\mathit{T}$ if and only if $(Y_\mathit{T} - X_\mathit{T}) \in S^\mathit{nonneg}$ for all $X_\mathit{T}, Y_\mathit{T} \in C^\mathit{sc}$.

By this definition, it is immediately verified that $Z_\mathit{T} \le X_\mathit{T}$ for all $X_\mathit{T} \in 
S^\mathit{nonneg}$, namely, the t-scalar $Z_\mathit{T}$ is the least element of $S^\mathit{nonneg}$. The 
relation $Z_\mathit{T} \le X_\mathit{T}$ is synonymous with the claim that the t-scalar $X_\mathit{T}$ is nonnegative.

\textbf{Nonpositive t-scalar}.\
The binary relation ``$\le$'' is reflexive, antisymmetric, and 
transitive. Those properties qualify ``$\le$'' a relation of a partial order. 
The partial order helps define nonpositive t-scalars ------ a t-scalar $X_\mathit{T} \in C^\mathit{sc}$ is called 
nonpositive if and only if $X_\mathit{T} \le Z_\mathit{T} $, or equivalently, $-1\cdot{}X_\mathit{T} \in 
S^\mathit{nonneg}$.

Let $S^\mathit{nonpos} \doteq \{-X_\mathit{T} \,|\, X_\mathit{T} \in S^\mathit{nonneg} \} \equiv \{X_\mathit{T} \,|\, X_\mathit{T} \le Z_\mathit{T}, X_\mathit{T} 
\in C^\mathit{sc} \}$ be the set of nonpositive t-scalars. The set $S^\mathit{nonpos}$ is a monoid under the 
addition, with the additive identity $Z_\mathit{T}$ being the greatest element of $S^\mathit{nonpos}$. 
A t-scalar $X_\mathit{T}$ is nonpositive iff its Fourier transform $F(X_\mathit{T})$ is a 
nonpositive real array.

Both $S^\mathit{nonneg}$ and $S^\mathit{nonpos}$ are proper subsets of $C^\mathit{sc}$. There is usually a ``gap'' 
between $S^\mathit{nonneg}$ and $S^\mathit{nonpos}$ such that $S^\mathit{nonneg} \cup S^\mathit{nonpos} \neq C^{sc}$ 
unless $I_1 = \cdots = I_N = 1$. 
Given two self-conjugate t-scalars $X_\mathit{T}$ and $Y_\mathit{T}$,  
if and only if their subtraction falls in this ``gap'', namely $X_\mathit{T} - Y_\mathit{T} \notin (S^\mathit{nonneg} \cup 
S^\mathit{nonpos})$, $X_\mathit{T}$ and $Y_\mathit{T}$ are 
called incomparable under the partial order ``$\le$''.

When $I_1 =\cdots  = I_N = 1$, the partial order under the relation ``$\le $'' reduces to the usual total 
order of real numbers under the relation ``$\leqslant$''. The set $S^\mathit{nonneg}$ reduces to the 
interval $[0, +\infty)$ of real numbers, and $S^\mathit{nonpos}$ reduces to the interval $(-\infty, 0]$ 
of real numbers.

\textbf{Nonnegative $p$-th root of a nonnegative t-scalar}.\;
For each 
integer $p \geqslant 1$ and   
nonnegative t-scalar $Y_\mathit{T}$, there is a unique 
nonnegative t-scalar $X_\mathit{T}$ such that 
$Y_\mathit{T} = X_\mathit{T}^{p}\,$. The proof of the unique existence for $p = 2$ is given in 
\cite{liao2020generalized}, and the proof for $p > 2$ can be given analogously. The nonnegative t-scalar 
$X_\mathit{T}$ is called the 
$p$-th arithmetic root of the nonnegative t-scalar $Y_\mathit{T}$ and is denoted by 
\begin{equation} 
X_\mathit{T} \doteq \sqrt[p]{ \scalebox{0.9}{$\,Y_\mathit{T}$ } } \equiv Y_\mathit{T}^{1/p} \;\;.
\end{equation}

\textbf{Norm of a t-scalar}.\;
The notions of nonnegative t-scalars and nonnegative roots help define 
the norm of a t-scalar, 
also called the absolute value of a t-scalar ------ for all t-scalar $X_\mathit{T} \in C$, its norm 
$|X_\mathit{T}| \doteq r(X_\mathit{T})$ is a nonnegative t-scalar defined by
\begin{equation}
|X_\mathit{T}| \doteq r(X_\mathit{T}) 
\doteq 
\scalebox{1.1}{$\sqrt[2]{\scalebox{0.8}{$\myconj{X} \circ X_\mathit{T}$} }$}
\equiv  \sqrt[2]{\mathit{Re}(X_\mathit{T})^{2}  +  \mathit{Im}(X_\mathit{T})^{2} }  \,\in\, S^\mathit{nonneg}\;\;.
\label{equation:absolute-value}
\end{equation}

It is easy to verify that the following equalities, analogous to their canonical counterparts, hold for all $X_\mathit{T}, Y_\mathit{T} \in C$ and $\alpha \in \mathbb{C}$,   
\begin{equation}
\scalebox{0.95}{$
\begin{aligned}
r(\alpha \cdot X_\mathit{T}) = |\alpha| \cdot {\color{black}r(X_\mathit{T})}   \vspace{0.3em}\\
r(X_\mathit{T}) = Z_\mathit{T} \;\text{iff}\; X_\mathit{T} = Z_\mathit{T} \vspace{0.3em}\\
r(X_\mathit{T} \circ Y_\mathit{T}) = r(X_\mathit{T}) \circ r(Y_\mathit{T})  \\
r(X_\mathit{T} + Y_\mathit{T}) \le r(X_\mathit{T}) + r(Y_\mathit{T}) \\
\end{aligned}
$}\;\;\;.
\end{equation}

When $I_1 = \cdots = I_N = 1$, the norm $r(\cdot)$ reduces to the  absolute value of a complex 
number.

\textbf{Inner product of two t-scalars}.\;
Following the vein of equation (\ref{equation:absolute-value}), one can define the notion of orthogonality for a pair of t-scalars. 
First, the polarization identity 
\begin{equation}
\resizebox{0.9\textwidth}{!}{$
\myconj{X} \circ Y_\mathit{T} =   
\frac{1}{4}
\cdot \Big( 
r^{2}(X_\mathit{T} + Y_\mathit{T})
- i \cdot
r^{2}(X_\mathit{T} + i \cdot Y_\mathit{T}  ) 
-r^{2}( X_\mathit{T} - Y_\mathit{T} )  
+\,i \cdot
r^{2}(X_\mathit{T} - i \cdot Y_\mathit{T} )
\Big)
$}
\end{equation} 
holds for all t-scalars $X_\mathit{T}, Y_\mathit{T} \in C$.

By the polarization identity, 
we define the inner product of 
any pair t-scalars $X_\mathit{T}, Y_\mathit{T} \in C$ by 
\begin{equation}
\psi(\scalebox{0.92}{$X_\mathit{T}, Y_\mathit{T}$} ) = \scalebox{0.92}{$\myconj{X} \circ Y_\mathit{T}$} 
\;.   
\end{equation} 
. 

\newcommand{\myinnerXX}{\psi(\scalebox{0.95}{$X_\mathit{T}, X_\mathit{T}$} )}
\newcommand{\myinner}{\psi(\scalebox{0.95}{$X_\mathit{T}, Y_\mathit{T}$} )}
\newcommand{\myinnerYX}{\psi(\scalebox{0.95}{$Y_\mathit{T}, X_\mathit{T}$} )}

The following identities, 
analogous to their canonical counterparts for a linear space, hold
for all $X_\mathit{T}, Y_\mathit{T}, A_\mathit{T}, B_\mathit{T}  \in C$,
\begin{equation}
\begin{aligned} 
r(X_\mathit{T}) = \sqrt[2]{\myinnerXX } \\
\myinner =  \big(\myinnerYX \big)^{*} \\
\psi( \scalebox{0.95}{$A_\mathit{T} \circ X_\mathit{T}, B_\mathit{T} \circ Y_\mathit{T}$}   ) 
= \psi(\scalebox{0.95}{$A_\mathit{T}, B_\mathit{T}) \circ \psi(X_\mathit{T}, Y_\mathit{T}$}  ) \\
\psi(\scalebox{0.95}{$X_\mathit{T} + Y_\mathit{T}, A_\mathit{T} + B_\mathit{T}$}   ) 
= \psi(\scalebox{0.95}{$X_\mathit{T}, A_\mathit{T}$} ) 
+ \psi(\scalebox{0.95}{$X_\mathit{T}, B_\mathit{T}$} ) + \psi(\scalebox{0.95}{$Y_\mathit{T}, A_\mathit{T}$} ) 
+ \psi(\scalebox{0.95}{$Y_\mathit{T}, B_\mathit{T}$} ) 
\end{aligned}\;\;\;.
\end{equation}

The inner product $\psi: \scalebox{0.95}{$(X_\mathit{T}, Y_\mathit{T})$} \mapsto \scalebox{0.9}{$\myconj{X} \circ 
Y_\mathit{T}$} $ is employed to define the 
notion of orthogonal t-scalars, which is used in the decomposition of  
the t-algebra $C$ to a finite number of simple algebras. Two t-scalars $X_\mathit{T}, Y_\mathit{T} \in C$ are said to be orthogonal 
over the t-algebra $C$, 
iff their inner product is equal to $Z_\mathit{T}$, more precisely,
\begin{equation}
\psi(X_\mathit{T}, Y_\mathit{T}) = Z_\mathit{T}\;.
\label{equation:orthogonality-on-the-ring}
\end{equation}

The condition $\psi(X_\mathit{T}, Y_\mathit{T}) = Z_\mathit{T}$ is equivalent to the condition $X_\mathit{T} \circ Y_\mathit{T}  = Z_\mathit{T}$ 
for all $X_\mathit{T}, Y_\mathit{T} \in C$.  The trivial case of t-scalar orthogonality is that $Z_\mathit{T}$ is orthogonal 
to all t-scalars. In non-trivial cases of t-scalar orthogonality where $X_\mathit{T}$ and $Y_\mathit{T}$ are not equal to $Z_\mathit{T}$, both $X_\mathit{T}$ and $Y_\mathit{T}$ must be non-invertible.

Since the t-algebra $C$ is a ring and a linear space, the notion of 
inner product $\psi: C\times C \rightarrow C$ over the ring $C$ has 
a canonical counterpart over the linear space $C$. 
The canonical inner product
$\langle \cdot, \cdot \rangle: C\times C \rightarrow \mathbb{C}$ is a sesquilinear 
form defined by
\begin{equation}
\langle X_\mathit{T}, Y_\mathit{T} \rangle \doteq 
\scalebox{1}{$
\sum\nolimits_{(i_1,\cdots,i_N) \in [I_1] \times \cdots \times [I_N] }
$} 
\,\overline{(\raisebox{-0.1em}{$X_\mathit{T}$})_{i_1,\cdots,i_N}}  \,\cdot\, (Y_\mathit{T})_{i_1,\cdots,i_N}   
\end{equation} 
for all t-scalars
$X_\mathit{T}, Y_\mathit{T} \in C$.

Two t-scalars 
$X_\mathit{T}$ and $Y_\mathit{T}$, as two elements of a linear space, are said orthogonal over the linear space if and 
only if $\langle X_\mathit{T}, Y_\mathit{T} \rangle = 0$.

Since the t-algebra $C$ is both a linear space and a ring, 
orthogonality over the linear space $C$ is a companion notion of orthogonality over the ring $C$. Orthogonality over the ring is a sufficient condition for orthogonality 
over the linear space. 
When $I_1 = \cdots = I_N = 1$, the two notions of orthogonalities become identical.

\section{Decomposition of T-algebra via Direct Product}
\label{section:idem}

\subsection{Idempotence}
The orthogonality introduced by equation (\ref{equation:orthogonality-on-the-ring}) plays an essential 
role in decomposing the t-algebra $C$ to a finite number of simple algebras. To this end, we introduce the notion of idempotence over the ring $C$. 
An element ${P}_\mathit{T} \in C$ is called idempotent iff ${P}_\mathit{T} \circ {P}_\mathit{T} = {P}_\mathit{T} $.

It is easy to follow that a t-scalar ${P}_\mathit{T}$ is 
idempotent iff its Fourier transform $F(P_\mathit{T})$ is an array with entries either $0$ or $1$, and 
the t-scalars $Z_\mathit{T}$ and $E_\mathit{T}$ 
are idempotent. All idempotent t-scalars are nonnegative and form a multiplicative 
monoid with the identity $E_\mathit{T}$.

Let $S^\mathit{idem} \doteq \{X_\mathit{T} \,|\, X_\mathit{T} \circ X_\mathit{T} = X_\mathit{T}, X_\mathit{T} \in C \}$ be the set of all idempotent 
t-scalars. 
The cardinality of $S^\mathit{idem}$  is equal to 
$2^{K}$, where  $K \doteq \myK$. 
It also shows that, given any idempotent t-scalar ${P}_\mathit{T}$, $(E_\mathit{T} - {P}_\mathit{T})$ is also idempotent. 
Besides being both idempotent, $P_\mathit{T}$ and $E_\mathit{T} - P_\mathit{T}$ are orthogonal. Namely, 
$\psi({P}_\mathit{T}, E_\mathit{T} - {P}_\mathit{T} ) = Z_\mathit{T}$ holds for all ${P}_\mathit{T} \in S^\mathit{idem}$.

Let
$\myperp{P} \doteq E_\mathit{T} - P_\mathit{T}$ 
for all $P_\mathit{T} \in S^\mathit{idem}$. Then,  it shows that each t-scalar $Y_\mathit{T} \in C$ can be written as a sum of two 
orthogonal constituents in the form 
\begin{equation}
Y_\mathit{T} = P_\mathit{T} \circ Y_\mathit{T} + \myperp{P} \circ Y_\mathit{T} \equiv P_\mathit{T} \circ Y_\mathit{T} +  (E_\mathit{T} - P_\mathit{T}) \circ Y_\mathit{T} 
\label{equation:orthogonal-two-compotents}
\end{equation} 
such that $\psi(\scalebox{0.95}{$P_\mathit{T} \circ Y_\mathit{T}$}, \scalebox{0.95}{$\myperp{P} \circ Y_\mathit{T}$}  ) = Z_\mathit{T} $ 
for all $P_\mathit{T} \in S^\mathit{idem}$ and $Y_\mathit{T} \in C$.

\newcommand{\myspace}{\vspace{0.35em}}

Further, the following equalities hold for all
$\lambda \in \mathbb{C}$ and $X_\mathit{T}, Y_\mathit{T} \in C$. 
\begin{equation}
\scalebox{1}{$
\begin{aligned}
\myconj{X} = \big(\scalebox{0.92}{$P_\mathit{T} \circ X_\mathit{T}$}  \big)^{*} + 
\big(\scalebox{0.92}{$\myperp{P} \circ X_\mathit{T}$}  \big)^{*}   \myspace\\
r(X_\mathit{T})  = r(\scalebox{0.92}{$P_\mathit{T} \circ X_\mathit{T}$} ) + 
r(\myperp{P} \circ X_\mathit{T} )  \myspace\\
\lambda \cdot X_\mathit{T} = \lambda \cdot  
\big(\scalebox{0.92}{$P_\mathit{T} \circ X_\mathit{T} $}   \big) + 
\lambda \cdot 
\big( \scalebox{0.92}{$\myperp{P} \circ X_\mathit{T} $ } \big)  \myspace\\
X_\mathit{T} + Y_\mathit{T} = \big(
\scalebox{0.92}{$P_\mathit{T} \circ X_\mathit{T} + P_\mathit{T} \circ Y_\mathit{T}$}  
\big) +  
\big(
\scalebox{0.92}{$\myperp{P} \circ X_\mathit{T} + \myperp{P} \circ  Y_\mathit{T} $}
\big)     \myspace\\
\psi(X_\mathit{T}, Y_\mathit{T}) = 
\psi(\scalebox{0.92}{$P_\mathit{T} \circ X_\mathit{T},\, P_\mathit{T} \circ Y_\mathit{T}$}    )  
 + \psi(\scalebox{0.92}{$\myperp{P} \circ X_\mathit{T},\,  \myperp{P} \circ Y_\mathit{T}$}  )  \myspace\\
X_\mathit{T} \circ  Y_\mathit{T} 
= \big( \scalebox{0.92}{$P_\mathit{T} \circ X_\mathit{T}$} \big) \circ  
\big(\scalebox{0.92}{$P_\mathit{T} \circ Y_\mathit{T}$}   \big) +  
\big(\scalebox{0.92}{$\myperp{P} \circ X_\mathit{T}$}   \big) \circ  
\big(\scalebox{0.92}{$\myperp{P} \circ Y_\mathit{T}$}  \big)  \myspace\\
X_\mathit{T} \le Y_\mathit{T}  \Leftrightarrow   
P_\mathit{T} \circ X_\mathit{T} \le  P_\mathit{T} \circ Y_\mathit{T} 
\text{\;\;and\;\;}
\myperp{P} \circ X_\mathit{T} \le  \myperp{P}  \circ Y_\mathit{T} 
\end{aligned}\;\;\;\;. 
$}
\label{equation:direct-product-equalities}
\end{equation}

\subsection{Direct Product and Semisimplicity}
By the observations as in equation (\ref{equation:direct-product-equalities}), the t-algebra $C$ is 
a direct product of two algebras $C(P_\mathit{T})$ and $C(\myperp{P})$, written as follows. 
\begin{equation} 
C = C(P_\mathit{T}) \times C(\myperp{P})
\label{equation:direct-product}
\end{equation}
where $C(P_\mathit{T}) \doteq P_\mathit{T} \circ C \doteq \{P_\mathit{T} \circ Y_\mathit{T} \,|\, Y_\mathit{T} \in C  \} $ 
and $C(\myperp{P}) \doteq \myperp{P} \circ C \doteq \{\myperp{P} \circ Y_\mathit{T} \,|\, Y_\mathit{T} \in C  \}$ for all $P_\mathit{T} \in S^\mathit{idem}$. \footnote{For most finite-dimensional algebraic structures, the notions of direct product and direct sum are equivalent. However, on the underlying ring of $C$, the two notions are not equivalent. Interested readers are referred to \cite{lang2002graduate} for a relevant discussion.
}

For all $P_\mathit{T} \in S^\mathit{idem}$, under the t-scalar addition and 
multiplication, $C(P_\mathit{T})$ and $C(\myperp{P})$ 
are both principal ideals of the underlying ring of $C$. It implies that $C(P_\mathit{T})$ and $C(\myperp{P})$ are closed under the t-scalar 
addition and multiplication.

Under the t-scalar addition, $C(P_\mathit{T})$ is a subgroup of the underlying additive group of $C$. 
On the other hand, 
the equality $X_\mathit{T} \circ P_\mathit{T} = X_\mathit{T}$ holds  
for all $X_\mathit{T} \in C(P_\mathit{T})$. 
Hence, $C(P_\mathit{T})$ is a ring with the additive 
identity $Z_\mathit{T}$ and the multiplicative identity $P_\mathit{T}$. A similar 
conclusion is obtained that $C(\myperp{P})$ is also a 
ring with the additive identity $Z_\mathit{T}$ and the multiplicative identity $\myperp{P}$.

However, usually, neither $C(P_\mathit{T})$ nor $C(P_\mathit{T}^{\perp})$ is a subring of $C$ 
since $E_\mathit{T}$ is not an element of either $C(P_\mathit{T})$ or $C(\myperp{P})$ unless 
that one of $C(P_\mathit{T})$ and $C(\myperp{P})$ is equal to $C$, and the other is just a singleton set $\{Z_\mathit{T}\}$.

\textbf{Orthogonal algebras}.\;
The 
algebra $C(\myperp{P})$ is the orthogonal complement of the algebra $C(P_\mathit{T})$ in the sense that 
\begin{equation}
\begin{matrix}
C(P_\mathit{T}) \,\cap\, C(\myperp{P} ) = \{Z_\mathit{T}\}   \vspace{0.3em} \\
C(\myperp{P}) \equiv  \{X_\mathit{T} \in C \;|\; \psi(X_\mathit{T}, Y_\mathit{T}) = Z_\mathit{T},\, \forall\, Y_\mathit{T}  \in 
C(P_\mathit{T})  \}  
\end{matrix} \;\;\;\;.
\end{equation}

\textbf{Primitive idempotent t-scalars}.\;
By equation (\ref{equation:orthogonal-two-compotents}), each idempotent t-scalar $P_\mathit{T}$ can be written as a sum of 
two orthogonal idempotent t-scalars, more precisely,
\begin{equation}
P_\mathit{T} = X_\mathit{T} + Y_\mathit{T}
\label{equation:idempotent-decomposition}
\end{equation} 
such that $X_\mathit{T}, Y_\mathit{T} \in S^\mathit{idem}$ and  $\psi(X_\mathit{T}, Y_\mathit{T}) = Z_\mathit{T}$ for all $P_\mathit{T} \in 
S^\mathit{idem}$.

An idempotent t-scalar $P_\mathit{T}$ is called primitive, if and only if $P_\mathit{T}$ can not be written as a sum of 
two non-zero orthogonal idempotent t-scalars. By definition, it is easy to show that $Z_\mathit{T}$ is a primitive 
idempotent t-scalar.

There are $K \doteq \myK$ non-zero primitive idempotent t-scalars. Let 
$\myQ{1}, \cdots, \myQ{K} $
be these t-scalars, and 
$S^\mathit{pidem} \doteq  \{\myQ{1}, \cdots,  \myQ{K}   \}$
be the set of them. 
It is easy to verify that the Fourier transform $F(\myQ{k})$ contains only one entry of $1$ and other 
entries of $0$ for all $\myQ{k} \in S^\mathit{pidem}$.

Any two elements of $S^\mathit{pidem}$ are orthogonal to each other and 
are incomparable under 
the partial order ``$\le$''.  
Each element of $S^\mathit{pidem}$
is multiplicatively non-invertible and one of the $K$ minimal elements of the poset $S^\mathit{idem} \setminus \{Z_\mathit{T} 
\}$. Furthermore, 
the following identities hold 
\begin{equation}
E_\mathit{T} = \scalebox{1}{$
\sum\nolimits_{k=1}^{K} \myQ{k} 
$}
\end{equation}
and 
\begin{equation}
Y_\mathit{T} \equiv Y_\mathit{T} \circ E_\mathit{T} 
\equiv 
\scalebox{1}{$
\sum\nolimits_{k=1}^{K} Y_\mathit{T}  \circ \myQ{k} 
$}
\;,\;\forall Y_\mathit{T} \in C\;.
\label{equation:decompositionOverTheRing}
\end{equation}

These primitive elements $\myQ{1}, \cdots,  \myQ{K} $ play an essential role in decomposing $C$ and all $C$-modules.

Via direct product, one can discuss the semisimplicity of $C$ and the modules over $C$. Semisimplicity is 
a concept with a rigorous definition in mathematical disciplines such as linear algebra, abstract 
algebra, representation theory, etc. In brief, a semisimple object can be represented as a non-trivial direct 
product of simple objects. Simple objects are non-representable by a non-trivial direct product.

In this article, we are only concerned with semisimple algebras and modules. One
can find a rigorous definition of semisimple algebras and semisimple modules in Chapter 4 of a recent book by Karin Erdmann and Thorsten Holm \cite{erdmann2018algebras} or equivalently in Chapter IX of Thomas Hungerford's time-honored book on algebra \cite{hungerford1980texts}. 
For the reader’s convenience, the relevant definition of semisimplicity is given as follows.

\begin{definition}[Simple algebra] 
\label{definition:SemisimplicityOfC001}
A non-zero algebra is called 
simple or irreducible if the algebra has no two-sided ideals besides the zero ideal and itself. 
\end{definition}

\begin{definition}[Semisimple algebra] 
\label{definition:SemisimplicityOfC002}
If a non-zero algebra is a direct product of simple algebras, then the algebra is called semisimple.  
\end{definition}

A simple algebra is a special case of 
semisimple algebras in the sense that, up to isomorphism,  a simple algebra can be written as a 
direct proproduct of the zero algebra and itself.

It immediately follows that a non-zero algebra $C(P_\mathit{T}) \doteq P_\mathit{T}\; \circ \;C$ is simple if and only if the non-zero idempotent t-scalar $P_\mathit{T}$ is primitive, namely, $P_\mathit{T} \in S^\mathit{pidem} $.
It is  immediately verified that a non-trivial  semisimple
t-algebra $C$ is the direct product of $K$ orthogonal factors 
as follows.
\begin{equation}
C = C(\myQ{1}) \,\times\, \cdots \,\times\,  C(\myQ{K})  \doteq 
\scalebox{1}{$\prod\nolimits_{k = 1}^{K} C(\myQ{k}) $} 
\label{equation:direct-product-via-primitive-idempotent}
\end{equation}
where $C(\myQ{k}) \doteq \myQ{k} \circ C 
\doteq 
\{
\scalebox{0.92}{$\myQ{k} \circ X_\mathit{T}$} \,|\, \, 
\scalebox{0.95}{$ X_\mathit{T} \in C$}
\}
$ for all $\myQ{k} \in S^\mathit{pidem}$.

By definition, the field $\mathbb{C}$ of complex numbers is also  
a one-dimensional algebra.
It is easy to follow that each algebra $C(\myQ{k})$  
is isomorphic to the algebra $\mathbb{C}$  for all $\myQ{k} \in S^{pidem}$.  Hence, the following 
isomorphism holds in the form of a direct product  
\begin{equation}
C \doteq \scalebox{1}{$\raisebox{0em}{$\prod_{k = 1}^{K}$} $} \, C(\myQ{k}) 
\;\cong\;
\scalebox{1}{$\raisebox{0em}{$\prod_{k = 1}^{K}$} $} \,\mathbb{C}
\;\;.
\label{equation:direct-product-equivalence}
\end{equation}

\textbf{Orthogonality series}.\;
Following the vein of equation (\ref{equation:direct-product-via-primitive-idempotent}), 
it shows that the set $S^{pidem} \doteq \{\myQ{1}, \cdots, \myQ{K} \}$ is a generating-set of 
$C$. More precisely, each t-scalar $Y_\mathit{T} \in C$ 
is a linear combination of $\myQ{1}, \cdots, \myQ{K}$  in the following form
\begin{equation}
Y_\mathit{T} \equiv Y_\mathit{T} \circ E_\mathit{T} = 
\scalebox{1}{$\sum\nolimits_{k = 1}^{K}$} Y_{T} \circ \myQ{k}
\equiv
\scalebox{1}{$\sum\nolimits_{k = 1}^{K}$}  \mymu(\scalebox{1}{$\myQ{k}, Y_\mathit{T}$} )   \cdot \myQ{k} 
\label{equation:orthogonality-series}
\end{equation} 
where  
\begin{equation}
\mymu(\myQ{k},   Y_\mathit{T}) \doteq K \cdot \raisebox{0em}{$\langle \myQ{k}, Y_\mathit{T} \rangle$}  
\end{equation}
is the $k$-th complex coordinate of $Y_\mathit{T} \in C$ in terms of $\myQ{k} \in S^\mathit{pidem}$.

It is easy to follow the primitive idempotent t-scalars $\myQ{1},\cdots,\myQ{K}$
are orthogonal basis vectors of the underlying vector space of $C$, and the Gram matrix of 
$\myQ{1},\cdots,\myQ{K}$ is given by 
$G_\mathit{mat} \doteq \big(\langle  \myQ{k}, \myQ{k'}   \rangle \big) =  
K^{-1 } \cdot  I_{mat}$ where $I_\mathit{mat}$ denotes the $K\times K$ identity matrix.

By the nature of equation (\ref{equation:direct-product-equivalence}),  
following equalities hold for all $\lambda \in \mathbb{C}$ and $X_\mathit{T}, Y_\mathit{T} \in C$, 
\begin{equation}
\begin{aligned}
\myconj{X} = \scalebox{0.92}{$\sum\nolimits_{k = 1}^{K}$} 
\overline{
\scalebox{0.92}{$
\mymu(\scalebox{1}{$\myQ{k}, Y_\mathit{T}$} )
$} }
\cdot \myQ{k}  \\
r(X_\mathit{T}) = \scalebox{0.92}{$\sum\nolimits_{k = 1}^{K}$} 
|\scalebox{0.92}{$
\mymu(\scalebox{1}{$\myQ{k}, Y_\mathit{T}$} )
$} | 
\cdot \myQ{k}   \\
\lambda \cdot X_\mathit{T} = \scalebox{0.92}{$\sum\nolimits_{k = 1}^{K}$}    \big(\lambda \cdot  
\scalebox{0.92}{$
\mymu(\scalebox{1}{$\myQ{k}, Y_\mathit{T}$} )
$}
\big) \cdot \myQ{k} \\
X_\mathit{T} + Y_\mathit{T} = \scalebox{0.92}{$\sum\nolimits_{k = 1}^{K}$}  \big(
\scalebox{0.92}{$
\mymu(\scalebox{1}{$\myQ{k}, X_\mathit{T}$} )
$}
+  
\scalebox{0.92}{$
\mymu(\scalebox{1}{$\myQ{k}, Y_\mathit{T}$} )
$}
\big) \cdot \myQ{k} \\
\psi(X_\mathit{T}, Y_\mathit{T}) \doteq \myconj{X} \circ Y_\mathit{T} =  
\scalebox{0.92}{$\sum\nolimits_{k = 1}^{K}$}  
\big(
\overline{\,
\scalebox{0.92}{$
\mymu(\scalebox{1}{$\myQ{k}, X_\mathit{T}$} )
$} }   \cdot  
\scalebox{0.92}{$
\mymu(\scalebox{1}{$\myQ{k}, Y_\mathit{T}$} )
$\,}
\big)
\cdot \myQ{k}
\\
X_\mathit{T} \circ Y_\mathit{T} = \scalebox{0.92}{$\sum\nolimits_{k = 1}^{K}$}  \big(
\scalebox{0.92}{$
\mymu(\scalebox{1}{$\myQ{k}, X_\mathit{T}$} )
$}
\cdot  
\scalebox{0.92}{$
\mymu(\scalebox{1}{$\myQ{k}, Y_\mathit{T}$} )
$}
\big) \cdot \myQ{k} \\
X_\mathit{T} \le Y_\mathit{T} \Leftrightarrow    
\mymu(\scalebox{1}{$\myQ{k}, X_\mathit{T}$} ) \leqslant
\mymu(\scalebox{1}{$\myQ{k}, Y_\mathit{T}$} ), 
\forall k \in [K]\;.
\end{aligned}
\label{equation:many-direct-product-equalities}
\end{equation}

These equalities in equation (\ref{equation:many-direct-product-equalities}) are analogous to those in equation (\ref{equation:direct-product-equalities}).

\section{Generalized Matrices over T-algebra and Beyond} 
\label{section:generalized-matrix-over-talgebra}
\subsection{T-matrix}
With various notions defined on $C$, one can establish algebraic structures over $C$. For 
example, one can define matrices over $C$, which generalizes matrices over $\mathbb{C}$ (i.e., complex 
matrices).  A matrix over $C$, called t-matrix,  is a rectangular array of 
t-scalars arranged in rows and columns. 
T-matrices follow the same algebraic principles of complex matrices and hence are backward-compatible to 
complex matrices 
\cite{liao2020generalized}.

For instance, 
for each t-matrix $X_\mathit{TM} \in C^{M_1\times M_2}$, let $(X_\mathit{TM})_{m_1, m_2}$ be its 
$(m_1, m_2)$-th t-scalar entry
for all $(m_1, m_2) \in [M_1] \times [M_2]$.  
Then, some operations on t-matrices are given as follows.

\textbf{T-matrix addition}.\;
The t-matrix addition 
$X_\mathit{TM} +  Y_\mathit{TM} $ for all $X_\mathit{TM}, Y_\mathit{TM} \in C^{M_1\times M_2}$
is a t-matrix in $C^{M_1\times M_2}$ such that 
\begin{equation}\label{equation:t-matrix-addition}
(X_\mathit{TM} + Y_\mathit{TM})_{m_1, m_2} =
(X_\mathit{TM})_{m_1, m_2} + 
(Y_\mathit{TM})_{m_1, m_2} \in C
\;, \forall (m_1, m_2) \in [M_1] \times [M_2]\;. 
\end{equation}

\textbf{T-scalar multiplication}.\;
The t-scalar multiplication $\lambda_\mathit{T} \circ  X_\mathit{TM} \in C^{M_1\times M_2}$ for all $\lambda_\mathit{T} \in C$,
$X_\mathit{TM} \in C^{M_1\times M_2}$ is a t-matrix in $C^{M_1\times M_2}$ such that
\begin{equation}\label{equation:t-scalar-multiplication}
(\lambda_\mathit{T} \circ X_\mathit{TM})_{m_1, m_2} = \lambda_\mathit{T} \circ (X_\mathit{TM})_{m_1, m_2} 
\in C
\;,
\forall m_1, m_2 \;.
\end{equation}

\textbf{T-matrix multiplication}.\;
The t-matrix multiplication 
$X_\mathit{TM} \circ Y_\mathit{TM}$ for all  
$X_\mathit{TM} \in C^{M_1\times M}$, $Y_\mathit{TM} \in C^{M\times M_2}$ is 
a t-matrix in $C^{M_1\times M_2}$ 
such that
\begin{equation}\label{equation:t-matrix-multiplication}
\raisebox{0.05em}{$(\scalebox{0.92}{$X_\mathit{TM} \circ Y_\mathit{TM}$}  )_{m_1, m_2}$}
= 
\scalebox{1}{$\sum\nolimits_{m = 1}^{M}$}
(X_\mathit{TM})_{m_1, m} \circ 
(Y_\mathit{TM})_{m, m_2}  \in C\,,
\forall m_1, m_2\,.
\end{equation}

\textbf{Scalar multiplication}.\;
The scalar multiplication  $\lambda \cdot  X_\mathit{TM} $ for all $\lambda 
\in \mathbb{C}$,
$X_\mathit{TM} \in C^{M_1\times M_2}$ is a t-matrix in $C^{M_1\times M_2}$ such that
\begin{equation}\label{equation:canonical-scalar-multiplication}
(\lambda \cdot X_\mathit{TM})_{m_1, m_2} = \lambda \cdot (X_\mathit{TM})_{m_1, m_2} 
\in C
\;,
\forall m_1, m_2 \;.
\end{equation}

\newcommand\myconjm[1]{ {#1}^{\raisebox{0.05em}{$*$} }_\mathit{TM} }

\textbf{Conjugate transpose of a t-matrix}.\;
The conjugate transpose 
$\myconjm{X}$ (the original notation in \cite{liao2020generalized} is 
\scalebox{0.92}{$
X_\mathit{TM}^\mathcal{H}
$})
of a t-matrix $X_\mathit{TM} \in C^{M_1\times M_2}$ is a t-matrix in $C^{M_2\times M_1}$ such that 
\begin{equation}\label{equation:t-multiplication-conjugate}
\scalebox{1.1}{$
(\scalebox{0.88}{$\myconjm{X}$})
$}_{m_2,\,m_1} = 
\scalebox{1.1}{$
(\scalebox{0.92}{$X_\mathit{TM}$})
$}_{m_1,\,m_2}^{\raisebox{-0.1em}{$*$}} \in  C \,,\,\, \forall m_1, m_2 \;.
\end{equation}

\textbf{Multiplication of a matrix and a t-scalar}.\;
The multiplication $Y_\mathit{TM} \doteq Y_\mathit{mat} \ltimes X_\mathit{T}$ 
is a t-matrix in $C^{M_1\times M_2}$ for all $Y_\mathit{mat} \in \mathbb{C}^{M_1\times M_2}$ and $X_\mathit{T} \in 
C$ such that the $(m_1, m_2)$-th t-scalar entry of the product 
$Y_\mathit{TM}$ is given by
\begin{equation} 
(Y_\mathit{TM} )_{m_1, m_2} = (Y_\mathit{mat})_{m_1,\,m_2} \cdot X_\mathit{T} \in C\,, \;\forall m_1, m_2
\label{equation:matrix-multiplication-TM}
\end{equation}
where $(Y_\mathit{mat})_{m_1,\,m_2} $ denotes the $(m_1, m_2)$-th complex entry of the matrix $Y_\mathit{mat}$.

Equation (\ref{equation:matrix-multiplication-TM}) extends equation
(\ref{equation:canonical-scalar-multiplication}) since
the former reduces to the latter when $M_1 = M_2 = 1$.
One has the notion of t-vector via
the notion of t-matrix ------ 
a t-matrix 
$X_\mathit{TM} \in C^{M_1\times M_2}$ reduces to a t-vector in $X_\mathit{TV} \in C^{M_1\times 1} \equiv 
C^{M_1}$ when $M_2 = 1$.

All t-matrices of the same size form a module over the ring $C$ 
by equations (\ref{equation:t-matrix-addition}) and (\ref{equation:t-scalar-multiplication}). They also 
form a vector space over the field $\mathbb{C}$ by equations (\ref{equation:t-matrix-addition}) 
and (\ref{equation:canonical-scalar-multiplication}).

\subsection{Semisimplicity and {Decomposability} of A Module of T-matrices}
The semisimplicity of a module over an algebra can be defined analogously to the semisimplicity of an algebra. 
Here, we rephrase the definition of module semisimplicity given in \cite{erdmann2018algebras} as follows.

\begin{definition}[Simple module] 
A non-zero module is called simple or irreducible if it has no submodule besides the zero submodule and itself. 
\end{definition}

\begin{definition}[Semisimple module] 
A non-zero module 
is called semisimple if 
the module is a direct product of simple modules.
\end{definition}

Every simple module is a special case of semisimple module since, up to isomorphism, a simple module can 
always be written as a direct product of the zero submodule and itself.

A non-trivial $C$-module of can be written as a direct product via a finite number of primitive idempotent elements. More concretely, let $G \equiv C^{M_1\times M_{2} }$ be a non-trivial module over $C \equiv \mathbb{C}^{I_1\times \cdots 
\times I_N}$ and 
$\myQ{1},\cdots,\myQ{K}$ be the primitive idempotent 
elements of $C$. Following the direct product 
in equation (\ref{equation:direct-product-via-primitive-idempotent}), 
the module $G$ is semisimple and hence a direct product 
by $K$ simple submodules 
as follows. 
\begin{equation}
G = G(\myQ{1}) \,\times \cdots \times\, G(\myQ{K}) \,\doteq\,  
\scalebox{1}{$\prod\nolimits_{k = 1}^{K} G(\myQ{k}) $} 
\label{equation:module-decomposition}
\end{equation}
where $G(\myQ{k}) \doteq \myQ{k} \circ G 
\doteq 
\{
\myQ{k} \circ X_\mathit{TM}
\;|\;\, 
X_\mathit{TM} \in G
\}
$ for all $k \in [K]$.

These submodules are simple such that none of them can be written 
as a direct product of two non-trivial proper submodules.  
Furthermore, 
the submodules 
$\myQ{1},\cdots,\myQ{K}$
are orthogonal to each other.

One can also extend the notion of inner 
product $\psi$ over $C$ to over $G \equiv C^{M_1\times M_2}$. 
The extended inner product 
is a $C$-sesquilinear form   $\psi: G \times G \rightarrow C$ defined as follows.
\begin{equation}
\psi(
\scalebox{0.92}{$X_\mathit{TM}, Y_\mathit{TM}$} 
) \doteq 
\scalebox{1}{$\sum\nolimits_{m_1,m_2} $}\,
(X_\mathit{TM})_{m_1,m_2}^{\scalebox{1}{$*$} }
\circ 
(Y_\mathit{TM})_{m_1,m_2} \in C \;,\forall X_\mathit{TM}, Y_\mathit{TM} \in G\;.
\end{equation}

\textbf{Orthogonality on $C$-module}.\;
Two t-matrices $X_\mathit{TM}, Y_\mathit{TM} $ are said orthogonal on $G$ iff
their inner product is equal to $Z_\mathit{T}$, namely,  
$\psi(X_\mathit{TM}, Y_\mathit{TM}) = Z_\mathit{T}$.  
It follows that any two submodules  
$G(\myQ{k})$ and 
$G(\myQ{k'})$ with $k \neq k'$ in equation (\ref{equation:module-decomposition}), 
are orthogonal
in the following sense    
\begin{equation}
\psi(
\scalebox{1}{$X_\mathit{TM}, \,Y_\mathit{TM}$} 
) = Z_\mathit{T}\;, \;\forall 
X_\mathit{TM} \in G(\myQ{k}),  
Y_\mathit{TM} \in G(\myQ{k'}) \;\;.
\end{equation}

\textbf{Orthogonality series}.\;
In the light of equation 
(\ref{equation:orthogonality-series}), for all t-matrix 
$Y_\mathit{TM} \in G \equiv  C^{M_1\times M_2}$,
the t-matrix $Y_\mathit{TM}$ can be  written as a unique series as follows
\begin{equation}\label{equation:TM-orthogonality-series}
Y_\mathit{TM} 
= \scalebox{1.1}{$\sum\nolimits_{k = 1}^{K}$}\, Y_{\mathit{TM},\,k} 
\doteq \scalebox{1.1}{$\sum\nolimits_{k = 1}^{K}$}\,  
Y_\mathit{mat, k} \ltimes \myQ{k} 
\doteq 
\scalebox{1.1}{$\sum\nolimits_{k = 1}^{K}$}\,  
f_{k}(Y_\mathit{TM})
 \ltimes \myQ{k}
\;.
\end{equation}
where $f_{k}(Y_\mathit{TM}) \doteq Y_{\mathit{mat}, k} \in \mathbb{C}^{M_1\times M_2}$ is the $k$-th 
matrix constituent, such that 
the $(m_1, m_2)$-th complex entry of $Y_{\mathit{mat}, k}$ is given by 
\begin{equation}
(Y_{\mathit{mat}, k})_{m_1, m_2} 
= \mymu(\scalebox{0.92}{$\myQ{k}$}, \scalebox{0.92}{$(Y_\mathit{TM})_{m_1, m_2}$}   )
\doteq K \cdot \langle  
\scalebox{0.92}{$\myQ{k}$}, \,\scalebox{0.92}{$(Y_\mathit{TM})_{m_1, m_2}$}  
\rangle \in \mathbb{C}
\end{equation}
for all $
(k, m_1, m_2) \in \scalebox{0.92}{$[K] \times [M_1] \times [M_2]$} 
$.

It shows that 
the t-matrices $Y_\mathit{TM, k} \doteq Y_{\mathit{mat}, k} \times \myQ{k} $ and 
$Y_{\mathit{TM}, k'} \doteq   Y_{\mathit{mat}, k'} \times \myQ{k'} $ are orthogonal on the module $G$
and its underlying vector space for all $k \neq k'$, more symbolically, 
the following equalities hold for all $k \neq k'$,
\begin{equation}
\begin{aligned}
\langle Y_{\mathit{TM}, k},\, Y_{\mathit{TM}, k'}  \rangle = 0 \\
\psi(Y_{\mathit{TM}, k},\, Y_{\mathit{TM}, k'}) = Z_\mathit{T}  \\
Y_{\mathit{TM}, k} \cap Y_{\mathit{TM}, k'} = \{Z_\mathit{T}\} 
\end{aligned} \;\;\;.
\end{equation}

Equation (\ref{equation:TM-orthogonality-series}) is called the orthogonality series of $Y_\mathit{TM}$.

For  
$X_\mathit{TM} = \scalebox{1}{$\sum\nolimits_{k = 1}^{K}$}\,X_\mathit{mat, k} \ltimes \myQ{k} \in G$, 
$Y_\mathit{TM} = \scalebox{1}{$\sum\nolimits_{k = 1}^{K}$}\,Y_\mathit{mat, k} \ltimes \myQ{k} \in G$, 
$\lambda_\mathit{T} \in C$ and $\alpha \in \mathbb{C}$,  
it is easy to verify the following equalities hold 
\begin{equation}
\begin{aligned}
X_\mathit{TM}^{\scalebox{1}{$*$}} = \scalebox{1}{$\sum\nolimits_{k = 1}^{K}$}\, 
X_{\mathit{mat}, k}^{\scalebox{1}{$*$}}  \ltimes \myQ{k}  \\
\alpha \cdot X_\mathit{T} = 
\scalebox{1}{$\sum\nolimits_{k = 1}^{K}$}\, 
(\alpha   \cdot  X_\mathit{mat, k}) 
\ltimes \myQ{k} \\
\lambda_\mathit{T} \circ X_\mathit{T} = 
\scalebox{1}{$\sum\nolimits_{k = 1}^{K}$}\, 
\big(\scalebox{0.92}{$\mymu(\myQ{k}, \lambda_\mathit{T})$}   \cdot  X_\mathit{mat, k} \big) 
\ltimes \myQ{k}  \\
X_\mathit{TM} + Y_\mathit{TM} =  \scalebox{1}{$\sum\nolimits_{k = 1}^{K}$}\, 
(X_\mathit{mat, k} + Y_\mathit{mat, k}) \ltimes \myQ{k}  \\
\psi(X_\mathit{TM}, Y_\mathit{TM}) = \scalebox{1}{$\sum\nolimits_{k = 1}^{K}$}\, 
\langle X_{\mathit{mat}, k}, Y_{\mathit{mat}, k}  \rangle   \ltimes \myQ{k} \\
X_\mathit{TM} \circ Y_\mathit{TM} = \scalebox{1}{$\sum\nolimits_{k = 1}^{K}$}\, 
(X_{\mathit{mat}, k} \cdot Y_{\mathit{mat}, k}  ) \ltimes \myQ{k}   \\
\end{aligned}\;\;\;.
\label{equation:direct-product-for-module}
\end{equation}

It shows that these equalities in equation (\ref{equation:direct-product-for-module}) are analogous to 
those in equations (\ref{equation:direct-product-equalities})
and (\ref{equation:many-direct-product-equalities}).

Equation (\ref{equation:direct-product-for-module}) shows that an operation or a notion on a t-matrix 
$X_\mathit{TM} = \scalebox{1}{$\sum\nolimits_{k = 1}^{K}$}\,X_\mathit{mat, k} \ltimes \myQ{k}$
is reducible to its canonical counterparts $X_{\mathit{mat}, k}$ for all $k \in [K]$. 
This helps define more notions of t-matrices.

\textbf{Singular value decomposition of a t-matrix}.\;
One can give the notion of singular value decomposition on a t-matrix.  
Given a t-matrix $Y_\mathit{TM} = \scalebox{1}{$\sum\nolimits_{k = 1}^{K}$}\, 
Y_{\mathit{mat}, k} \ltimes \myQ{k} \in C^{M_1\times M_2} $, let the singular value decomposition (SVD) of
the $k$-th matrix constituent
$Y_{\mathit{mat}, k}$ of the t-matrix $Y_\mathit{TM}$ 
be 
\begin{equation}
Y_{\mathit{mat}, k} = 
U_{\mathit{mat}, k} \cdot S_{\mathit{mat}, k} \cdot 
V_{\mathit{mat}, k}^{\scalebox{1}{$*$}} \,\,,\,  \forall k \in [K]
\end{equation} 
where 
$M \doteq \operatorname{min}(M_1, M_2)$,
$U_{\mathit{mat}, k} \in \mathbb{C}^{M_1\times M}$, 
$V_{\mathit{mat}, k} \in \mathbb{C}^{M_2 \times M}$  
and 
$
S_{\mathit{mat}, k} \doteq \operatorname{diag}(
\scalebox{0.9}{$\lambda_{1}^{(k)},\cdots,\lambda_{M}^{(k)}$}) 
$
such that 
$U_{\mathit{mat}, k}^{\scalebox{1}{$*$}} \cdot U_{\mathit{mat}, k} = 
V_{\mathit{mat}, k}^{\scalebox{1}{$*$}} \cdot V_{\mathit{mat}, k} =
I_\mathit{mat}$ and $\lambda_{1}^{(k)}  
\;\scalebox{0.9}{$\geqslant$}\;   \cdots \;\scalebox{0.9}{$\geqslant$}\; \lambda_{M}^{(k)} \;\scalebox{0.9}{$\geqslant$}\; 0 $ for all $k \in [K]$.

Then, the following t-matrices 
are given
\begin{equation}
\begin{aligned}
U_\mathit{TM} \doteq \scalebox{0.92}{$\sum\nolimits_{k = 1}^{K}$}\, U_{\mathit{mat}, k} \ltimes \myQ{k}
\in C^{M_1\times M}  \\ 
V_\mathit{TM} \doteq \scalebox{0.92}{$\sum\nolimits_{k = 1}^{K}$}\, V_{\mathit{mat}, k} \ltimes \myQ{k}
\in C^{M_2\times M}   \\
S_\mathit{TM} \doteq \scalebox{0.92}{$\sum\nolimits_{k = 1}^{K}$}\, S_{\mathit{mat}, k} \ltimes \myQ{k}
\doteq \operatorname{diag}(\lambda_{\mathit{T},\,1},\cdots, \lambda_{\mathit{T},\,M})
\in C^{\,M\times M}
\end{aligned} \;\;\;\;.
\end{equation}

It follows that the following equality holds 
\begin{equation}\label{equation:TSVD}
Y_\mathit{TM} = U_\mathit{TM} \circ S_\mathit{TM} \circ V_\mathit{TM}^{\scalebox{1}{$*$}} 
\end{equation}
where 
\begin{equation}
\begin{matrix}
U_\mathit{TM}^{\scalebox{1}{$*$}} \circ U_\mathit{TM} =  
V_\mathit{TM}^{\scalebox{1}{$*$}} \circ V_\mathit{TM} = I_\mathit{TM} \doteq 
\operatorname{diag}(E_\mathit{T}, \cdots,E_\mathit{T}) \;,  \vspace{0.5em}\\
\lambda_{\mathit{T},\,1} \geq \cdots \geq \lambda_{\mathit{T},\,M} \geq Z_\mathit{T} \;\;.
\end{matrix}
\end{equation}

Equation (\ref{equation:TSVD}), called TSVD (Tensorial Singular Value Decomposition), 
is a higher-order generalization of a matrix's singular value decomposition. When $I_1 = \cdots = I_N = 1$, 
equation (\ref{equation:TSVD}) reduces to the canonical singular 
value decomposition of a matrix. 
When $I_1 > 1$ and $I_2 = \cdots = I_N = 1$, equation (\ref{equation:TSVD}) 
reduces to Kilmer's version of SVD called t-SVD in  \cite{Kilmer2011Factorization-TProduct001-0006}.

\textbf{Pseudoinverse of a t-matrix}.\; 
The notion of pseudoinverse can also be defined via the orthogonality series in equation 
(\ref{equation:TM-orthogonality-series}).

The pseudoinverse of a t-matrix  
$Y_\mathit{TM} = \scalebox{0.92}{$\sum\nolimits_{k = 1}^{K}$}\,Y_\mathit{mat, k} \ltimes \myQ{k} \in C^{M_1\times M_2}$ is given by 
\begin{equation}\label{equation:moore-penrose-inverse}
Y_\mathit{TM}^{\scalebox{1}{$+$}} = 
\scalebox{1}{$\sum\nolimits_{k = 1}^{K}$}\, Y_{\mathit{mat}, k}^{\scalebox{1}{$+$}} \ltimes \myQ{k}
\end{equation}
where $Y_{\mathit{mat}, k}^{\scalebox{1}{$+$}}$ denotes the canonical Moore-Penrose inverse of  $Y_{\mathit{mat}, k}$ for all $k \in [K]$.\footnote{The pseudoinverse can be of any type. However, in this article, we only discuss the Moore-Penrose inverse and its generalization over $C$.}

\textbf{Rank of a t-matrix}.\;
The notions of TSVD and pseudoinverse of a t-matrix helps define the rank of a t-matrix. 
Let $Y_\mathit{TM} = U_\mathit{TM} \circ S_\mathit{TM} \circ V_\mathit{TM} \in C^{M_1\times M_2}$ be the 
compact TSVD of $Y_\mathit{TM}$,  
namely $S_\mathit{TM}$ is a diagonal t-matrix in $C^{M\times M}$ and $M \doteq \min(M_1, M_2)$, 
the rank of $Y_\mathit{TM}$ is given by 
\begin{equation}
\operatorname{rank}(Y_\mathit{TM}) \doteq 
\operatorname{trace} (S_\mathit{TM} \circ S_\mathit{TM}^{\scalebox{1}{$+$}}) 
\label{equation:t-matrix-rank}
\end{equation}
where $\operatorname{trace}(\cdot)$ returns the sum of diagonal t-scalar entries of a square t-matrix.

It show that the inequality $Z_\mathit{T} \le  \operatorname{rank}(Y_\mathit{TM}) \le M\cdot E_\mathit{T}$ holds for all $Y_\mathit{TM} \in C^{M_1\times M2}$.
On the other hand, for all t-matrix 
$Y_\mathit{TM} \doteq \scalebox{1}{$\sum\nolimits_{k = 1}^{K}$}\,Y_\mathit{mat, k} \ltimes \myQ{k} \in C^{M_1\times M_2}$, one has the following equality  
\begin{equation}
\operatorname{rank}(Y_\mathit{TM}) = \scalebox{0.92}{$\sum\nolimits_{k = 1}^{K}$}\,
\operatorname{rank}(Y_\mathit{mat, k}) \cdot \myQ{k}\;\;.
\end{equation}

When $I_1 =  \cdots  I_N = 1$,  equation (\ref{equation:t-matrix-rank}) reduces to the rank of a 
$M_1\times M_2$ complex matrix. When $M_1 = M_2 = 1$, equation (\ref{equation:t-matrix-rank}) reduces to 
the rank of a t-scalar, i.e., an idempotent t-scalar in $S^\mathit{idem}$. Namely, 
there are $2^{K}$ different possible values as the rank of a t-scalar
where $K \doteq \myK$.

If and only if a t-scalar $X_\mathit{T} \in C$ is multiplicatively 
invertible, the rank of $X_\mathit{T}$ is equal to $E_\mathit{T}$, i.e., the greatest element of the poset 
$S^\mathit{idem}$. Hence, the invertible t-scalar $X_\mathit{T}$ is also equivalently called of full rank.

Furthermore, the rank of a t-matrix $Y_\mathit{TM} \in C^{M_1\times M_2}$
is reducible to the ranks of its singular values. Namely, let the generalized singular values of 
$Y_\mathit{TM}$ be $\lambda_{\mathit{T}, 1},\cdots,\lambda_{\mathit{T}, M}$ where $M \doteq \min(M_1, M_2)$, the following equality 
holds  
\begin{equation}
\operatorname{rank}(Y_\mathit{TM}) = 
\scalebox{1}{$\sum\nolimits_{k = 1}^{M}$}\,
\operatorname{rank}(\lambda_{\mathit{T}, k})\;\;\;.
\label{equation:sum-of-rank-tscalars}
\end{equation}

\textbf{Generalized Frobenius norm}.\;
Another fundamental notion for general data analytics is the norm of a t-matrix. Similar to its canonical 
counterparts, a t-matrix can have different types of norms. Among them is the so-called Frobenius norm 
------ given a t-matrix $X_\mathit{TM} \in C^{M_1\times M_2}$, its Frobenius norm, 
analogous to its canonical counterpart, is defined
as follows
\begin{equation} 
r(X_\mathit{TM})_{F} = \sqrt[2]{\psi(X_\mathit{TM}, X_\mathit{TM})} 
\;\in S^\mathit{nonneg}\; \;.
\label{equation:generalized-norm}
\end{equation}

When $M_2 = 1$, $X_\mathit{TM} \in C^{M_1\times M_2}$ reduces to a t-vector denoted by $X_\mathit{TV}$. 
The norm 
in equation (\ref{equation:generalized-norm}) in this context  
is denoted by $r(X_\mathit{TV})_2 \in S^\mathit{nonneg}$.

\textbf{Generalized distance}.\;
Equation (\ref{equation:generalized-norm}) help extend the notion of distance over $C$. 
For each pair of  t-matrices $X_\mathit{TM}, Y_\mathit{TM} \in G \equiv C^{M_1\times M_2}$, a generalized 
distance between 
$X_\mathit{TM}$ and $Y_\mathit{TM}$ is defined by a nonnegative t-scalar as follows
\begin{equation}
\scalebox{1}{$d\,$}(X_\mathit{TM}, Y_\mathit{TM}) = r(X_\mathit{TM} - Y_\mathit{TM})_{F} \in 
S^\mathit{nonneg} \;\;.
\label{equation:generalized-distance}
\end{equation}

It is easy to prove that the following 
conditions, analogous to the axioms of canonical distance, hold
for all $X_\mathit{TM}, Y_\mathit{TM}, 
X'_\mathit{TM} \in G$, 
\begin{equation}
\begin{aligned}
\scalebox{1}{$d\,$}(X_\mathit{TM}, Y_\mathit{TM}) = \scalebox{1}{$d\,$}(Y_\mathit{TM}, X_\mathit{TM}) \\
\scalebox{1}{$d\,$}(X_\mathit{TM}, Y_\mathit{TM}) = Z_\mathit{T} \Leftrightarrow X_\mathit{TM} = Y_\mathit{TM} \\
\scalebox{1}{$d\,$}(X_\mathit{TM}, X'_\mathit{TM}) \le \scalebox{1}{$d\,$}(X_\mathit{TM}, Y_\mathit{TM}) + 
\scalebox{1}{$d\,$}(Y_\mathit{TM}, X'_\mathit{TM})
\end{aligned}\;\;\;.
\end{equation}

\begin{remarks}[\textbf{Generalized metric space}]
The pair $(G, d)$ generalizes the notion of metric space, and the 
function $d$, called t-metric, sends 
any pair of elements in $G$ to an element of the poset $S^\mathit{nonneg}$.
\end{remarks}

\subsection{Optimization}

One can minimize a nonnegative-tscalar-valued function characterized by 
the generalized notions over $C$.\footnote{A maximization problem
can be reformulated as a minimization problem. Hence, we only discuss the minimization problems over $C$ in this subsection.}

Given 
a function $f: A \rightarrow  S^\mathit{nonneg}$,  
if the range 
$f(A) $, a poset, has the least element under partial order ``$\le$'', namely 
$\inf f(A) \doteq  
\min\nolimits_{\,\alpha \,\in A} f(\alpha) \in f(A)
$, 
and $f$ is an injection,  
there is a unique element $\hat{\alpha} \in A$ satisfying $f(\hat{\alpha}) 
= \inf f(A) $. In this scenario, the notation $\hat{\alpha} \doteq \operatorname{argmin}\nolimits_{\,\alpha \in A} 
f(\alpha)$ 
is used.

The function $f: \alpha \mapsto f(\alpha) \in S^\mathit{nonneg}$ can also be written in the following form 
\begin{equation}
f(\alpha) = \scalebox{1}{$\sum\nolimits_{k = 1}^{K}$}\, 
\mymu(\scalebox{0.9}{$\myQ{k}$}, 
\scalebox{0.9}{$f(\alpha)$}
)
\cdot \myQ{k} 
\end{equation}
where $
\mymu(\scalebox{0.85}{$\myQ{k}$}, \scalebox{0.9}{$f(\alpha)$} ) 
\doteq K \cdot \langle \myQ{k}, f(\alpha) \,\rangle 
\geqslant 0 
$ for all $k \in [K]$.

If the range $f(A)$ has the least element, the unique element 
$\inf f(A) \doteq \min\nolimits_{\alpha\,\in A}f(\alpha)$ can be written as follows 
\begin{equation} 
\inf f(A) \doteq 
\min\nolimits_{\alpha\,\in A}f(\alpha) = \scalebox{1}{$\sum\nolimits_{k = 1}^{K}$}\, 
\Big(  
\min\nolimits_{\alpha \,\in A} \mymu(\scalebox{0.9}{$\myQ{k}$}, \scalebox{0.9}{$f(\alpha)$} 
)
\Big) 
\cdot \myQ{k} 
 \;\;\;.
\end{equation}

In other words, one can seek a minimizer to $f(\alpha)$ via investigating the minimizers 
to a finite number of canonical subfunctions 
$\mymu(\scalebox{0.9}{$\myQ{1}$}, \scalebox{0.9}{$f(\alpha)$} ),\cdots,
\mymu(\scalebox{0.9}{$\myQ{K}$}, \scalebox{0.9}{$f(\alpha)$} )
$.

\section{Applications of T-matrices in General Visual Information Analysis}
\label{section:Applications-of-T-matrices}

\subsection{Tensorial Representation of T-matrices }
A t-matrix is an order-two array of t-scalars.
On the other hand, each t-scalar entry of a t-matrix can be represented by a fixed-sized order-$N$ complex 
array. 
Thus, a convenient numerical representation of a t-matrix is an order-$(N+2)$ array of 
complex numbers.

There are many equivalent tensorial representations for t-matrices. 
Following the convention in \cite{liao2020generalized}, we represent a t-matrix 
$X_\mathit{TM} \in C^{M_1\times M_2}$ by a complex array in $\mathbb{C}^{I_1 \times \cdots \times I_N\times 
M_1\times M_2}$.

T-matrices can characterize much information. 
For example, an RGB image in the form of an order-three real array can be characterized by a t-matrix of order-one t-scalars. A color video in the form of an order-four real array can be represented by a t-matrix of order-two t-scalars. Even a monochrome image in the form of an order-two array can be represented by a t-matrix of order-zero t-scalars.

However, we contend, to have an effective t-matrix representation of high-order data,  the complex entries of 
a t-scalar need to be correlated. Otherwise, the {\color{black}t-scalar} multiplication based on circular convolution is 
pointless.

Hence, a convenient application arena of the t-matrix paradigm is for analyzing visual 
information including but not limited to images, videos, and sequential data such as time series, where for a 
raw data sample, there are always spatially-correlated neighborhoods available for exploitation.

\subsection{T-matrix Representation of Legacy Visual Information}
\label{section:t-matrix-representation}
To reuse the legacy data representation and, on the other hand, exploit the potential of the 
t-matrix paradigm, one needs a consistent neighborhood strategy for t-matricizing visual information.

\begin{figure}[h]
\centering
\includegraphics[width=\textwidth]{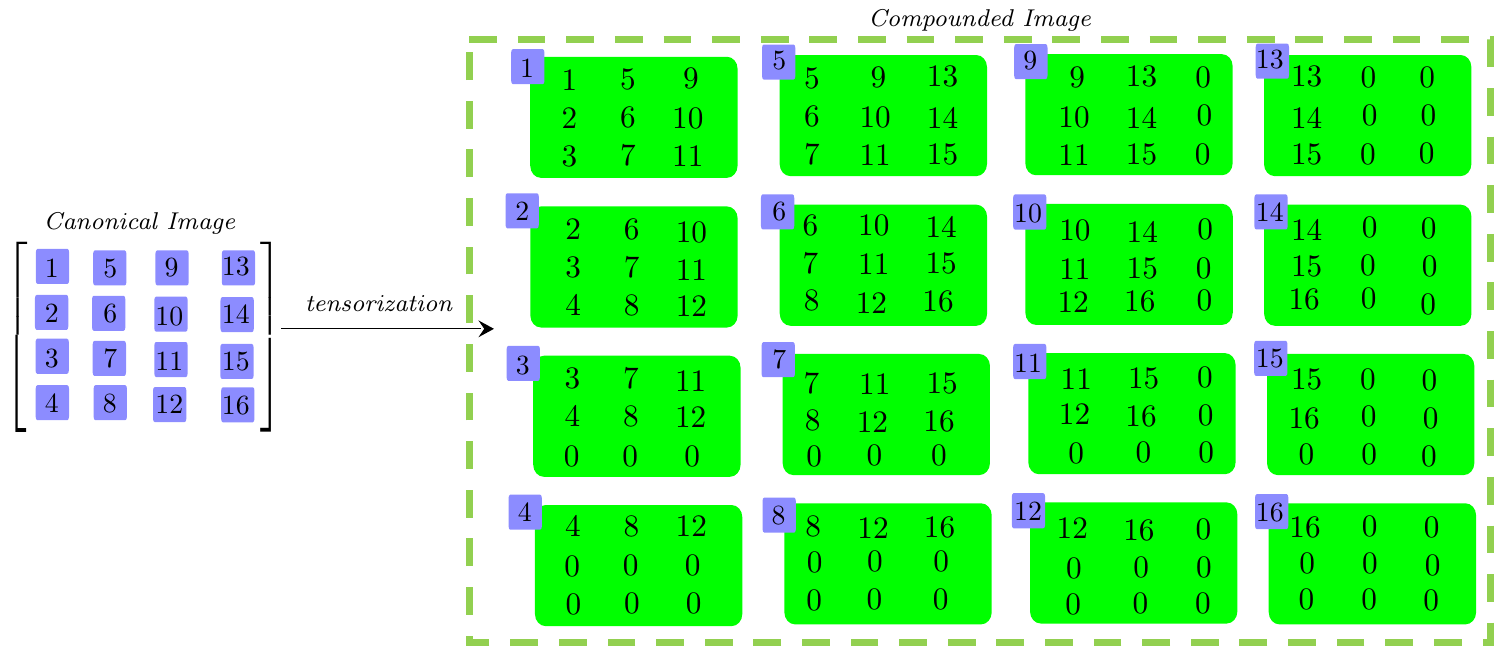} 
\caption{A t-matricization strategy of a 2D legacy image  
using $3\times 3$ ``inception'' neighborhoods ------ 
from an input order-two array in $\scalebox{0.9}{$\mathbb{C}^{4\times 4}$} $ (i.e., canonical matrix) to an 
order-four array in  $\scalebox{0.9}{$C^{4\times 4} \equiv \mathbb{C}^{3\times 3\times 4\times 4} $} $
(i.e., a t-matrix).}
\label{figure:tensorization001}
\end{figure}

Figure \ref{figure:tensorization001} demonstrates a ``$3\times 3$-neighborhood'' strategy for t-matricizing a 
small grey image of $16$ pixels, i.e., $16$ real numbers, in the form of a $4 \times 4$ real array.  
The t-matricization yields a t-matrix in $C^{4\times 4}$, i.e., an order-four array in 
$\mathbb{C}^{3\times 3\times 4 \times 4}$.

Each small blue box corresponds to a scalar entry of the $4\times 4$ input matrix in Figure \ref{figure:tensorization001}.
There are many distinct $3\times 3$ neighborhoods available for each scalar of the 
input matrix. For example, one can either have a ``central'' neighborhood set $\{1, 2, 3, 5, 6, 7, 9, 10, 
11\}$ for 
the scalar $6$ or alternatively, a so-called ``inception'' neighborhood set $\{6, 7, 8, 10, 11, 
12, 14, 15, 16\}$.

Figure \ref{figure:tensorization001} adopts the so-called ``inception'' neighborhood 
of each scalar for t-matricizing the input $4\times 4$ matrix. If a scalar is located at the image border, {\color{black}one can} 
pad with $0$ when necessary to have a $3\times 3$ neighborhood.  
Each ``inception'' neighborhood in the form of a $3\times 3$ green box is highlighted by the corresponding 
scalar represented by a small blue box at the top-left corner of each green box.

The feasibility of the ``neighborhood'' strategy, as demonstrated in Figure \ref{figure:tensorization001}, is under the condition that input data is spatially-constrained. Hence, the t-matrix paradigm with the demonstrated neighborhood strategy is suitable to analyzing images or other visual 
information. \footnote{If the given matrix is not spatially-constrained, the spatially-correlated neighborhood  strategy makes no sense. It is possible to 
analyze spatially-correlated data with the t-matrix paradigm. However, one needs a different t-matricization strategy to extend legacy data to higher-order. }

\begin{figure}[tbh]
\includegraphics[width=\textwidth]{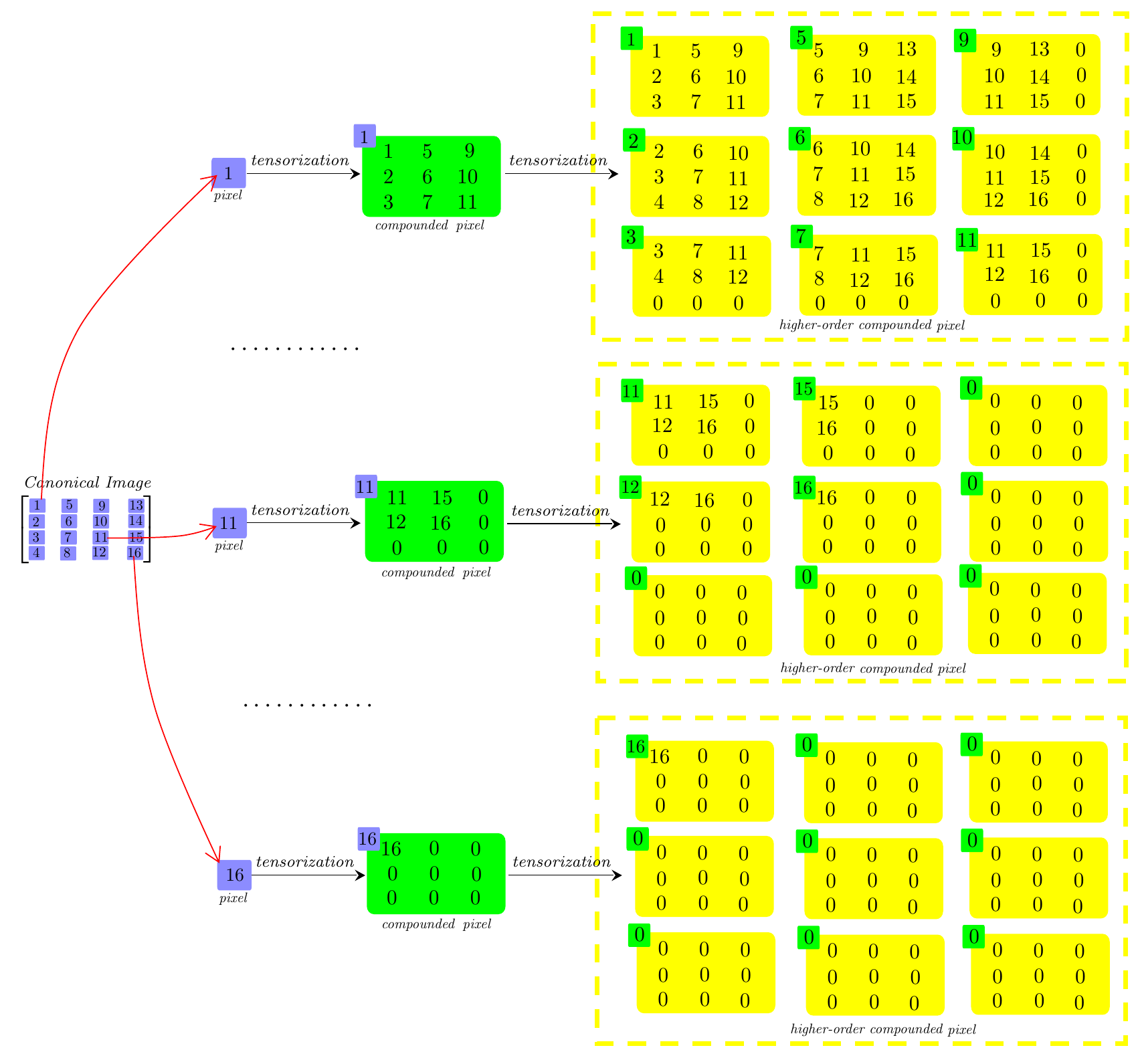}
\caption{Reuse the neighborhood strategy to extend to a legacy grey image 
(i.e., an array in $\mathbb{R}^{4\times 4}$) to 
a low-order compounded image (i.e., an array in $\mathbb{R}^{3\times 3\times 4\times 4}$) and, then 
to a higher-order compounded image (i.e., an array in $\mathbb{R}^{3\times 3\times 3\times 3\times 4\times 4}$). 
}
\label{figure:higher-order-tensorization}
\end{figure}

The neighborhood strategy can be reused to extend input data to higher-orders. 
Figure \ref{figure:higher-order-tensorization} demonstrates how to extend 
a canonical grey image (i.e., an order-two array in $\mathbb{R}^{4\times 4}$ ) to 
a low-order compounded image (i.e., an order-four array in $\mathbb{R}^{3\times 3\times 4\times 14}$ )
and then to 
a higher-order compounded image (i.e., an order-six array in $\mathbb{R}^{3\times 3\times 3\times 
3\times 4\times 4}$).

The spatial neighborhood strategy enables general visual information  
analysis with the higher-order t-matrix paradigm.

\subsection{Generalized Low-rank Approximation over $C$}
\label{equation:higher-order-generalization-low-rank}
With the t-matrix paradigm, many applications can be straightforwardly 
generalized. To this end, we discuss a high-order generalization of the Eckart-Young-Mirsky 
theorem, named after the authors of the theorem \cite{eckart1936approximation, mirsky1960symmetric}.

Low-rank approximation plays an important role in modeling many applications in machine learning 
and data analytics. The problem is to find a low-rank optimal approximation to a given matrix. 
Specifically, given a matrix $X_\mathit{mat} \in \mathbb{C}^{M_1\times M_2}$, one seeks an 
approximation matrix $\hat{X}_\mathit{mat}$ to $X_\mathit{TM}$, satisfying 
\begin{equation}
\begin{aligned}
\|X_\mathit{mat} - \hat{X}_\mathit{mat}  \|_{F} = 
\min\nolimits_{\,\operatorname{rank}(Y_\mathit{mat}) 
\,\leqslant\, r } \|X_\mathit{mat} - Y_\mathit{mat}\|_F 
\\
\;\text{subject to}\;
\operatorname{rank}(\hat{X}_\mathit{mat}) \leqslant r \leqslant \operatorname{rank}(X_\mathit{mat}) 
\;\;. 
\end{aligned}
\label{equation:low-rank-approximation} 
\end{equation}

\newcommand\myconjcanonicalm[1]{ {#1}^{\raisebox{0.05em}{$*$} }_\mathit{mat} }

The optimal approximation is given by the Eckart-Young-Mirsky  theorem via 
the SVD (Singular Value Decomposition) of $X_\mathit{mat}$. More specifically, 
given a complex matrix $X_\mathit{mat} \in \mathbb{C}^{M_1\times M_2}$, let 
$M \doteq \min(M_1, M_2)$ and $X_\mathit{mat} = U_\mathit{mat} \cdot S_\mathit{mat} \cdot 
\myconjcanonicalm{V}$
be the compact SVD of $X_\mathit{mat}$
such that 
$S_{mat} \doteq 
\operatorname{diag}(\lambda_1,\cdots,\lambda_M)$ where
$\lambda_1  \geqslant \cdots \lambda_M \geqslant 0$.

The Eckart-Young-Mirsky theorem gives the optimal approximation $\hat{X}_\mathit{mat}$ via the SVD (Singular Value Decomposition) of $X_\mathit{mat}$ as follows. 
\begin{equation}
\hat{X}_\mathit{mat} = U_\mathit{mat} \cdot \hat{S}_\mathit{mat} \cdot \myconjcanonicalm{V}
\label{equation:canonical-low-rank-approximation}
\end{equation}
where $\hat{S}_\mathit{mat} \doteq \operatorname{diag}(\lambda_{\,1}, \cdots,\lambda_{\,r}, 
\underset{(M-r)\;\text{copies}}{\underbrace{0,\,\,\cdots\,\,,0}})$.

With the canonical paradigm where 
a nonnegative integer defines rank, 
a solution to a higher-order generalization of equation (\ref{equation:low-rank-approximation}) is NP-hard
\cite{hillar2013most,vannieuwenhoven2014generic,boralevi2017orthogonal}, 
and as a consequence, 
``naive approach to this problem is doomed to failure''
\cite{de2008tensor}.

However, with the t-matrix paradigm or Kilmer's t-product model, a higher-order 
generalization of equation (\ref{equation:low-rank-approximation}) with an analytical solution analogous to 
equation (\ref{equation:canonical-low-rank-approximation}) 
is straightforward.  
The higher-order generalization with the t-matrix paradigm is as follows.

For a t-matrix $X_\mathit{TM} \in C^{M_1\times M_2} \equiv \mathbb{C}^{I_1\times \cdots \times I_N \times 
M_1\times M_2}$, the 
generalized optimization over $C$ is to find a low-rank t-matrix $\hat{X}_\mathit{TM} \in C^{M_1\times M_2}$ such 
that 
\begin{equation}
\begin{aligned}
r(X_\mathit{TM} - \hat{X}_\mathit{TM} )_F = \min\nolimits_{\,\operatorname{rank}(Y_\mathit{TM}) \,\le\, H_{T} } r(X_\mathit{TM} - Y_\mathit{TM} )_{F} \\
\text{subject to}\; \operatorname{rank}(\hat{X}_\mathit{TM}) \,\le\, H_{\,T} \,\le\, 
\operatorname{rank}(X_\mathit{TM})
\;\;.
\end{aligned}
\label{equation:generalized-low-rank-approximation}
\end{equation}

Let $H_{T} = \scalebox{1}{$\sum\nolimits_{k = 1}^{K}$}\,r_k \cdot \myQ{k} \ge Z_{T}$
be the orthogonality series of the nonnegative t-scalar $H_{T} $. 
Without loss of generality, let's assume that $0 \leqslant r_1,\cdots,r_K \leqslant M$ are all nonnegative integers where $M \doteq \min(M_1, M_2)$. In other words, 
there must exist a t-matrix in $C^{M_1\times M_2}$ whose rank is equal to 
$H_{T}$.

The nonnegative t-scalar $H_{T}$ can be uniquely represented 
by the sum of $M$ idempotent t-scalars as follows 
\begin{equation}
H_{T} =  \scalebox{1}{$\sum\nolimits_{m = 1}^{M}$}\, 
\delta_{T,\,m}
\label{equation:sum-of-Rank}
\end{equation}
where $\delta_{T,\,1},\cdots,\delta_{T,\,M} \in S^\mathit{idem}$ and  
$\delta_{T,\,1} \ge\, \cdots \ge\, \delta_{T,\,M} $ .

The sum in equation (\ref{equation:sum-of-Rank}) is unique, and the $m$-th idempotent t-scalar $\delta_{T,\,m} $ is given as follows. 
\begin{equation}
\delta_{T,\,m} = \scalebox{1}{$\sum\nolimits_{k = 1}^{K}$}\, 
\raisebox{-0.105em}{$\mathbf{1}$}_{\,m \,\leqslant\, r_k} \cdot \myQ{k} \;\;
\label{equation:delta}
\end{equation}
where $\raisebox{-0.105em}{$\mathbf{1}$}_{\,m \leqslant r_k}$ is the indicator function, which returns $1$ when 
$m \leqslant r_{k}$ and otherwise, returns $0$.

Let $X_\mathit{TM} = U_\mathit{TM} \,\circ\, S_\mathit{TM} \,\circ\, V_\mathit{TM}^{\scalebox{1}{$*$}}$ be the 
compact TSVD of $X_\mathit{TM}$ where $S_\mathit{TM} = \operatorname{diag}(\lambda_{T, 1},\cdots,\lambda_{T, M} )$ 
and $\lambda_{T,\,1} \ge \cdots \ge \lambda_{T,\,M} \ge Z_{T}$.
The analytical solution of equation (\ref{equation:generalized-low-rank-approximation}) is given by 
\begin{equation}
\hat{X}_\mathit{TM} = U_\mathit{TM} \circ \hat{S}_\mathit{TM} \circ V_\mathit{TM}^{\scalebox{1}{$*$}}
\label{equation:generalized-solution}
\end{equation}
where $\hat{S}_\mathit{TM} \doteq 
\operatorname{diag}(\lambda'_{T,\,1},\cdots,\lambda'_{T,\,M}   )
$ and $\lambda'_{T,\,k} \doteq \lambda_{T,\,k} \circ \delta_{T,\,k}$ for all $k \in [M]$.

A simplified version of the above generalization is given when $H_{T}$ is in the form $H_{T} = r \cdot E_{T} \equiv 
\scalebox{0.92}{$\sum\nolimits_{k = 1}^{K}$}\, r \cdot \myQ{k}$ where $r \leqslant M$. Under this condition, 
equation (\ref{equation:delta}) reduces to 
\begin{equation}
\left\{
\begin{aligned}
&\delta_{T,\,1} = \cdots =\delta_{T,\,r} = E_{T} \\
&\delta_{T,\,r+1} = \cdots =\delta_{T,\,M} = Z_{T} 
\end{aligned}
\right. \;\;\;\;.
\end{equation}

Namely, the t-matrix $\hat{S}_\mathit{TM}$
in equation (\ref{equation:generalized-solution}) reduces to 
$\scalebox{0.90}{
$
\hat{S}_\mathit{TM} \doteq \operatorname{diag}(\lambda_{T,\,1},\cdots,\lambda_{T,\,r}, \, 
\underset{(M-r)\; \text{copies} }{\underbrace{Z_{T},\cdots,Z_{T}}}    )
$}$.

In this case, 
the approximation $\hat{X}_\mathit{TM} \doteq U_\mathit{TM} \circ \hat{S}_\mathit{TM} \circ  
V_\mathit{TM}^{\scalebox{1}{$*$}} \in 
C^{M_1\times M_2} \equiv \mathbb{C}^{I_1\times \cdots \times I_N\times M_1\times M_2}$  
is analogous to the canonical approximation $\hat{X}_\mathit{mat}$ in 
equation (\ref{equation:canonical-low-rank-approximation})
and is called the ``truncated''  TSVD approximation.

When $I_1 = \cdots = I_N = 1$, 
equation (\ref{equation:generalized-low-rank-approximation}) reduces to equation 
(\ref{equation:low-rank-approximation}), and the generalized solution given by equation (\ref{equation:generalized-solution}) reduces to the canonical solution given by equation (\ref{equation:canonical-low-rank-approximation}).

In other words, 
equation (\ref{equation:generalized-low-rank-approximation}) is a 
straightforward generalization of 
the analytical solution given by the Eckart-Young-Mirsky theorem.

\subsection{Generalized Least-squares over $C$}

By the semisimplicity of $C$, many 
canonical applications can be
generalized using the t-matrix paradigm. These generalizations are completely compatible with their canonical 
counterparts.

For example, in \cite{liao2020generalized, Liao2017Hyperspectral}, Liao and Maybank et al. generalize the
algorithms of HOSVD (Higher-Order Singular Value Decomposition), 
PCA  (Principal Component Analysis), 2DPCA (Two Dimensional PCA), and Grassmannian Component Analysis over $C$, 
used for analyzing or classifying visual data.

To show the general principles and particularly the backward-compatibility of the t-matrix paradigm, we discuss the 
backward-compatible generalization of the well-known least-squares, which belongs to a special class of 
convex optimization. The principles demonstrated in the following discussion apply to generalize other canonical 
optimization applications, even those not convex.

The least-squares optimization over $C$ is backward-compatible with the canonical least-squares optimization over $\mathbb{C}$ and 
is formulated 
as follows  
\begin{equation}
r(W_\mathit{TM} \circ  \hat{\beta}_\mathit{TV} - A_\mathit{TV})_2 = \min\nolimits_{\beta_\mathit{TV} \in C^{M}} \,r(W_\mathit{TM} \circ \beta_\mathit{TV} - A_\mathit{TV})_2 \;\;.
\label{equation:generalized-least-squares}
\end{equation}

In the above equation, $r(\cdot)$ is the generalized norm defined by equation (\ref{equation:generalized-norm}).  
The t-matrix $W_\mathit{TM} \in C^{D \times M}$ ($D \geqslant M$) and
the t-vector $A_\mathit{TV} \in C^{D}$ are given in advance. 
The t-vector 
$\beta_\mathit{TV} \in C^{M}$ is optimizable, and the t-vector $\hat{\beta}_{TV} \in C^{M}$ is the optimal 
solution of $\beta_\mathit{TV}$.

The generalized least-squares has a geometric interpretation. Precisely, the column t-vectors of the t-matrix 
$W_\mathit{TM} \in C^{D\times M}$ form a generating set 
spanning a submodule $\mathcal{M} \subseteq C^{D}$ 
with a generalized dimension $\operatorname{dim}(\scalebox{0.92}{$\mathcal{M}$}) \doteq   
\operatorname{rank}(\scalebox{0.92}{$W_\mathit{TM}$}) $.

The projection $A'_\mathit{TV} $ 
of the t-vector $A_\mathit{TV} $ on the submodule $\mathcal{M}$ 
is given by 
\begin{equation}
\begin{aligned}
{A}'_\mathit{TV} &\doteq W_\mathit{TM} \circ \hat{\beta}_\mathit{TV} \\
&= W_\mathit{TM} \circ 
\big(\myconjm{W} \circ W_\mathit{TM} \big)^{+} \circ \myconjm{W} \circ A_\mathit{TV} 
\\
& \doteq P_\mathit{TM} \circ A_\mathit{TV} \in \mathcal{M} 
\end{aligned}
\end{equation}
where 
$P_{TM} \doteq   
W_\mathit{TM} \circ 
\big(\myconjm{W} \circ W_\mathit{TM} \big)^{+} \circ \myconjm{W} 
\in C^{D\times D}
$ is called the projection t-matrix 
for the submodule  $\mathcal{M}$.

The t-matrix $P_\mathit{TM}$ is idempotent in the sense  that 
\begin{equation}
P_\mathit{TM} \circ P_\mathit{TM} = P_\mathit{TM}\;\;\;. 
\end{equation}

Also, the following equalities hold for all t-matrix $W_\mathit{TM} \in C^{D\times M}$, 
\begin{equation}
\begin{matrix}
\operatorname{rank}(W_\mathit{TM}) \equiv \operatorname{rank}(P_\mathit{TM}) \vspace{0.5em} \\
W_\mathit{TM}^{\scalebox{1}{$+$}} \equiv (\myconjm{W} \circ  W_\mathit{TM})^{+} \circ \myconjm{W} \\
\end{matrix} \;\;\;.
\end{equation}

The generalized least-squares 
is equivalently 
defined by the generalized distance between 
$A_\mathit{TV}$ and the submodule $\mathcal{M}$, i.e., the generalized distance between 
$A_\mathit{TV} \in C^{D}$ and $A'_\mathit{TV} \in \mathcal{M}$. More precisely,  
\begin{equation}
\begin{aligned}
r(A'_\mathit{TV} - A_\mathit{TV} )_2 \equiv
d{\hspace{0.05em}}(A'_\mathit{TV}, \,A_\mathit{TV} ) \ge Z_{T} \;\;\;.
\end{aligned}
\end{equation}

Note that the generalized least-squares 
$r(W_\mathit{TM} \circ \hat{\beta}_\mathit{TV} - A_\mathit{TV} )$
is unique for all t-matrix $W_\mathit{TM} \in C^{D\times M}$ and all t-vector
$A_\mathit{TV} \in C^{D}$. 
However, the t-vector $\hat{\beta}_\mathit{TV} \in C^{M}$ is not necessarily unique and is given as follows.
\begin{equation}
\hat{\beta}_\mathit{TV} \in \left\{W_\mathit{TM}^{\scalebox{1}{$+$} } 
\circ A_\mathit{TV} + (I_\mathit{TM} - W_\mathit{TM}^{\scalebox{1}{$+$}} \circ W_\mathit{TM}) \circ 
{\xi}_\mathit{TV} \,\,|\,\, {\xi}_\mathit{TV} \in C^{M}   
\right\}
\end{equation}
where $I_\mathit{TM} \doteq \operatorname{diag}(\,\underset{M\;\text{copies}}{\underbrace{E_{T},\cdots,E_{T}}} \,)$ is the identity t-matrix.

\newcommand\myinvm[1]{ {#1}^{\raisebox{0.07em}{\scalebox{0.8}{$\,-1$} } }_\mathit{TM} }

The t-vector $\hat{\beta}_\mathit{TV} \in C^{M}$ is unique if and only if the column t-vectors of $W_\mathit{TM} \in C^{D\times 
M}$, where $D \geqslant M$, are independent over $C$, or in other words, the t-matrix $W_\mathit{TM}$ is of full rank.

The condition that the column t-vectors of $W_\mathit{TM}$ are independent over $C$ 
is equivalent to one
of the following conditions 
\begin{equation} 
\begin{aligned}
\text{(\romannumeral1)}&\;\; \operatorname{rank}(W_\mathit{TM}) = M \cdot E_{T}  \;\;\text{where}\;\;  M \doteq \min(M_1, M_2)  \\
\text{(\romannumeral2)}&\;\; W_\mathit{TM}^{\scalebox{1}{$+$}} \circ W_\mathit{TM} = I_\mathit{TM} \doteq \operatorname{diag}(\,\underset{M\;\text{copies}}{\underbrace{E_{T},\cdots,E_{T}}} \,) 
\end{aligned} \;\;\;.
\label{equation:equalities-full-rank}
\end{equation}

When the column t-vectors of $W_\mathit{TM}$ are independent over $C$, the minimizer 
$\hat{\beta}_\mathit{TV} \in C^{M}$ is unique and 
is given by 
\begin{equation}
\begin{aligned}
\hat{\beta}_\mathit{TV} &\doteq \mathop{\operatorname{argmin}}\nolimits_{\beta_\mathit{TV} \in C^{M}} 
\;r(W_\mathit{TM} \circ \beta_\mathit{TV} - A_\mathit{TV})_2 \\ 
&=  W_\mathit{TM}^{\scalebox{1}{$+$}} \circ A_\mathit{TV} \\
& \equiv 
\big(\myconjm{W} \circ W_\mathit{TM} \big)^{+} \circ \myconjm{W} \circ A_\mathit{TV} \\
& \equiv
\big(\myconjm{W} \circ W_\mathit{TM} \big)^{-1} \circ \myconjm{W} \circ A_\mathit{TV}
\;\;
\end{aligned}
\end{equation}
where $\big(\scalebox{0.9}{$\myconjm{W} \circ W_\mathit{TM}$} \big)^{-1}$, 
called the inverse of the t-matrix $\scalebox{0.9}{$\myconjm{W} \circ W_\mathit{TM}$} \in C^{M\times M}$, 
is a special case of 
the pseudoinverse  
$\big(\scalebox{0.9}{$\myconjm{W} \circ W_\mathit{TM}$} \big)^{+}$ when the t-matrix 
$\big(\scalebox{0.9}{$\myconjm{W} \circ W_\mathit{TM}$} \big) \in C^{M\times M}$ is of full rank.

When $I_1 = \cdots = I_N  = 1$, the generalized least-squares over $C$ reduces to the canonical least-squares over 
$\mathbb{C}$.

\subsection{Generalized Principal Component Analysis over $C$}
Using generalized least-squares over $C$, one can generalize the well-known method of 
PCA (Principal Component Analysis). 
The generalized PCA is called TPCA (Tensorial PCA).

Precisely,   
given $N$ t-vectors $X_{\mathit{TV},\,1},\cdots, 
X_{\mathit{TV},\,N} \in C^{D}$,  
the generalized component analysis of 
these t-vectors is to find a finite number of   
principal components 
$U_{\mathit{TV},\,1},\cdots, 
U_{\mathit{TV},\,Q} \in C^{D}$
such that $\myconjm{U} \circ U_\mathit{TM} = I_\mathit{TM} \doteq \operatorname{diag}(E_{T},\cdots,E_{T}) \in C^{\,Q\times 
Q} $ where the $k$-th column t-vector of $U_\mathit{TM} \in C^{D\times Q}$, denoted by 
$(U_\mathit{TM})_{:,\,k}$, is the 
principal component $U_{\mathit{TV},\,k}$, namely, 
$(U_\mathit{TM})_{:,\,k} \doteq U_{\mathit{TV},\,k}, \forall k \in [Q] $.

\newcommand\myconjv[1]{ {#1}^{\raisebox{0.05em}{$*$} }_\mathit{TV} }

The principal components $U_{\mathit{TV},\,1},\cdots,U_{\mathit{TV},\,Q} \in C^{D}$ capture the dominant information of the t-vectors 
$X_{\mathit{TV},\,1}, \cdots, X_{\mathit{TV},\,N} \in C^{D}$ such that the first principal component 
$U_{\mathit{TV},\,1}$ 
is given by 
\begin{equation}
\begin{aligned}
U_{\mathit{TV},\,1} 
&\doteq 
\mathop{\operatorname{argmax}}\nolimits_{\,r(Y_\mathit{TV})_2 = E_{T}}
\;\Big\{
\scalebox{1}{$
\sum\nolimits_{k=1}^{N}
$}
\,\big|
\scalebox{0.9}{$\myconjv{Y} \circ (X_{\mathit{TV},\,k} - \bar{X}_\mathit{TV})$}  
\big|^{2}
\Big\} \\
&\equiv
\mathop{\operatorname{argmax}}\nolimits_{\,r(Y_\mathit{TV})_2 = E_{T}}
\;r^{2}(
\scalebox{0.9}{$\myconjv{Y} \circ W_{\mathit{TM}}$}  
)_F  \\
& \equiv  
\mathop{\operatorname{argmax}}\nolimits_{\,r(Y_\mathit{TV})_2 = E_{T}}
\;\scalebox{0.95}{$
\myconjv{Y} \circ W_{\mathit{TM}} \circ \myconjm{W} \circ Y_\mathit{TV}
$}  \\
\end{aligned}
\label{equation:maxization}
\end{equation}
where 
\begin{equation}
\bar{X}_\mathit{TV} \doteq \scalebox{0.92}{$(1/N) \cdot \sum\nolimits_{k = 1}^{N}$}\, X_{\mathit{TV},\,k}
\end{equation} 
is the mean of the t-vectors $X_{\mathit{TV},\,1},\cdots,X_{\mathit{TV},\,N}$
and 
$W_\mathit{TM} \in C^{D\times N}$ denotes the t-matrix whose $k$-th column $(W_\mathit{TM})_{:,\,k} \in C^{D}$
is given by
\begin{equation}
(W_\mathit{TM})_{:,\,k} \doteq X_{\mathit{TV},\,k} - \bar{X}_\mathit{TV} \;\;,\;\forall k \in [N]\;.
\end{equation}

Note that the condition $r(Y_\mathit{TV})_2 = E_{T}$ is equivalent to $\myconjv{Y} \circ Y_\mathit{TV} = 
E_{T}$. Equation (\ref{equation:maxization}) is to find the stationary point(s) 
of the following formulation with a generalized Lagrange multiplier $\lambda_{\,T} \in C$, 
\begin{equation}
\mathcal{L}(Y_\mathit{TV}) = \myconjv{Y} \circ W_\mathit{TM} \circ \myconjm{W} \circ Y_\mathit{TV} -\lambda_{\,T} \circ (\myconjv{Y} \circ Y_\mathit{TV} - E_{T}  ) \;\;.
\label{equation:generalized-lagarange-multilier}
\end{equation}

The stationary point(s) of equation  
(\ref{equation:generalized-lagarange-multilier}) can be determined
by its derivative over $C$, equal to $Z_{T}$.\footnote{A rigorous investigation of differentiation over $C$ is beyond the scope of this article.}  
\begin{equation}
\begin{aligned}
&\hspace{4.6em}\frac{\partial\, \mathcal{L}(Y_\mathit{TV}) }{\partial\, Y_\mathit{TV} } = 
\scalebox{0.92}{$2$} \cdot 
\Big(
\myconjv{Y} \circ W_\mathit{TM} \circ \myconjm{W} -  \lambda_{\,T} \circ \myconjv{Y}
\Big)
 = Z_{T} \\ 
&\Rightarrow\;\; W_\mathit{TM} \circ \myconjm{W} \circ Y_\mathit{TV} = Y_\mathit{TV} \circ \lambda_{\,T}  
\;\Rightarrow\; \myconjv{Y} \circ W_\mathit{TM} \circ \myconjm{W} \circ Y_\mathit{TV}  = \lambda_{\,T} \;\;.
\end{aligned}
\label{equation:generalized-differentation-over-talgebra}
\end{equation}

It shows that $U_{\mathit{TV},\,1}$ is the generalized eigenvector in $C^{D} \equiv \mathbb{C}^{I_1\times \cdots \times 
I_N\times D} $ with the generalized maximum eigenvalue 
$\lambda_{\,T} \in S^\mathit{nonneg}$ of the Hermitian t-matrix $
{\color{black}\scalebox{1}{$W_\mathit{TM} \circ \myconjm{W}$}}
\in C^{D\times D} \equiv \mathbb{C}^{I_1\times \cdots \times I_N\times D\times D}
$.

The t-vector
$U_{\mathit{TV},\,1}$ is also the 
dominant singular t-vector with the generalized
maximum singular value, i.e., a nonnegative t-scalar, of the t-matrix $W_\mathit{TM} \in C^{D\times N} 
\equiv \mathbb{C}^{I_1\times \cdots \times I_N\times D\times N}$.

We provide an unrigorous interpretation of the derivative as in 
equation (\ref{equation:generalized-differentation-over-talgebra}) ------ 
given a mapping $\mathcal{L}:  Y_\mathit{TV} \mapsto \mathcal{L}(Y_\mathit{TV})$, the dependable 
$\mathcal{L}(Y_\mathit{TV}) \in C$ 
can be written as follows
\begin{equation} 
\left\{
\begin{aligned}
&\mathcal{L}(Y_\mathit{TV}) = 
\scalebox{1}{$\sum\nolimits_{k = 1}^{K}$} \mathcal{L}_{k}(Y_{\mathit{vec},\,k})  \cdot  \myQ{k} \, \\
&Y_\mathit{TV} =  \scalebox{1}{$\sum\nolimits_{k = 1}^{K}$} Y_{\mathit{{vec}},\,k} \ltimes \myQ{k} 
\end{aligned}
\right. 
\end{equation}
where $\mathcal{L}_{k}$ is the $k$-th sub-mapping of $\mathcal{L}$ for each $k \in [K]$.

Then, the derivative of $\mathcal{L}(Y_\mathit{TV})$ with respect to 
$Y_\mathit{TV}$ is given by  
\begin{equation}
\frac{\partial \mathcal{L}(Y_\mathit{TV}) }{\partial Y_\mathit{TV}} \doteq 
\scalebox{1}{$\sum\nolimits_{k = 1}^{K}$} 
\scalebox{1}{$
\frac{\partial 
\scalebox{0.95}{$\mathcal{L}_{k}(Y_{\mathit{vec},\,k})$} 
}{\partial 
\scalebox{0.95}{$Y_{\mathit{vec},\,k}$}
}   \cdot  \myQ{k} 
$} \in C
\label{equation:generalized-differentiation-of-partial-derivative}
\end{equation}
where the derivative of the left side of the equation is the generalized derivative on $C$, and the derivatives of the right 
side denote the canonical derivatives on complex numbers.

\newcommand\myconjcanonicalv[1]{ {#1}^{\raisebox{0.05em}{$*$} }_\mathit{vec} }

In a simple case as equation (\ref{equation:generalized-differentation-over-talgebra}), 
the sub-mappings $\mathcal{L}_{1},\cdots,\mathcal{L}_{K}$ are the identical 
real-valued vector functions 
given by 
\begin{equation}
\mathcal{L}_{k}:   Y_\mathit{vec} \,\mapsto\, \myconjcanonicalv{Y} \cdot W_\mathit{mat} \cdot  
\myconjcanonicalm{W}  \cdot 
Y_\mathit{vec} -\lambda \cdot (\myconjcanonicalv{Y} \cdot Y_\mathit{vec} - 1  )\;,\; 
\forall k \in [K]\;\;.
\end{equation}

It shows that equation (\ref{equation:generalized-differentation-over-talgebra}) is a result given by equation (\ref{equation:generalized-differentiation-of-partial-derivative}).

When the first $q$ principal t-vectors $U_{\mathit{TV},\,1},\cdots,U_{\mathit{TV},\,q}$ are obtained, one uses the following equation to project the t-matrix {\color{black}$W_\mathit{TM}$} 
on the orthogonal complement submodule of the submodule spanned by the principal t-vectors 
$U_{\mathit{TV},\,1},\cdots,U_{\mathit{TV},\,q} \in C^{D}$. More 
precisely,  
\begin{equation}
\begin{aligned}
W_{\mathit{TM},\,(q+1)} 
=   
\Big(
I_\mathit{TM} - \scalebox{1}{$\sum\nolimits_{i = 1}^{q}$} \,U_{\mathit{TV},\,i} \circ  {\myconjv{U}}_{,\,i}
\Big) \circ 
W_\mathit{TM}  \in C^{D \times N} \;\;.
\end{aligned} 
\label{equation:tmatrix-residual}
\end{equation}

Let the t-matrices $P_{\mathit{TM},\,q}, \,P_{\mathit{TM},\,q}^{\raisebox{0.1em}{\scalebox{0.8}{$\perp$}}} \in C^{D\times D}$ be given 
by 
\begin{equation}
\left\{
\begin{aligned}
&P_{\mathit{TM},\,q} \doteq \scalebox{1}{$\sum\nolimits_{i = 1}^{q}$} \,
\scalebox{0.95}{$U_{\mathit{TV},\,i}$}  
\circ  {\myconjv{U}}_{,\,i} \in C^{D\times D} \\
&P_{\mathit{TM},\,q}^{\raisebox{0.1em}{\scalebox{0.8}{$\perp$}}} \doteq I_\mathit{TM} - P_{\mathit{TM},\,q}\, \in C^{D\times D}
\end{aligned}
\right. \;\;\;.
\end{equation}

Then, the following equalities hold for all $q \in [Q]$, 
\begin{equation}
\begin{aligned}
P_{\mathit{TM},\,q} \circ P_{\mathit{TM},\,q} = P_{\mathit{TM},\,q} = P_{\mathit{TM},\,q}^{\raisebox{0.1em}{\scalebox{0.8}{$*$}}} \\
P_{\mathit{TM},\,q}^{\raisebox{0.1em}{\scalebox{0.8}{$\perp$}}} \circ 
P_{\mathit{TM},\,q}^{\raisebox{0.1em}{\scalebox{0.8}{$\perp$}}} = P_{\mathit{TM},\,q}^{\raisebox{0.1em}{\scalebox{0.8}{$\perp$}}} = 
\big(P_{\mathit{TM},\,q}^{\raisebox{0.1em}{\scalebox{0.8}{$\perp$}}} \big)^{*} 
 \\ 
\operatorname{rank}(P_{\mathit{TM},\,q}^{\raisebox{0.1em}{\scalebox{0.8}{$\perp$}}}) + \operatorname{rank}(P_{\mathit{TM},\,q})  = D \cdot E_{T}  \\
P_{\mathit{TM},\,q} \circ P_{\mathit{TM},\,q}^{\raisebox{0.1em}{\scalebox{0.8}{$\perp$}}} 
= P_{\mathit{TM},\,q}^{\raisebox{0.1em}{\scalebox{0.8}{$\perp$}}} \circ P_{\mathit{TM},\,q} = Z_{T} 
\end{aligned} \;\;\;\;.
\end{equation}

When the t-matrix $W_{\mathit{TM},\,(q+1)} \in C^{D\times N}$ is obtained as in equation (\ref{equation:tmatrix-residual}), the $(q+1)$-th 
principal t-vector 
$U_{\mathit{TV},\,(q+1)} \in C^{D} $
is given by
\begin{equation}
\begin{aligned}
U_{\mathit{TV},\,(q+1)} 
&\doteq 
\mathop{\operatorname{argmax}}\nolimits_{\,r(Y_\mathit{TV})_2 = E_{T}}
\;r^{2}(
\scalebox{0.95}{$\myconjv{Y} \raisebox{0.05em}{$\,\circ\, W_{\mathit{TM},\,(q+1)}$}$}    \,  
)_F  \\
&\equiv \mathop{\operatorname{argmax}}\nolimits_{\,r(Y_\mathit{TV})_2 = E_{T}}  
\;\scalebox{0.95}{$
\myconjv{Y} \circ W_{\mathit{TM},\,(q+1)} \circ {\myconjm{W}}_{,\,(q+1)} \circ Y_\mathit{TV}
$}\;,\, \forall q \in [Q] \;.
\end{aligned}
\label{equation:other-TPCA-component}
\end{equation}

The t-vector 
$U_{\mathit{TV},\,(q+1)} \in C^{D}$ is the $(q+1)$-th dominant generalized eigenvector of
$\scalebox{1}{$W_{TM} \circ \myconjm{W}$} \in C^{D\times D}$, also the $(q+1)$-th dominant singular t-vector of $W_\mathit{TM} \in C^{D\times N}$. The maximum t-scalar 
$\max\nolimits_{r(Y_\mathit{TV})_2 = E_{T} } \,r(\scalebox{1}{$\myconjv{Y} \raisebox{0.05em}{$\,\circ\, W_{\mathit{TM},\,(q+1)}$}$}  \,)_F \in S^\mathit{nonneg}$ is 
the $(q+1)$-th dominant singular value, i.e., a t-scalar, of $W_\mathit{TM} \in C^{D\times M}$. 

Note that 
the t-vectors $(X_{\mathit{TV},\,1} - \bar{X}_\mathit{TV})  ,\cdots,
(X_{\mathit{TV},\,N} - \bar{X}_\mathit{TV})  $ are not independent on the module $C^{D}$. 
This leads to 
\begin{equation}
\begin{aligned}
\operatorname{rank}(W_\mathit{TM}) \le Q_{m}  \cdot E_{T}\;\;\text{where}\;\;  Q_{m} \doteq \min(D, N - 1)\;\;.
\end{aligned}
\end{equation}

Let the compact TSVD of the t-matrix $W_\mathit{TM} \in C^{D\times N}$ 
be $W_\mathit{TM} = U_\mathit{TM} \circ S_\mathit{TM} \circ \myconjm{V}$ 
such that 
$U_\mathit{TM} \in C^{D\times Q_m}$, 
$V_\mathit{TM} \in C^{N\times Q_m}$, 
$\myconjm{U} \circ U_\mathit{TM}  = 
\myconjm{V} \circ V_\mathit{TM}  =
I_\mathit{TM} \in C^{\,Q_m \times Q_m}$ 
and 
$S_\mathit{TM} \doteq \operatorname{diag}(\lambda_{T,\,1}, \cdots,  \lambda_{T,\,Q_m})$
where $\lambda_{T,\,1} \ge \cdots \ge \lambda_{T,\,Q_m} \ge Z_{T}$.

If and only if $\operatorname{rank}(W_\mathit{TM}) = Q_m \cdot E_{T}$, 
the generalized singular values $\lambda_{T,\,1},\cdots,\lambda_{T,\,Q_m}$ are all multiplicatively invertible, and all $Q_m$ column t-vectors of $U_\mathit{TM}$ are the principal t-vectors given by equations
(\ref{equation:maxization}) and  (\ref{equation:other-TPCA-component}).

We call the above-mentioned generalized principal component analysis TPCA (Tensorial PCA). TPCA is backward-compatible with its canonical counterpart PCA. Analogous to its canonical counterpart PCA, TPCA applies to reduce the dimension of data.

More precisely, $N$ t-vectors $X_{\mathit{TV},\,1},\cdots,X_{\mathit{TV},\,N} \in C^{D}$ 
are given in advance and let the nonnegative t-scalar $H_{T}$ be a generalized dimension
subject to 
\begin{equation}
H_{T} \le \operatorname{rank}(W_\mathit{TM}) 
\le Q_m \cdot E_{T} \;.
\end{equation}

Without loss of generality,   
the generalized dimension $H_{T} \ge Z_{T}$ is uniquely represented in the form of equation (\ref{equation:sum-of-Rank}), 
namely, 
\begin{equation}
H_{T} = \scalebox{0.92}{$\sum\nolimits_{q = 1}^{Q_m}$}\, \delta_{T,\,q} 
\end{equation}  
where $\delta_{T,\,q} \in S^\mathit{idem}$ for each $q \in [Q_m]$ 
and 
$\delta_{T,\,1} \ge \cdots \ge \delta_{T,\,Q_m} \ge Z_{T} $.

\newcommand\myYtvector{Y_\mathit{TV}^{ \raisebox{0.1em}{\scalebox{0.75}{$\mathit{raw}$}} } }
\newcommand\myYtvectorTPCA{Y_\mathit{TV}^{ \raisebox{0.1em}{\scalebox{0.6}{$\mathit{TPCA}$}} } }
\newcommand\myYprojection{Y_\mathit{TV}^{ \raisebox{0.1em}{\scalebox{0.75}{$\mathit{proj}$}} } }
\newcommand\myYtvectorConstruction{Y_\mathit{TV}^{ \raisebox{0.1em}{\scalebox{0.75}{$\mathit{recon}$}} } }

A query sample $\myYtvector \in C^{D}$ is then reduced to a t-vector 
$\myYtvectorTPCA \in C^{\,Q_m}$
as follows 
\begin{equation} 
\myYtvectorTPCA \doteq \myconjm{\hat{U}} \circ
(\myYtvector - \bar{X}_\mathit{TV})  
\end{equation}
where 
\begin{equation}
\hat{U}_\mathit{TM} \doteq U_\mathit{TM} \circ 
\operatorname{diag}(\delta_{T,\,1},\cdots,\delta_{T,\,Q_m}) \in C^{D\times Q_m}\;\;\;.
\end{equation}

The projection $\myYprojection$ of the t-vector 
$(\scalebox{0.92}{$\myYtvector - \bar{X}_\mathit{TV}$} )  \in C^{D}$ on the submodule $\mathcal{M}$
spanned by the columns of the t-matrix $\hat{U}_\mathit{TM} \in C^{D\times Q_m}$
is given by the following generalized least-squares problem  
\begin{equation} 
\begin{aligned}
\myYprojection 
&\doteq \mathop{\operatorname{argmin}}\nolimits_{Y_{\mathit{TV}} \,\in \mathcal{M}} 
\;r(\myYtvector - \bar{X}_\mathit{TV} - Y_\mathit{TV} )_2    \\
&= U_\mathit{TM} \circ \myYtvectorTPCA \\ 
&\equiv 
\hat{U}_\mathit{TM} \circ \myYtvectorTPCA \\
&\equiv
\hat{U}_\mathit{TM} \circ \myconjm{\hat{U}} 
\circ (\myYtvector - \bar{X}_\mathit{TV}) \in C^{D} \;\;.
\end{aligned}
\end{equation}

It follows that the t-matrix $P_\mathit{TM} \doteq \hat{U}_\mathit{TM} \circ \myconjm{\hat{U}} \in 
C^{D\times D} $ is idempotent, hermitian, and low-rank over $C$, namely $P_\mathit{TM} \circ P_\mathit{TM} = 
P_\mathit{TM} = \myconjcanonicalm{P} $ and $\operatorname{rank}(P_\mathit{TM}) = H_{T}$.

The reconstruction $\myYtvectorConstruction$ of the query t-vector $\myYtvector$ is 
then given by 
\begin{equation}
\begin{aligned}
\myYtvectorConstruction &= \myYprojection  + \bar{X}_\mathit{TV}  
= P_\mathit{TM}  \circ \myYtvector 
+ (I_\mathit{TM} - P_\mathit{TM}) \circ \bar{X}_\mathit{TV} \;\;\;.
\end{aligned}
\label{equation:TPCA-reconstruction}
\end{equation}

When the given generalized dimension is in the form $H_{T} = Q \cdot E_{T}$, where $Q$ is a positive integer, 
the t-matrix $\hat{U}_\mathit{TM}$  reduces to 
\begin{equation}
\hat{U}_\mathit{TM} \doteq U_\mathit{TM} \circ \operatorname{diag}(
\underset{Q \text{\,\,copies}}{\underbrace{E_{T},\cdots,E_{T}}},     
\underset{(Q_m - Q) \text{\,\,copies}}{\underbrace{Z_{T},\cdots,Z_{T}}} 
) \in C^{D\times Q_m}
\end{equation} 
or, equivalently, 
\begin{equation}
\hat{U}_\mathit{TM} = (U_{TM})_{:, 1:\,Q} \;\in\; C^{D\times Q}
\end{equation}
where
$\hat{U}_\mathit{TM}$ is the sub-structure containing the first $Q$ columns of $U_\mathit{TM} \in C^{D\times 
Q_m}$.

In this situation, the last $(Q_m - Q)$ t-scalar entries of  
$\myYtvectorTPCA \in C^{\,Q_m}$ are discarded.  
When $I_1 = \cdots = I_N = 1$, TPCA reduces to its canonical counterpart PCA.

\section{Experimental Verifications}
\label{section:Experimental-Verifications}
In this section, we demonstrate the t-matrix paradigm for general visual information analytics. We give some experiment results via t-matrices with quantitative comparison to their canonical counterparts.

\subsection{Generalized Low-rank Approxiamtion}
In the first experiment of low-rank approximation, we compare the approximation results of SVD and TSVD 
on the publicly available images.

\subsubsection{``Baboon'' image}
The RGB image used in the first experiment in the $512\times 512\times 3$ is the ``baboon'' image.

\begin{figure}
\begin{minipage}[h] {\textwidth}
\centering
\includegraphics[width=0.95\textwidth]{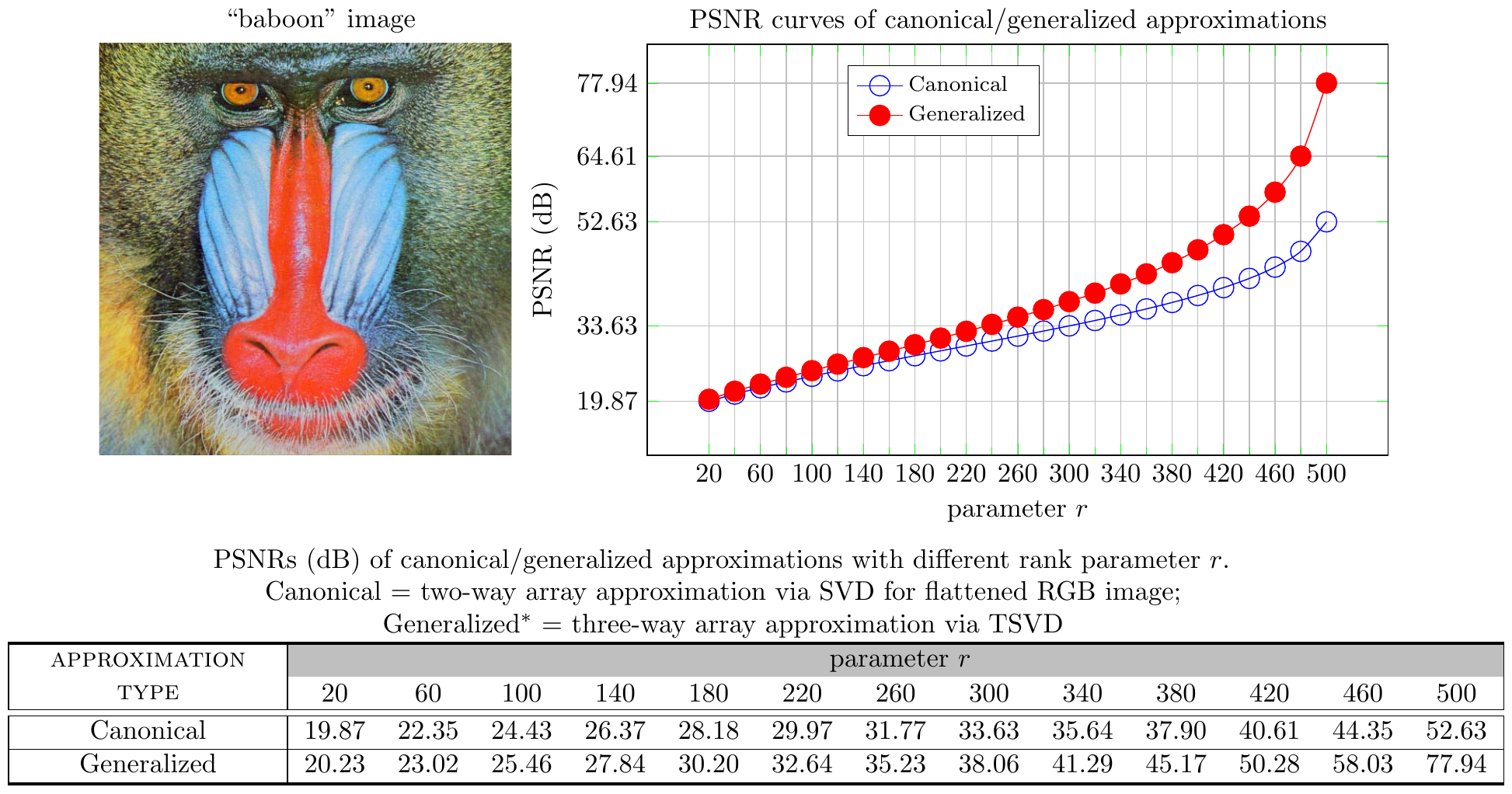}
\caption{A comparison of low-rank approximation by SVD and TSVD on the ``baboon'' image} 
\label{fig:baboo-image}
\end{minipage}
\begin{minipage}[h] {\textwidth}
\centering
\includegraphics[width=0.85\textwidth]{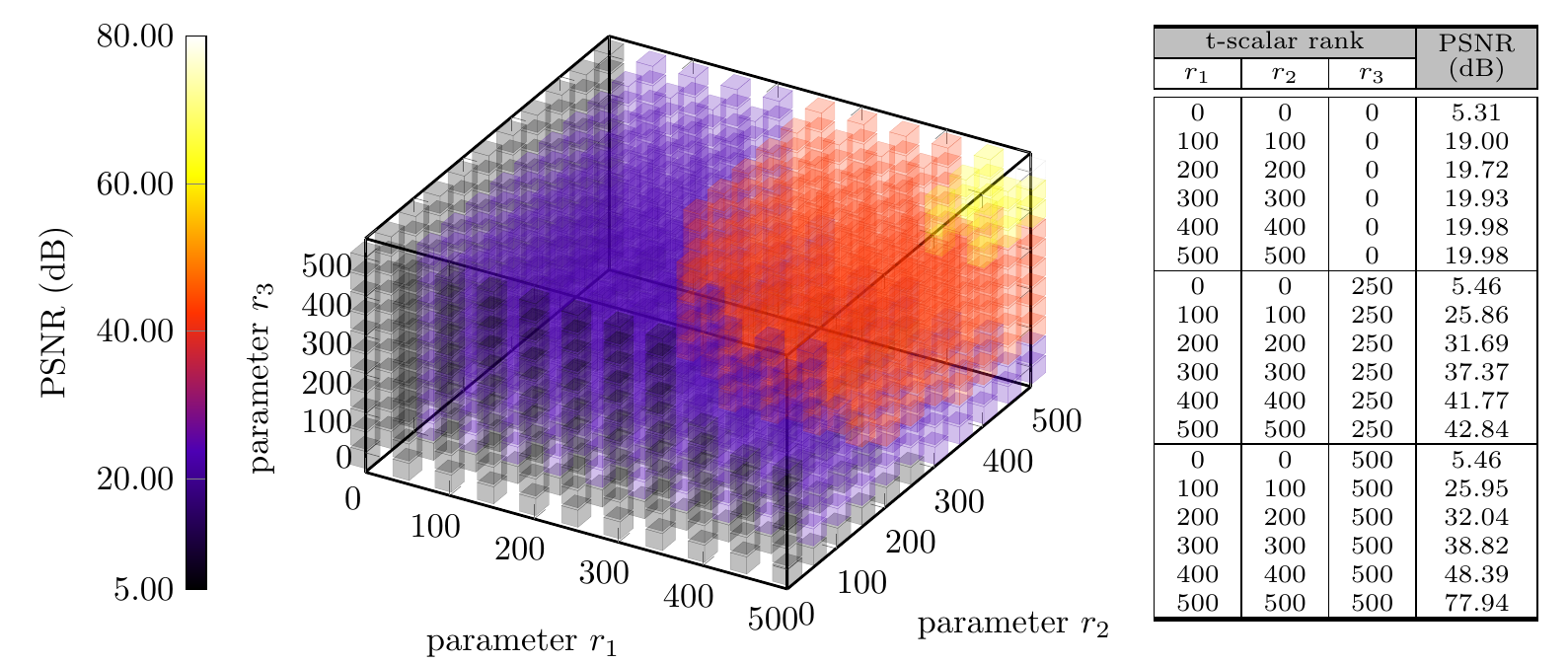} \\
\includegraphics[width=\textwidth]{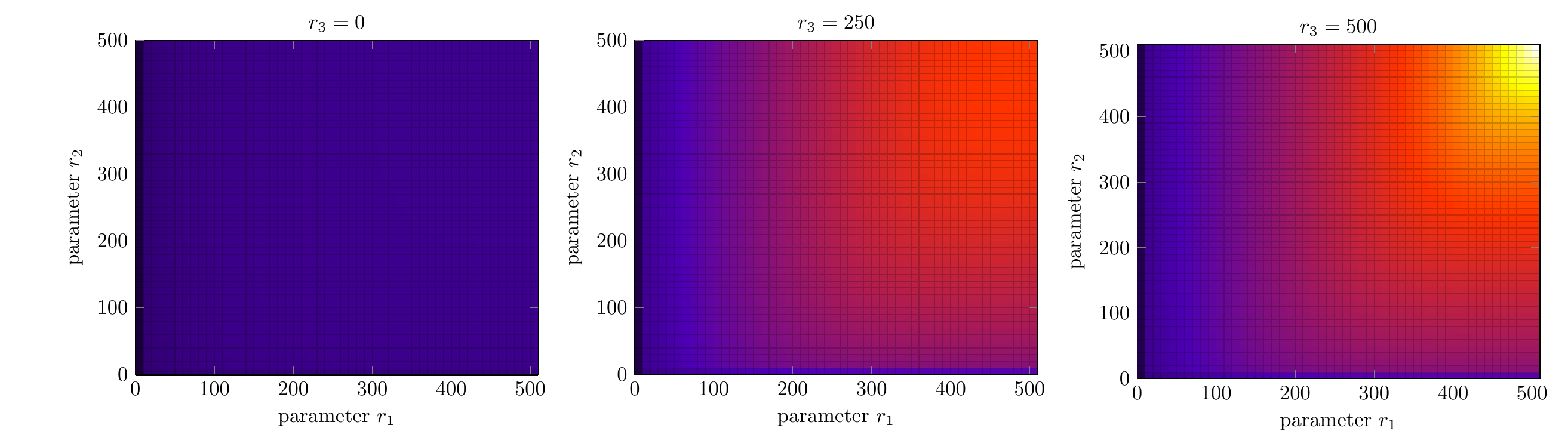}
\caption{PSNR (dB) heat maps of TSVD approximations with different t-matrix ranks, characterized by the tuple $r_1, r_2, r_3$ on the RGB ``baboon'' image}
\label{fig:TSVD-with-general-rank}
\end{minipage}
\end{figure}

Since TSVD applies to order-three arrays while SVD is only applicable to order-two arrays,    
to use 
SVD, the ``baboon'' image is flattened to a 
$512\times 1536$ matrix by 
concatenating each $512\times 512$ sub-image along the horizontal direction.

TSVD applies to order-three arrays of complex numbers, while SVD is only applicable to order-two 
arrays of complex numbers. 
Using SVD, the 
``baboon'' image is flattened to a $512\times 1536$ matrix by concatenating each $512\times 512$ sub-image 
along the horizontal direction. 
Using TSVD, the ``baboon'' image is represented by a t-matrix in 
$C^{\,512\times 512} \equiv \mathbb{C}^{3\times 512\times 512}$. The only 
requirement of transforming the RGB image to the underlying $3\times 512\times 512$ array of the t-matrix is 
a permutation of the indices of the raw $512\times 512\times 3$ array.

To make a reasonable comparison with the SVD approximation, we use a 
simplified TSVD approximation under the constraint 
$\operatorname{rank}(\scalebox{0.85}{$\hat{X}_\mathit{TM}$})  \le r \cdot E_{T}$ where $r \in 
\{0,\cdots,512\}$.

To give a quantitative comparison, when an approximation array $\hat{X}$ of a 
given array $X$ is obtained, the PSNR (Peak Signal Noise Ratio) of the 
approximation is given as follows. 
\begin{equation}
\scalebox{0.92}{$\mathit{PSNR}$} 
= 20 \cdot \log\nolimits_{10} \left(  {\sqrt{N^\mathit{entry} }  \cdot 
\scalebox{0.93}{$\mathit{MAX}$} 
\cdot \|X - \hat{X} \|_F^{-1} }
\right)
\label{equation:PSNR}
\end{equation}
where $N^\mathit{entry}$ denotes the number of scalar entries of the array $X$, and
$\mathit{MAX}$ represents the maximum possible real value in $X$.

In the experiment of using the ``baboon'' image,  
$N^\mathit{entry} = 786432 
\equiv 512 \times 
512 \times 3$, $\mathit{MAX} = 255$.  
Figure \ref{fig:baboo-image} gives the PSNRs of canonical approximation (via SVD) and 
generalized approximation (via TSVD) with 
different rank parameters. It is easy to follow that TSVD consistently outperforms SVD. 
When $r = 500$, the PSNRs of TSVD and SVD differ by more than $25$ dB.

Also note, in an approximation problem as in equation 
(\ref{equation:generalized-low-rank-approximation}), the rank parameter $H_{T}$ can be any t-scalar $Z_{T}\le 
H_{T} \le  \operatorname{rank}(X_\mathit{TM}) $ rather than just in the form of $H_{T} \doteq r \cdot E_{T}$
where $r \in \{0,\cdots,512 \}$.

The peak signal-noise ratios (PSNRs in dB)
of TSVD approximation using different general 
t-matrix ranks subject to $Z_{T}\le H_{T} \le  \operatorname{rank}(X_\mathit{TM}) $
are given in Figure \ref{fig:TSVD-with-general-rank}.

\begin{figure}[thb]
\begin{minipage}[t] {1\textwidth}
\centering
\includegraphics[width=0.95\textwidth]{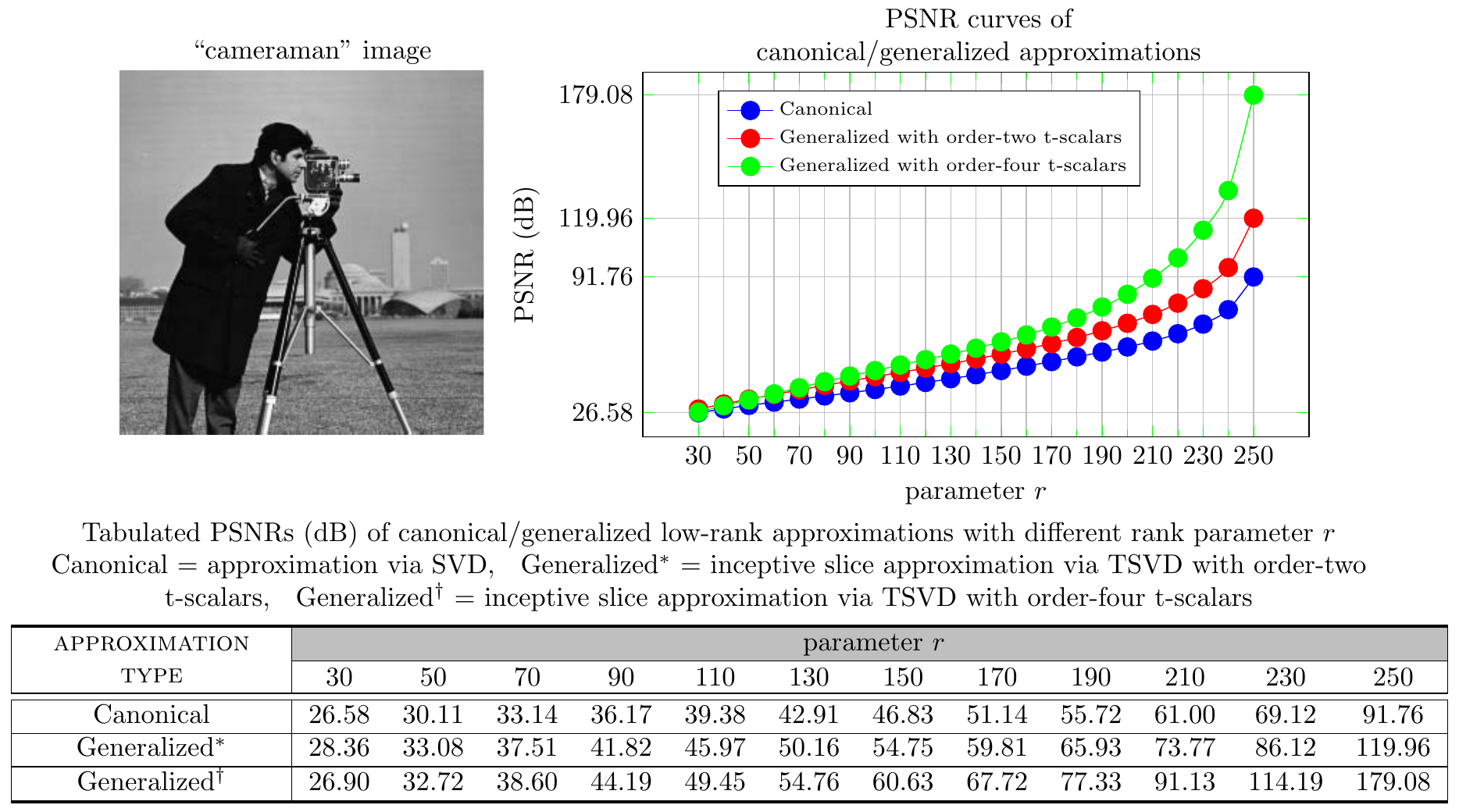} 
\caption{A comparison of approximations by TSVD, using inception slice and different array orders,  
 and SVD where $N^\mathit{entry} = 65536$.}
\label{fig:cameraman-approximation-inception-slice}
\end{minipage}
\begin{minipage}[t] {1\textwidth}
\centering
\includegraphics[width=0.95\textwidth]{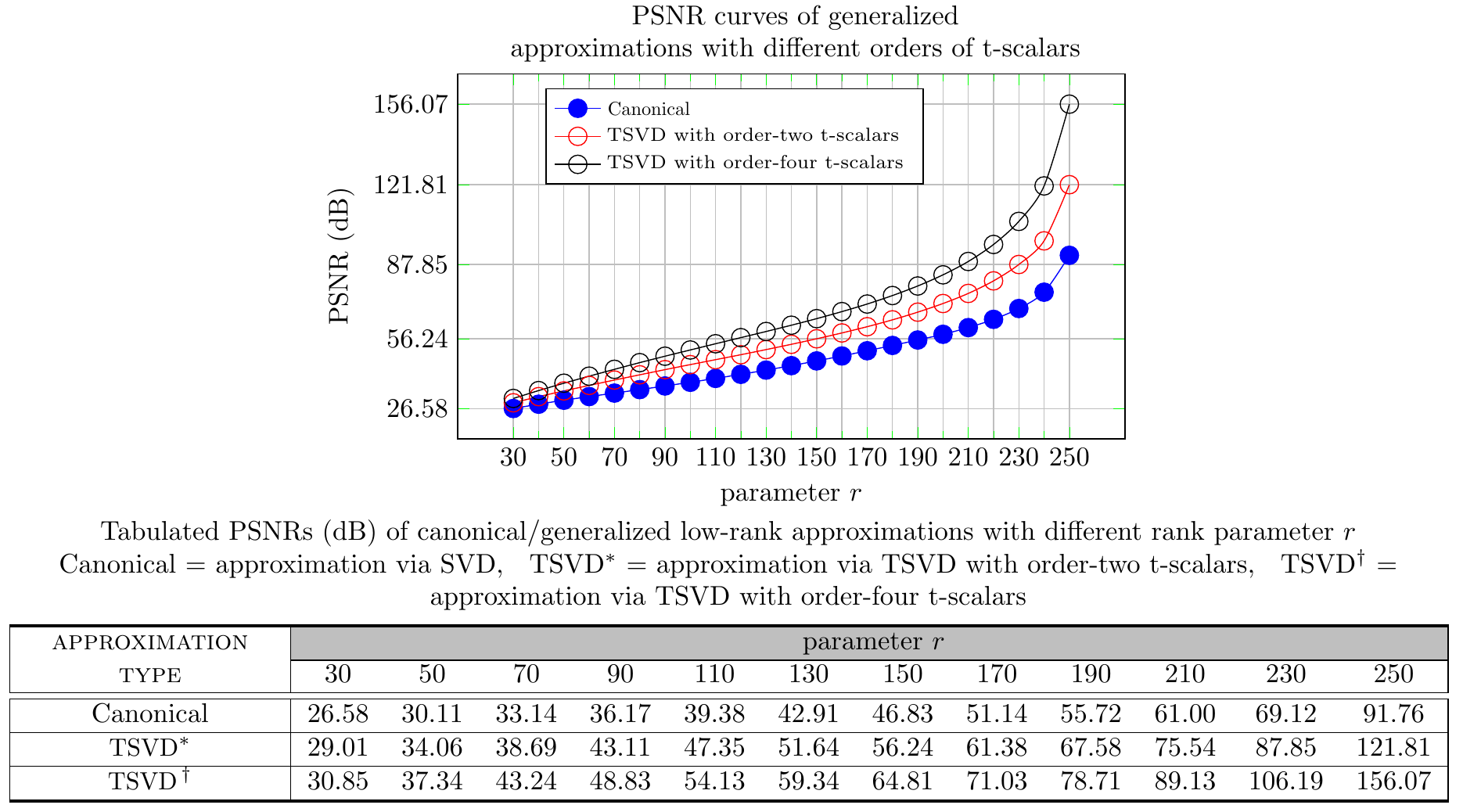} 
\caption{A comparison of approximations by TSVD with $K \in \{9, 81\}$  
 and SVD with $K = 1$ where PSNRs are computed with $N^\mathit{entry} = 65536\times K$.}
\label{fig:cameraman-approximation-higher-order-approximation-TSVD}
\end{minipage}
\end{figure}

Since the generalized rank of the t-matrix $\hat{X}_\mathit{TM} \in C^{512\times 512} \equiv \mathbb{C}^{3\times 512\times 512} $ can be written in the 
form of $\operatorname{rank}(\scalebox{0.92}{$\hat{X}_\mathit{TM}$} ) = 
\scalebox{0.92}{$\sum\nolimits_{k = 1}^{3} r_{k} \cdot  \scalebox{0.9}{$\myQ{k}$} $} 
$, the generalized rank of $\hat{X}_\mathit{TM}$ is equivalently characterized by the tuple $(r_1, r_2, r_3)$.

A 3D heat map 
and three 2D  heat maps
with different $r_1$, $r_2$ and $r_3$ are shown in the figure. It meets the expectation that 
a better approximation is obtained with higher values $r_1$, $r_2$ and $r_3$, or equivalently, a higher t-matrix rank of $\hat{X}_\mathit{TM} \in C^{512\times 512}$.

\subsubsection{``Cameraman'' image} For a second low-rank approximation experiment, the ``cameraman'' image is used. The size of the grey image $256\times 256$.  This image is easy to be approximated via SVD.

To exploit the potential of TSVD,  we use the $3\times 3$ ``inception'' neighborhood strategy (see Figure 
\ref{figure:tensorization001}) to t-matricize
the ``cameraman'' image to a t-matrix in $C^{256\times 256} \equiv \mathbb{C}^{3\times 3\times 256\times 256}$.

In the TSVD approximation experiment,
the rank condition is given by   
$\operatorname{rank}(\scalebox{0.9}{$\hat{X}_\mathit{TM}$} ) \le r \cdot  E_{T} $, where $r$ is a nonnegative integer. Namely, the ``truncated'' TSVD approximation is used.

Note that the approximation by TSVD is an array in $\mathbb{C}^{3\times 3\times 256\times 256}$, while 
the approximation by SVD is an array in $\mathbb{C}^{256\times 256}$. To give a more relevant comparison on 
PSNRs, we extract the ``inception'' slice of the TSVD approximation to compare with the approximation by 
SVD. The ``inception'' slice is the $256\times 256$ matrix by only keeping the first scalar entry of each 
t-scalar in a t-matrix.

To compute PSNRs for this experiment, 
the parameters in equation (\ref{equation:PSNR}) are    
$\scalebox{0.92}{$\mathit{MAX}$} = 255$ and 
$N^\mathit{entry} = 65536 \equiv 256\times 256$.

Figure \ref{fig:cameraman-approximation-inception-slice} shows that the inception slice approximations via TSVD 
consistently outperform the SVD approximation 
in PSNRs. 
When the parameter $r = 250$, the approximation by the inception slice via TSVD with order-two 
t-scalars outperforms the canonical approximation via SVD by more than $28$ dB.

By  reusing the neighborhood t-matricization solution demonstrated in Figure \ref{figure:higher-order-tensorization}, the order of an obtained t-matrix is increased.  
When using order-four 
t-scalars, 
Figure \ref{fig:cameraman-approximation-inception-slice} shows
that an additional gain of more than $59$ dB is obtained, reaching $179.08$ dB.

The approximation results by TSVD shown in Figure \ref{fig:cameraman-approximation-inception-slice} are 
computed via the inception slice of a TSVD approximation. One might be interested in the PSNR of a whole TSVD 
approximation rather than its slice.

To this end, another type of PSNRs is computed. More precisely, given a higher-order array 
$X_\mathit{TM} \doteq  X \in C^{256\times 256}$ and its approximation $\hat{X}_\mathit{TM} \doteq \hat{X} \in 
C^{256\times 256}$, its PSNR is computed as in equation (\ref{equation:PSNR}) with $MAX = 255$ and 
$N^\mathit{entry} = 256 \times 256 \times K \doteq 65536 \times K $, where $K \doteq \myK $ 
denotes the number of scalar-entries in a t-scalar.

Using the neighborhood strategy of data t-matricization (see Figures \ref{figure:tensorization001} and 
\ref{figure:higher-order-tensorization}), two distinct t-scalar sizes are 
adopted in this experiment. More precisely, the order-two t-scalars are elements of 
$C \equiv \mathbb{C}^{\,3\times 3} $, i.e., $K = 9$.  
The order-four t-scalars are elements of 
$C \equiv \mathbb{C}^{\,3\times 3\times 3 \times 3} $, i.e., $K = 81$.

Two distinct t-scalar sizes are adopted in approximating the ``cameraman'' image. More precisely, order-two 
t-scalars are elements of $C \equiv \mathbb{C}^{\,3\times 3} $, i.e., $K = 9$, and order-four t-scalars are 
elements of $C \equiv \mathbb{C}^{\,3\times 3\times 3 \times 3} $, i.e., $K = 81$.

Figure \ref{fig:cameraman-approximation-higher-order-approximation-TSVD} shows the PSNR curves of high-order 
approximations with respectively the order-two t-scalars ($K = 9$) and the 
order-four t-scalars ($K = 81$) and the PSNR curve obtained by SVD ($K = 1$) where 
the parameter $N^\mathit{entry} \doteq 65536\times K$ is given 
in equation (\ref{equation:PSNR}).  
The PSNR comparison shown in Figure 
\ref{fig:cameraman-approximation-higher-order-approximation-TSVD} corroborates 
the outperformance of TSVD over SVD, shown in Figure \ref{fig:cameraman-approximation-inception-slice}.

\begin{figure}[thb]
\begin{minipage}[h] {1\textwidth}
\includegraphics[width=0.95\textwidth]{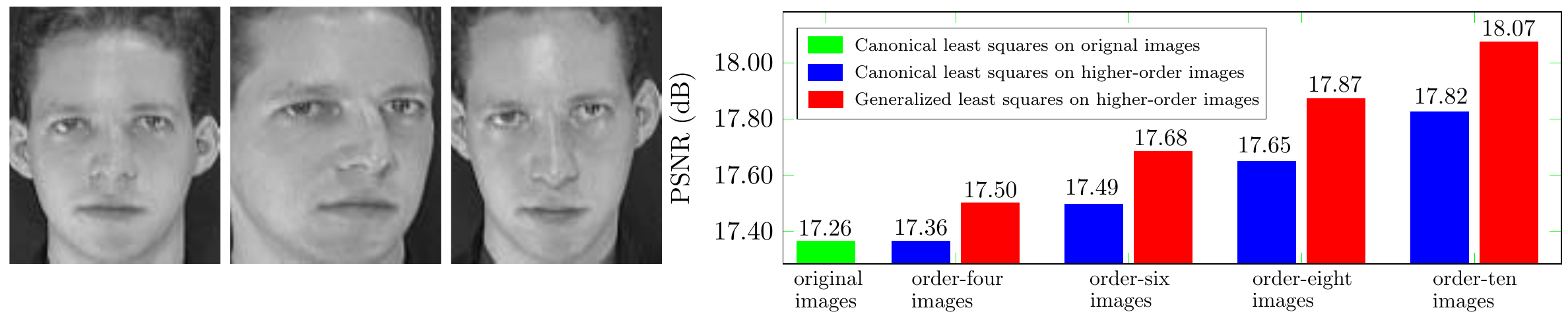}
\includegraphics[width=0.95\textwidth]{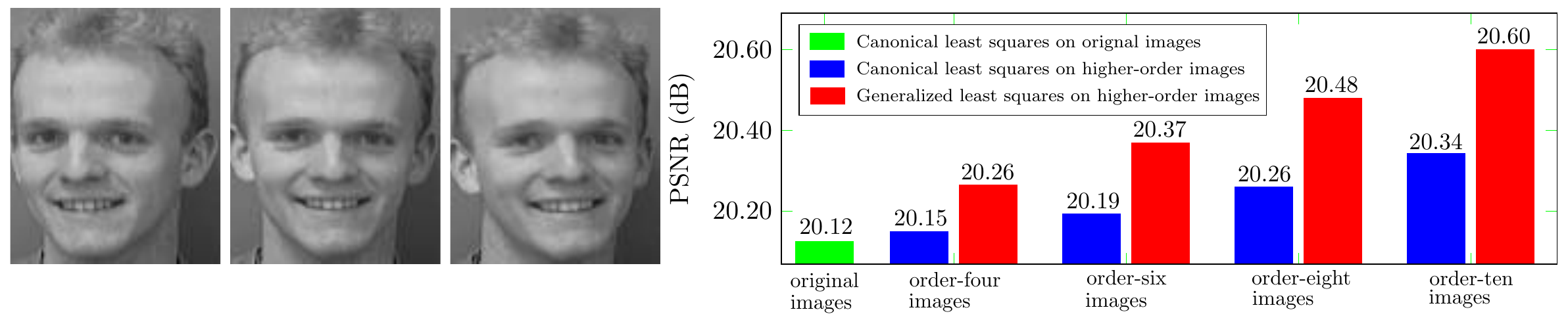}
\vspace{1.5em}
\caption{A quantitative comparison of approximations 
in PSNR 
by canonical least squares and generalized least squares on the ORL image dataset}
\label{figure:images-from-orl-data-set}
\end{minipage}
\end{figure}

\subsection{Generalized Least-squares}

To evaluate canonical least squares and generalized least squares, we compare their performances on 
approximating images. The experimental images are from the publically available ORL dataset, which 
contains $400$ facial images from $40$ 
subjects. \footnote{https://www.cl.cam.ac.uk/research/dtg/attarchive/facedatabase.html.}
Each of these images is monochrome and has $112\times 92$ pixels. We choose the images from two classes
for the experiment.

Figure 
\ref{figure:images-from-orl-data-set} shows the chosen images from each class ---
for each class, the first three images of each class are chosen. These images are further t-matricized to higher-order 
arrays, i.e., t-matrices.

For each class, the experiment uses the last two of the chosen images, i.e., t-matrices, to approximate 
the first one via the generalized least-squares.

Let the three images in the form of t-matrices be $A_\mathit{TM}, 
B_\mathit{TM}, C_\mathit{TM} \in  C^{112\times 92}$. 
The goal 
is to use a generalized linear combination of $\lambda_{T} \circ B_\mathit{TM} + \xi_{T} \circ C_\mathit{TM}$, where $\lambda_{T}, \xi_{T} \in C$, 
to approximate 
$A_\mathit{TM}$.

The following equation gives the optimal approximation 
$A_\mathit{TM}^\mathit{opt} \doteq \lambda_{T}^\mathit{opt} \circ B_\mathit{TM} + 
\xi_{T}^\mathit{opt} \circ C_\mathit{TM}  $ of the t-matrix $A_\mathit{TM}$ by
\begin{equation}
\begin{aligned}
&\hspace{1.3em}r(A_\mathit{TM}^\mathit{opt}  - A_\mathit{TM})_{F}
\doteq r(\lambda_{\,T}^\mathit{\,opt} \circ B_\mathit{TM}  + \xi_{\,T}^\mathit{\,opt} \circ C_\mathit{TM}  - 
A_\mathit{TM})_{F}   \\
&= \mathop{\operatorname{min}}\nolimits_{\lambda_{T}, \,\xi_{T} \,\in C} 
\,\,\scalebox{1.15}{$r$}\big(\lambda_{T} \circ B_\mathit{TM} + 
\xi_{\,T} \circ C_\mathit{TM} - A_\mathit{TM} 
\big)_F \ge Z_{T} \;\;\;.
\end{aligned}
\label{equation:experiment-generalized-least-squares}
\end{equation}

The problem 
in equation (\ref{equation:experiment-generalized-least-squares}) 
can be recast to a generalized least-squares problem in 
equation (\ref{equation:generalized-least-squares}) to obtain the t-matrix $A_\mathit{TM}^\mathit{opt} \in 
C^{112\times 92}$.

When the approximation t-matrix $A_\mathit{TM}^\mathit{opt} \in C^{\,112\times 92} 
\;{\color{black}\equiv}\; \mathbb{C}^{I_1\times 
\cdots \times I_N \times \,112 \times \,92}$ is obtained, 
the PSNR of the approximation 
is computed with $\scalebox{0.92}{$\mathit{MAX}$} \doteq 255$, 
$N^\mathit{entry} \doteq 112\times 92\times K$ where $K = \myK$.

To have a comprehensive comparison, we have the experiment images t-matricized using or 
reusing $3\times 3$ neighborhood strategy to arrays of order-four, order-six, order-eight, and order-ten.

To have a comprehensive comparison, we have the experiment images t-matricized 
using or reusing $3\times 3$ neighborhood strategy to arrays of 
order-four (where \scalebox{0.9}{$N = 2, I_1 = I_2 = 3$}), 
order-six (where \scalebox{0.9}{$N = 4, I_1 = \cdots = I_4 = 3$}), 
order-eight (where \scalebox{0.9}{$N = 6, I_1 = \cdots = I_6 = 3$}), 
and order-ten (where \scalebox{0.9}{$N = 8, I_1 = \cdots = I_8 = 3$}).

Besides the generalized least-squares, the canonical least-squares also applies to approximate higher-order images in the form of t-matrices.

More precisely, given 
higher-order images 
$A_\mathit{TM}, B_\mathit{TM}, C_\mathit{TM} \in \mathbb{C}^{I_1\times \cdots \times I_N\times M_1\times M_2}$, 
the canonical least-squares use the linear combination 
$\alpha \cdot B_\mathit{TM} + \beta \cdot C_\mathit{TM} $ 
to approximate $A_\mathit{TM}$, where $\alpha$ and $\beta$ are complex numbers.

The following equation gives the optimal approximation 
$A_\mathit{TM}^\mathit{opt} \doteq \alpha^\mathit{opt} \cdot B_\mathit{TM}
+ \beta^\mathit{opt} \cdot C_\mathit{TM}$ of the t-matrix $A_\mathit{TM}$ by
\begin{equation}
\begin{aligned}
&\hspace{1.3em}\big\|\scalebox{0.92}{$A_\mathit{TM}^\mathit{opt} - A_\mathit{TM}$}   \big\|_F   
\doteq \big\|\scalebox{0.92}{$\alpha^\mathit{opt} \cdot B_\mathit{TM} + \beta^\mathit{opt} \cdot C_\mathit{TM}$}    
- A_\mathit{TM} \big\|_F   \\
&= \min\nolimits_{\alpha,\,\,\beta \,\in\, \mathbb{C}}
\big\|\scalebox{0.92}{$\alpha \cdot B_\mathit{TM} + \beta \cdot C_\mathit{TM}$} - A_\mathit{TM}  \big\|_F  \;\geqslant\; 0 \;\;\;.
\end{aligned}
\end{equation}

Figure \ref{figure:images-from-orl-data-set} gives the PSNRs by the canonical 
least-squares and generalized least-squares. The PSNRs by the canonical least-squares on the original ORL images is $17.26$ dB and $20.12$ dB. 
Higher-order images contribute higher PSNRs by the canonical least-squares. 
However, the generalized least-squares outperform the canonical least-squares on 
higher-order images, yielding higher quality of approximation with higher PSNRs. The 
highest PSNRs are yielded by the generalized least-squares, namely $18.07$ dB and $20.60$ dB 
on the chosen experiment images of each class.

\subsection{Generalized Principal Component Analysis}
\label{section:image-reconstruction-TPCA}
To show its performance, we use the generalized principal component analysis (TPCA) to 
extract features of the public CIFAR-10 image dataset. 
\footnote{http://www.cs.toronto.edu/~kriz/cifar.html}

The CIFAR-10 dataset contains thousands of color images, each image a $32\times 32\times 3$ 
array. We choose the first $36$ images of the first training set of the dataset for 
extracting principal t-vectors or vectors, namely $N = 36$.

\subsubsection{TPCA}
\label{section:TPCA}

\begin{figure}[t]
\includegraphics[width=0.99\textwidth]{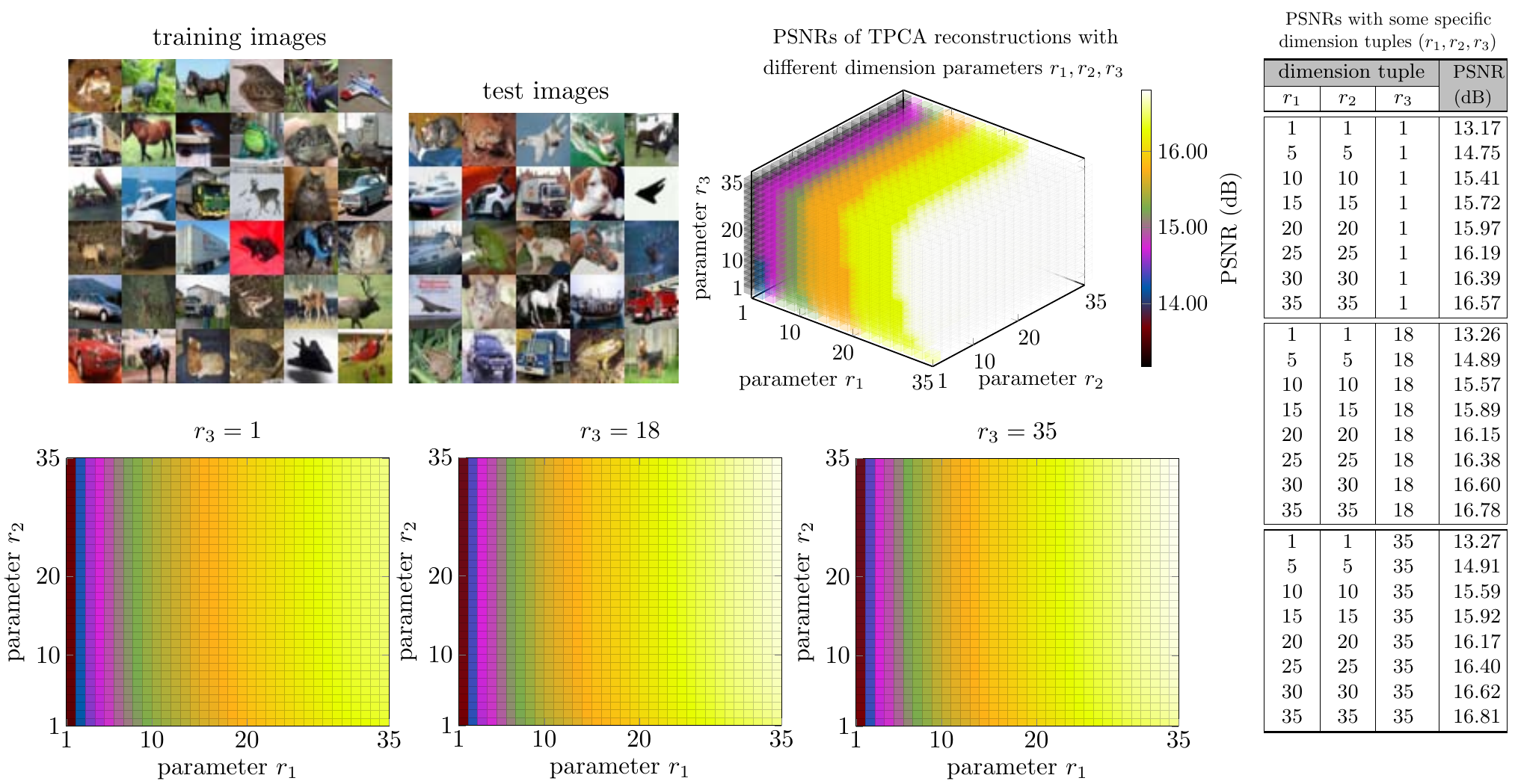}
\caption{PSNRs by TPCA with various parameters $(r_1, r_2, r_3)$, or equivalently,  
$H_{T} = \raisebox{0.1em}{\scalebox{0.9}{$\sum\nolimits_{k=1}^{3}$}}\,  r_k \cdot \myQ{k}
$, on the CIFAR-10 images.}
\label{figure:images-from-CIFAR-imageset}
\end{figure}

The first $25$ images from the test set of the dataset are chosen as the query 
images. 
The subfigures in the first column of Figure \ref{figure:images-from-CIFAR-imageset} show the 
chosen training images and query images.

Each raw CIFAR-10 image is an order-three array in $\mathbb{C}^{32\times 32\times 3}$. With a permutation 
of entry index and then an array reshaping, each order-three array can be transformed into  
a t-vector in $C^{1024} \equiv \mathbb{C}^{3 \times 1024} \equiv \mathbb{C}^{3\times 1\times \cdots \times 
1\times 1024\times 1}$. 
Namely, $I_1 = 3, I_2 = \cdots =I_N = 1$, $M_1 = 1024$, $M_2 = 1$, and $1024 = 32\times 32$ in the form of $\mathbb{C}^{I_1\times \cdots \times I_N \times 
M_1\times M_2}$.

In this image approximation experiment, TPCA 
works on the obtained t-vectors in $C^{1024} \equiv \mathbb{C}^{3\times 1024}$, where all t-scalars are 
order-one arrays, each containing three scalar-entries, namely, $K = 3$. 
Then, given a generalized dimension parameter $H_{T} \in S^\mathit{nonneg}$, it can be represented by a 
$3$-tuple of nonnegative integers $r_1$, $r_2$, and $r_3$ in the following form 
\begin{equation}
H_{T} = 
r_1 \cdot \myQ{1} \,+\, r_2 \cdot \myQ{2} \,+\, r_3 \cdot \myQ{3}
\label{equation:generalized-dimension}
\end{equation} 
where $r_1, r_2, r_3 \in \{0,\cdots, Q_{m}\} \equiv \{0,\cdots,35\}$.

With the parameter $H_{T}$ or equivalently, the tuple $(r_1, r_2, r_3)$, each of the $25$ query 
t-vector is reconstructed as in equation (\ref{equation:TPCA-reconstruction}). 
Each reconstructed image is then obtained by transforming its counterpart of the reconstructed t-vector into the original form, namely, a $32\times 32\times 3$ array.

After having all the $25$ raw and reconstructed query images, we arrange them in two 
$3072\times 25$ arrays. Precisely, each image is reshaped to a column of length 
$3072$ where $3072 = 32\times 32\times 3$. Then, the PSNR is computed as in equation (\ref{equation:PSNR})
with $N^\mathit{entry} = 76800 = 3072 \times 25 = (32\times 32\times 3) \times 25$ 
and $\mathit{MAX} = 255$.

The PSNRs of the TPCA reconstructions are given in Figure (\ref{figure:images-from-CIFAR-imageset}).
Figure (\ref{figure:images-from-CIFAR-imageset}) shows that a larger value of $H_{T}$ contributes to a higher PSNR 
of TPCA reconstruction.  When $H_{T} = E_{T}$, or equivalently $r_1 = r_2 = r_3 = 1$, the PSNR of TPCA 
reconstruction is $13.17$ dB. When $H_{T} \doteq Q_m \cdot E_{T} = 35 \cdot E_{T}$ (or 
equivalently, $r_1 = r_2 = r_3 = Q_m \doteq N - 1 = 35$), the PSNR is $16.81$ dB.

\subsubsection{PCA vs. TPCA}

Note that both PCA and TPCA apply to extract principal components and reconstruct RGB images in the form of 
higher-order arrays.

One might be interested in comparing the performance of PCA and TPCA. In this part, we compare the results of 
TPCA and PCA on reconstructing the CIFAR-10 images.

The underlying $32\times 32\times 3$ array of each RGB image is reshaped to a 
$3072$-dimensional vector When using PCA. 
Hence, with the same samples introduced in Section \ref{section:TPCA},  
we have $36$ training vectors and can extract $35$ principal component vectors. 
The first $r$ principal vectors are used to reconstruct each 
of the $25$ query vectors, where $r \leqslant 35$.

To have a fair and reasonable comparison to PCA, TPCA works on the same CIFAR-10 images, as already reported in 
Section \ref{section:TPCA}, but with the generalized dimension parameter $H_{T}$ in equation 
(\ref{equation:generalized-dimension})
constrained with the so-called ``truncated'' condition $r_1 = r_2 = r_3 \equiv r$. More precisely,
the parameter {\color{black}$H_{T} \geq Z_{T}$} is rewritten as follows
\begin{equation}
H_{T} \doteq r \cdot E_{T} \equiv  
r \cdot \myQ{1} \,+\, r \cdot \myQ{2} \,+\, r \cdot \myQ{3}
\label{equation:generalized-dimension}
\end{equation} 
where $r \in \{0,\cdots,Q_m\} \equiv \{0,\cdots, 35\}$.

Note that, in this experiment, only image reshape and scalar index permutation are 
adopted. No data t-matricization, i.e., the proposed neighborhood strategy, is employed. In other words, PCA and TPCA use the same raw images, only reorganized in different 
formats.

\begin{figure}[h]
\begin{minipage}[t] {0.95\textwidth}
\centering
\includegraphics[width=0.8\textwidth]{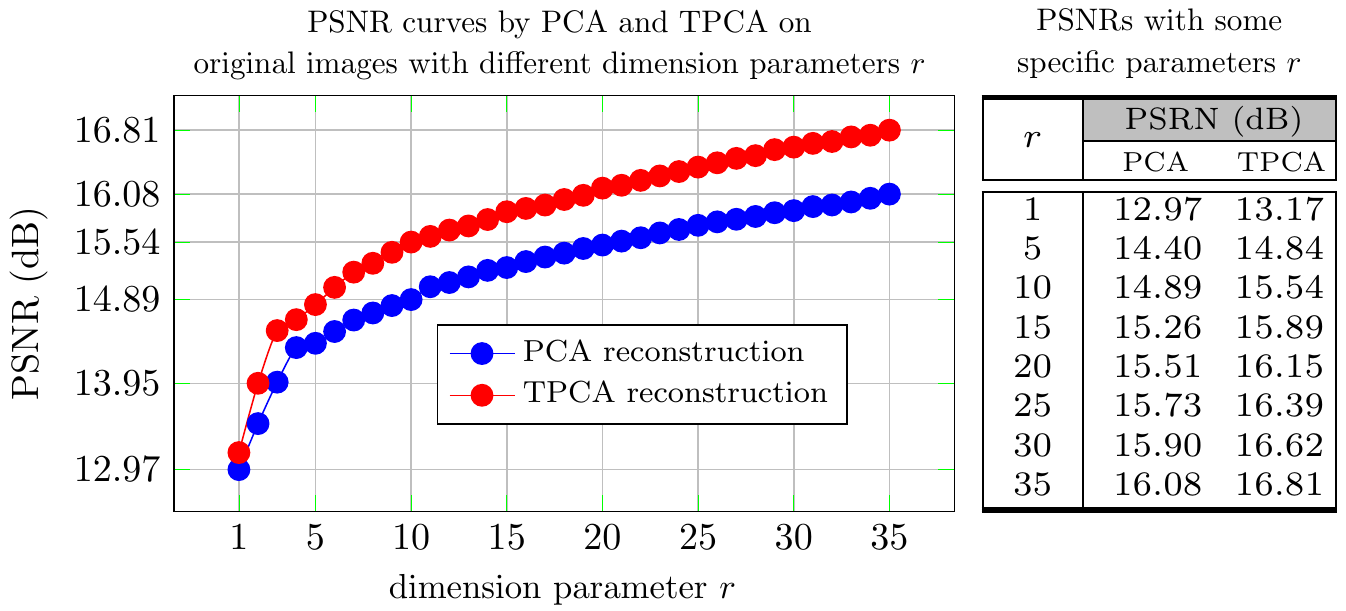}
\caption{PSNRs by PCA and TPCA with various paramters $r$ on the CIFAR-10 images}
\label{figure:images-reconstruction-PCA-TPCA-on-CIFAR-10-images}
\end{minipage}
\begin{minipage}[t] {0.95\textwidth}
\centering
\includegraphics[width=0.8\textwidth]{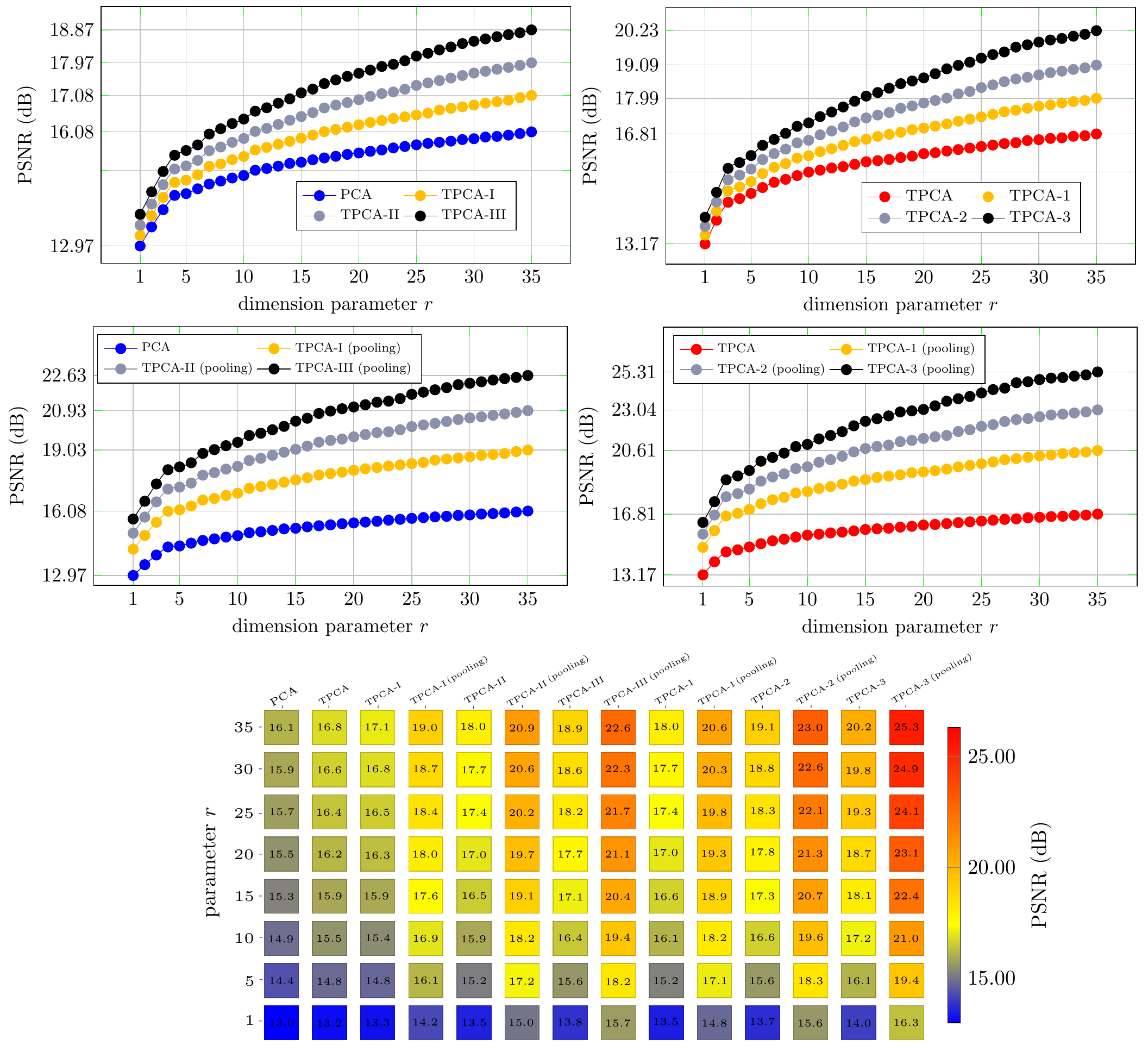}
\caption{A comparison of PRNS by PCA and TPCA variants with various t-scalar orders
on the CIFAR-10 images. \,\,\,
TPCA-I = TPCA with order-two t-scalars,\;
TPCA-II = TPCA with order-four t-scalars,\;
TPCA-III = TPCA with order-six t-scalars,\;
TPCA-1 = TPCA with order-three t-scalars,\;
TPCA-2 = TPCA with order-five t-scalars,\;
TPCA-3 = TPCA with order-seven t-scalars\;
}
\label{figure:images-from-CIFAR-10-images-with-higher-order-tscalar}
\end{minipage}
\end{figure}

The PSNRs of the PCA and TPCA reconstructions are computed with the same settings, as described 
in Section \ref{section:TPCA}.  
These PSNRs are both tabulated and are shown as the curves of parameter $r$ in 
Figure \ref{figure:images-reconstruction-PCA-TPCA-on-CIFAR-10-images}. 
It shows that TPCA consistently outperforms PCA on the CIFAR-10 images 
in terms of PSNR.

\subsubsection{TPCA with Higher-order T-scalars}

One might also be interested in the effect of higher-order t-scalars on the performance of a generalized application. To this end, we adopt t-scalars of different orders with TPCA to conduct image approximations.

Each experiment RGB image is a $32\times 32\times 3$ array of real numbers and has three monochrome 
subimages in the form of a $32\times 32$ array. Using the $3\times 3$ neighborhood 
strategy (see Figure \ref{figure:tensorization001}) on each 
monochrome subimage, an experiment RGB image is t-matricized from an order-three array in 
$\mathbb{C}^{32\times 32\times 3}$
to an order-five array $\mathbb{C}^{3\times 3\times 32\times 32\times 3}$.

With simple manipulations, an obtained order-five array can be transformed into at least two versions of 
the t-vector. The two versions are described as follows.

Version $1$: 
An order-five array in $\mathbb{C}^{3\times 3\times 32\times 32\times 3}$ is reshaped to an 
order-three array 
in $\mathbb{C}^{3\times 3\times 3072}$, where
the obtained array is algebraically interpreted as a t-vector in $C^{3072}$  
and $3072 = 32\times 32\times 3$,  
namely, in this scenario, $C \equiv \mathbb{C}^{3\times 3} \doteq \mathbb{C}^{3^{2}}$.   

Version $2$: 
Alternatively, an order-five array in $\mathbb{C}^{3\times 3\times 32\times 32\times 3}$ can be first
permuted to an array of the same order in $\mathbb{C}^{3\times 3 \times 3 \times 32\times 32}$,
and then reshaped to an order-four array in $\mathbb{C}^{3\times 3\times 3 \times 1024}$ with $1024 = 32\times 32$, which is algebraically  
interpreted as a t-vector in $C^{1024}$, namely, in this scenario, $C \equiv \mathbb{C}^{3\times 3\times 
3} \doteq \mathbb{C}^{3^{3}}$.

TPCA adopts the two versions of t-vectors in the experiment of this subsection. TPCA, with the above first version of t-vectors, is referred to as TPCA-I 
with $C \equiv \mathbb{C}^{3\times 3} \doteq \mathbb{C}^{3^{2}}$. The second version is referred to as TPCA-1 with $C \equiv \mathbb{C}^{3\times 3\times 3} \doteq \mathbb{C}^{3^{3}}$.

On the other hand, reusing the $3\times 3$ neighborhood strategy, 
as shown in Figure \ref{figure:higher-order-tensorization},
makes it easy to increase the order of adopted t-scalars. In the experiment, 
TPCA using t-vectors in $C^{3072}$ 
and $C \equiv \mathbb{C}^{3\times 3\times 3\times 3} \doteq \mathbb{C}^{3^{4}}$ is referred to as TPCA-II.  
TPCA using t-vectors in $C^{3072}$ 
and $C \equiv \mathbb{C}^{3\times 3\times 3\times 3\times 3\times 3} \doteq \mathbb{C}^{3^{6}}$ is referred 
to as TPCA-III.

Similarly, TPCA using t-vectors in $C^{1024}$ with $C \equiv \mathbb{C}^{3\times 3\times 3\times 3\times 3}
\doteq \mathbb{C}^{3^{5}}$ is referred to as TPCA-2, and 
TPCA using t-vectors in $C^{1024}$ with $C \equiv \mathbb{C}^{3\times 3\times 
3\times 3\times 3 \times 
3\times 3} \doteq \mathbb{C}^{3^{7}}$ is referred to as TPCA-3.

Thus, there are six variants of TPCA using t-scalars of different higher-orders in performance comparisons. 
For clarity, we summarize their t-vector settings t-vectors in Table \ref{tab:experiment-settings}. As a bottom line for performance comparison, Table \ref{tab:experiment-settings} also gives the settings of TPCA with low-order t-scalars and PCA.

\newcommand{\numberOfScalars}[2]{
$
\begin{matrix}
\text{#1} \\
\vspace{-1.5em}\\
\text{#2} \\
\end{matrix}
$
}

\newcommand{\TscalarsIsEqualToScalar}[4]{
$
\begin{matrix}
\vspace{-1.5em}\\
\text{#1} \\
\vspace{-1.5em}\\
\text{#2} \\
\vspace{-1.5em}\\
\text{#3} \\
\vspace{-1.5em}\\
\text{#4} \\
\end{matrix}
$
}

\newcommand{\myrasie}[1]{
\raisebox{-0.15em}{$#1$}
}

\newcommand{\mss}[1]{
$\underset{#1~\text{copies}}{\underbrace{3\times \cdots \times 3}}$
}

\begin{table}[tbp]
\centering
\caption{
\resizebox{0.75\textwidth}{!}{\sc
T-vector settings for the experiment of TPCA with t-scalars of different orders} 
} 
\vspace{0.1em}
\vspace{-0.99em}
\resizebox{\textwidth}{!}{

\begin{tabular}{V{6}cV{4}  cV{1}cV{6}V{6}  cV{1}cV{6}V{6}  cV{1}cV{6}V{6} cV{1}cV{6}}
\hline

\hline

\hline

\hline

\multirow{2}[0]{*}{\numberOfScalars{\footnotesize t-vector}{\footnotesize of settings}} & \multicolumn{8}{cV{6}}{\raisebox{-0.15em}{methods}} \\
    \cline{2-9}          &
          \raisebox{-0.08em}{\small  TPCA-I} & 
          \raisebox{-0.08em}{\small  TPCA-1}   & 
          \raisebox{-0.08em}{\small  TPCA-II}  & 
          \raisebox{-0.08em}{\small  TPCA-2}   & 
          \raisebox{-0.08em}{\small  TPCA-III} & 
          \raisebox{-0.08em}{\small  TPCA-3}   &
          \raisebox{-0.08em}{\small  TPCA}   & 
          \raisebox{-0.08em}{\small  PCA}  \\
\hline 
\hline
\raisebox{-0.5em}{
\numberOfScalars{\footnotesize shape}{\footnotesize of t-scalars} }    & 
\raisebox{0.1em}{\scalebox{0.9}{\mss{2}}}   & 
\raisebox{0.1em}{\scalebox{0.9}{\mss{3}}}   & 
\raisebox{0.1em}{\scalebox{0.9}{\mss{4}}}   & 
\raisebox{0.1em}{\scalebox{0.9}{\mss{5}}}   &
\raisebox{0.1em}{\scalebox{0.9}{\mss{6}}}   & 
\raisebox{0.1em}{\scalebox{0.9}{\mss{7}}}   &
\raisebox{-0.5em}{\scalebox{0.9}{$3$}} &
\raisebox{-0.5em}{\small 1}\\
\hline

\raisebox{-0.5em}{
\numberOfScalars{\footnotesize orders}{\footnotesize of t-scalars} }    & 
\raisebox{-0.5em}{\small order-two}   & 
\raisebox{-0.5em}{\small order-three}   & 
\raisebox{-0.5em}{\small order-four}   & 
\raisebox{-0.5em}{\small order-five}   &
\raisebox{-0.5em}{\small order-six}   & 
\raisebox{-0.5em}{\small order-seven}   &
\raisebox{-0.5em}{\small order-one} &
\raisebox{-0.5em}{\small order-zero}
\\
\hline

    \numberOfScalars{\footnotesize number of}{\footnotesize t-scalar entries}      & 
    \numberOfScalars{\footnotesize $3072$}{\footnotesize $=32 \times 32\times 3$}    & 
    \numberOfScalars{\footnotesize $1024$}{\footnotesize $=32 \times 32$}    & 
    \numberOfScalars{\footnotesize $3072$}{\footnotesize $=32 \times 32\times 3$}    & 
    \numberOfScalars{\footnotesize $1024$}{\footnotesize $=32 \times 32$}    & 
    \numberOfScalars{\footnotesize $3072$}{\footnotesize $=32 \times 32\times 3$}    & \numberOfScalars{\footnotesize $1024$}{\footnotesize $=32 \times 32$}  & 
    \numberOfScalars{\footnotesize $1024$}{\footnotesize $=32 \times 32$}  &
    \multirow{2}[0]{*}{
    \raisebox{0.8em}{
    \TscalarsIsEqualToScalar{\footnotesize $3072$}{\footnotesize $=32\times 32 \times 3$}{
    \footnotesize (i.e., t-scalars reduces}{\footnotesize to canonical scalars)}
    }
    }     \\
\cline{1-8}
    \numberOfScalars{\footnotesize $N^\mathit{scalar} = $ number}{\footnotesize of scalar entries}     & \numberOfScalars{\footnotesize $27648$}{\footnotesize $=3072 \times 3^{2}$}    & 
    \numberOfScalars{\footnotesize $27648$}{\footnotesize $=1024 \times 3^{3}$}    &
    \numberOfScalars{\footnotesize $248832$}{\footnotesize $=3072 \times 3^{4}$}     & 
    \numberOfScalars{\footnotesize $248832$}{\footnotesize $=1024 \times 3^{5}$}     & \numberOfScalars{\footnotesize $2239488$}{\footnotesize $=3072 \times 3^{6}$}       & 
    \numberOfScalars{\footnotesize $2239488$}{\footnotesize $=1024 \times 3^{7}$}    & 
    \numberOfScalars{\footnotesize $3072$}{\footnotesize $=1024 \times 3$} &       \\
\hline

\hline

\hline

\hline
\end{tabular}
}
\label{tab:experiment-settings}
\end{table}

Figure \ref{figure:images-from-CIFAR-10-images-with-higher-order-tscalar} shows 
the results with the dimension parameter $r$ by algorithms PCA, TPCA, 
and other variants of TPCA on the $25$ CIFAR-10 images appropriately t-matricized when necessary.

Among these algorithms, 
PCA, TPCA-I, TPCA-II, and TPCA-III respectively use order-zero, order-two, one-four, and order-six t-scalars entries, i.e., even-number-order t-scalars.\footnote{~Scalars are a special case of t-scalars, namely order-zero t-scalars.} 
The top-left subfigure of Figure \ref{figure:images-from-CIFAR-10-images-with-higher-order-tscalar} shows Their PSNR curves over the dimension parameter $r$. 
On the other hand, TPCA, TPCA-1, TPCA-2, and TPCA-3 respectively use 
order-one, order-three, order-five, and order-seven t-scalars 
entries, i.e., odd-number-order t-scalars. 
The top-right subfigure shows their PSNR curves.

Also, a few words for computing PSNRs, let $\mathit{N}^\mathit{scalar}$ be the 
number of scalar entries of each vector or t-vector employed by a specific 
algorithm. Then, given $25$ test images, or their t-matricized versions, cast to $25$ vectors, or t-vectors, one can arrange them and their approximation versions to two arrays of $N^\mathit{entry}$ 
scalars with 
$N^\mathit{entry} \doteq 25\cdot N^\mathit{scalar}$ and use equation (\ref{equation:PSNR}) to compute the PSNRs yielded by an algorithm.

Two observations are apparent from the first row of Figure 
\ref{figure:images-from-CIFAR-10-images-with-higher-order-tscalar}. 
(\romannumeral1) A higher-dimensional 
parameter $r$ always corresponds to a higher quality of reconstruction in terms of PSNR. 
(\romannumeral2) Higher-order methods outperform their lower-order counterparts in 
terms of reconstruction quality with the same parameter $r$.

Note that both PCA and TPCA are applied to the same information, cast in two different formats. Similar scenarios also occur to the pairs of TPCA-I and TPCA-1, TPCA-II and TPCA-2, as well as TPCA-III and TPCA-3, where TPCA-1, TPCA-2, and TPCA-3 
are respectively with higher-order t-scalars but a smaller number of t-scalar entries than their counterparts TPCA-I, TPCA-II, TPCA-III.

The first row of Figure 
\ref{figure:images-from-CIFAR-10-images-with-higher-order-tscalar} shows that, even on the same 
information, TPCA, TPCA-1, TPCA-2, and TPCA-3, using higher-order t-scalars, respectively outperform their counterparts PCA, TPCA-I, TPCA-II, and TPCA-III, using lower-order t-scalars. For example, when 
$r = 35$, TPCA-3 outperforms TPCA-III by $1.36$ dB, i.e., $1.36$ dB = $20.23$ dB - $18.87$ dB.

\textbf{Average Pooling.}\;
To adopt the generalized outputs in t-scalars, t-vectors, or t-matrices to canonical algorithms, 
one needs a mechanism to down-size generalized outputs over $C$ to canonical results over complex numbers. 
Average pooling is such a down-sizing mechanism, introduced as follows.

Given a t-matrix  $X_\mathit{TM} \in C^{M_1\times M_2} \equiv \mathbb{C}^{I_1\times \cdots \times I_N \times 
M_1\times M_2}$, one can use average pooling to down-size all t-scalar entries of $X_\mathit{TM}$ to have a 
matrix $X_\mathit{mat} \in \mathbb{C}^{M_1\times M_2}$ 
given by 
\begin{equation}
(X_\mathit{mat})_{m_1,\,m_2} = (1/K) \cdot 
\scalebox{1.1}{$
\sum\nolimits_{(i_1,\cdots,i_N) \in [I_1]\times \cdots \times [I_N]}
$}
(X_{\mathit{T},\,m_1,\,m_2})_{i_1,\cdots,i_N} \in \mathbb{C}
\end{equation}
where 
$K \doteq \myK $, 
$X_{{T},\,m_1,\,m_2} \doteq (X_\mathit{TM})_{m_1, m_2} \in C$ denotes 
the $(m_1, m_2)$-th t-scalar entry of $X_\mathit{TM}$, 
$(X_\mathit{mat})_{m_1,\,m_2} \in \mathbb{C}  $ denotes the 
$(m_1,\,m_2)$-th complex entry of the matrix $X_\mathit{mat}$, 
for all $m_1 \in [M_1]$ and $m_2 \in [M_2]$.

Using the average pooling in the experiment, one reduces all TPCA variants over 
t-scalars to their canonical counterparts over complex numbers.

As a consequence, each t-scalar is reduced to a vector
with $N^\mathit{scalar} = 3072$. When computing PSNRs in equation 
(\ref{equation:PSNR}), the parameter $N^\mathit{entry}$ is given by 
$N^\mathit{entry} = 25\cdot N^\mathit{scalar} = 76800$.

The second row of Figure \ref{figure:images-from-CIFAR-10-images-with-higher-order-tscalar} shows the PSNR curves of all the TPCA variants with average pooling. 
These curves corroborate the 
observation found from the subfigures in the first row of 
Figure \ref{figure:images-from-CIFAR-10-images-with-higher-order-tscalar}.  
Furthermore, it shows from the two rows of Figure 
\ref{figure:images-from-CIFAR-10-images-with-higher-order-tscalar}
 that a PSNR curve with average pooling is even higher 
than the associated PSNR curve without average pooling.

To give a panoramic comparison of different algorithms with different settings, 
Figure \ref{figure:images-from-CIFAR-10-images-with-higher-order-tscalar} gives a 2D heat-map of 
PSNRs in the last row, where TPCA-3, using order-seven t-scalars, has the highest PSNRs 
and PCA, using order-zero t-scalars, has the lowest PSNRs.

\section{Conclusion}
\label{section:conclusions}

A semisimple paradigm of tensorial matrices over an algebra of generalized scalars is proposed for general data analytics with visual information analysis applications. The algebraic paradigm generalizes and is backward-compatible with the canonical paradigm, combining the higher-order merits of multi-way arrays and the low-order intuition of canonical complex matrices.

In the algebraic paradigm, scalars are extended to the so-called t-scalars, which are implemented as multi-way complex arrays of a fixed-size. Under the bestowed algebraic operations, the set of t-scalars form a semisimple associative algebra, which is unital, commutative, and a novel *-algebra. Due to its semisimplicity, the semisimple algebra can be decomposed to a finite number of irreducible algebras, isomorphic to the field of complex numbers, which is also a simple algebra.

With the backward-compatible simple paradigm, many canonical algorithms and applications over complex numbers can be straightforwardly extended over the new algebra as long as the scalar entries of each t-scalars are correlated. To this end, we propose a neighborhood strategy to extend legacy visual information data to the higher-order versions. In theory, the computational cost of a higher-order generalization is a linear function of the size of a t-scalar, i.e., the number of entries of a t-scalar. To verify the semisimple paradigm's effectiveness and its backward-compatibility, we choose to generate several classical algorithms and applications to their higher-order versions and apply them to analyze legacy images. Our experiments on these publicly available images show the semisimple paradigm, generalized algorithms, and applications compare favorably with their canonical counterparts. Our experiments show that higher-order generalizations also outperform their low-order counterparts.

\section*{Acknowledgments}
Liang Liao would like to thank associate professor Xiuwei Zhang of the Northwestern Polytechnical University, professor Pingjun Wei, and Xinqiang Wang of the Zhongyuan University of Technology for supporting this work. The students or formal students, including Yuemei Ren, a former Ph.D. student at the Northwestern Polytechnical University, Xuechun Zhang, Shen Lin, Chengkai Yang, Haichang Ye, and Jie Yang,
all postgraduate students supervised by Liang Liao at the Zhongyuan University of Technology,
help a lot for the experiments.

This research of Liang Liao, by the suggestion of Professor Stephen John Maybank (Fellow of the IEEE, Member of Academia Europaea), was first launched in 2015, when Liang Liao served as a researcher at the Department of Computer Science, Birkbeck, University of London. Liang Liao is invited back to Birbeck and collaborates with Professor Stephen John Maybank to continue this research at the University of London. Liang Liao thanks Professor Yudong Zhang of the University of Leicester, Dr. Changzheng Ma, Chief engineer of MooVita Ltd., Singapore, and Hailong Zhu of the Guangdong University of Technology for helpful online discussions when Liang Liao was in London. The Henan Provincial Department of Science and Technology partially supported Liao's visit to London financially.

Liang Liao also would like to thank the Birkbeck Institute of Data Analytics for free using the 
high-performance computing facilities, although London's lockdown somehow affects the progress of this research.

This work was partially supported by the National Natural Science Foundation of China 
under the grant number No. U1404607 and the High-end Foreign Experts Program,  
under the grant number No. GDW20186300351, of the State Administration of Foreign Experts Affairs.

\section*{Credits}

Liang Liao and Stephen John Maybank contribute equally to the theory of t-scalars, t-matrices,  
the t-algebra, and the semisimple paradigm.

\section*{Open Source}
During this research, Liang Liao has completed a workable MATLAB library for 
numerical experiments and application prototyping. The library is pedagogy oriented with structure programming and follows the naming protocol of many MATLAB built-in functions. After the completion of this article, we are considering to open-source the version of the MATLAB library. We hope it can help the interested readers grasp the semisimple paradigm's philology and understand the mechanism and the backward-compatibility of the t-matrix theory. 

{\color{black}The code repository is at https://github.com/liaoliang2020/talgebra.}


An enhanced object-oriented version of this library and a version in a second programming language is also 
planned. However, at present, this research is only frugally founded, and the necessity of developing a new version significantly depends on the response for the current pedagogical version of this MATLAB library. If interested, please let us know and be free to contribute spiritually or financially to this project.

\section*{Contact Information}
All prospective fundings, supports, collaboration interests, job offers, suggestions, critics, 
and other responses are welcome. Be free to contact via liaolangis@126.com or 
liaoliang2018@gmail.com.

\section*{Appendix: Generalization of Supervised Classification and Neural 
Networks}

\subsection{Generalized Supervised Classification}
Besides the applications mentioned above, many other applications can be improved using the t-algebra paradigm as long as the scalar entries of each t-scalar are correlated, making sense of a linear transform, not necessarily the Fourier transform, of t-scalars.

The solution via a fixed-sized small neighborhood of each 
scalar, e.g., the solution in Section \ref{section:t-matrix-representation},  is one 
of the most convenient approaches 
to establish a correlation between the scalar entries of a t-scalar for spatially 
constrained data, including but not limited to images, videos, audios, and sequential 
data such as time series.

Using the solution in Section \ref{section:t-matrix-representation}, the generalization of canonical samples to their higher-order versions yields generalized inputs to generalized classifiers for supervised classification of legacy images.

\begin{figure}
\begin{minipage}[t] {1\textwidth}
\centering
\includegraphics[width=0.9\textwidth]{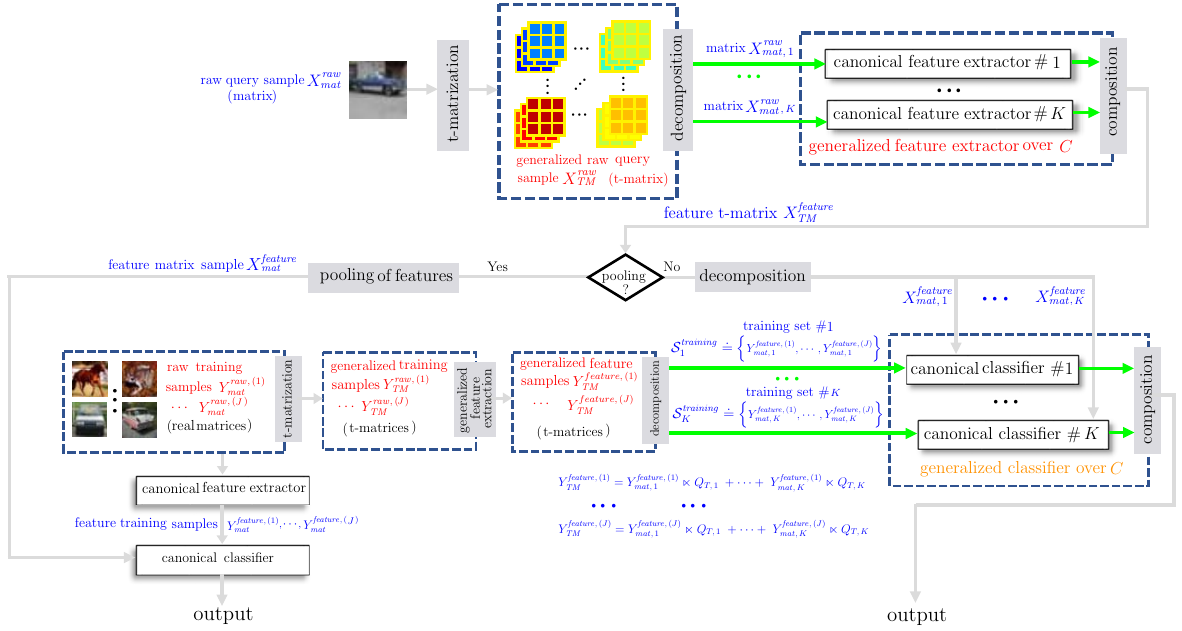}
\caption{Generalized classification over $C$ with or without pooling of features.}
\label{figure:generalized-classication-with-withhout-feature-pooling}
\end{minipage}
\begin{minipage}[t] {0.9\textwidth}
\centering
\includegraphics[width=1\textwidth]{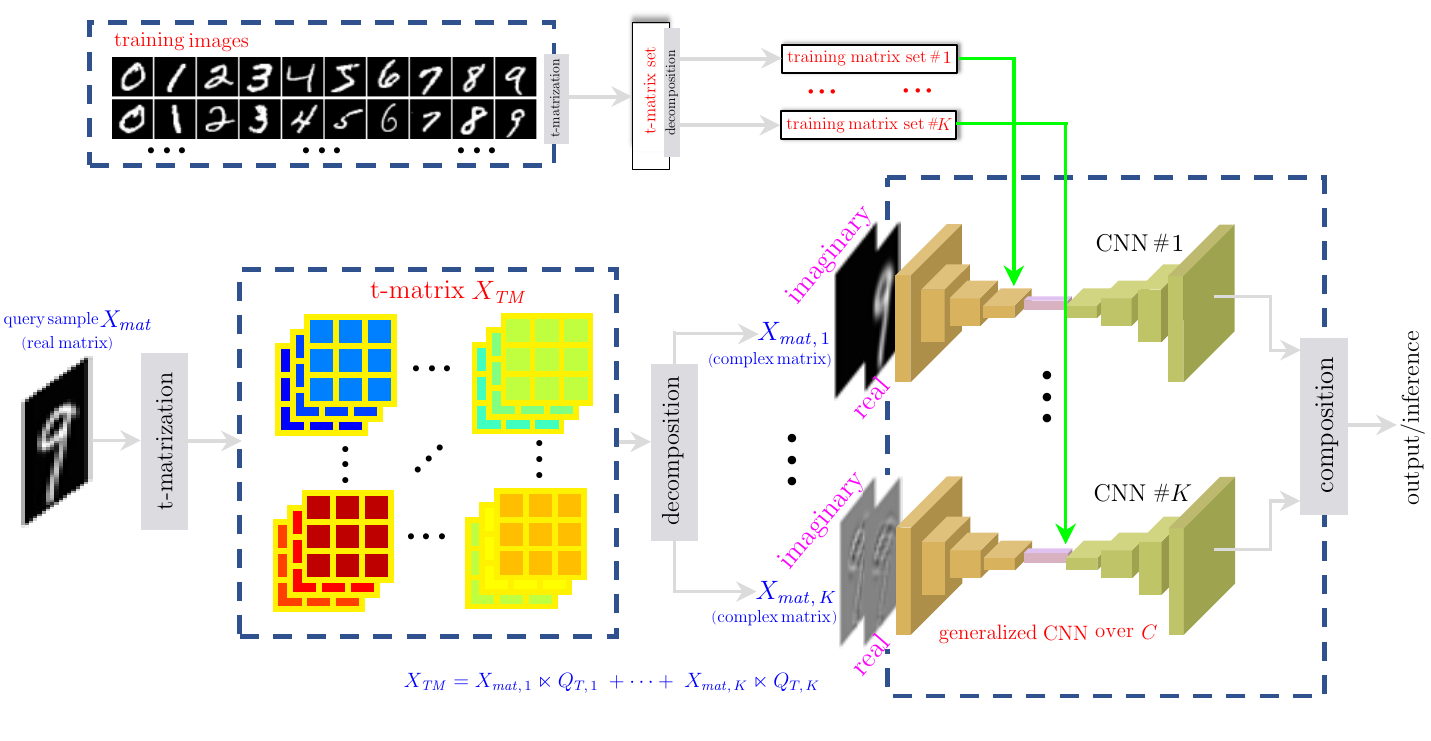}
\caption{A diagram of generalized CNN (Convolutional Neural Network) model over $C$.}
\label{figure:generalized-classifier}
\end{minipage}
\end{figure}

Figure \ref{figure:generalized-classication-with-withhout-feature-pooling} 
summarizes the generalized classification, over the t-algebra $C$, of a canonical 
matrix sample $X_\mathit{mat}^\mathit{raw}$. 
After 
t-matricizing the canonical sample 
$X_\mathit{mat}^\mathit{raw}$, using the solution in Section 
\ref{section:t-matrix-representation},  
to its higher-order version $X_\mathit{TM}^\mathit{raw}$, the 
t-matrix 
$X_\mathit{TM}^\mathit{raw}$ 
is sent to a generalized feature 
extractor, which is represented by $K$ canonical 
sub-extractors, over complex numbers, where $K \doteq \myK$.

The generalized extractor's output is a t-matrix ${X}_\mathit{TM}^\mathit{feature}$, either 
sent to a canonical classifier with pooling or sent without pooling to a generalized classifier 
represented by $K$ canonical sub-classifiers.

With pooling, the t-matrix ${X}_\mathit{TM}^\mathit{feature}$ is transformed into a canonical feature 
matrix ${X}_\mathit{mat}^\mathit{feature}$. 
The query matrix ${X}_\mathit{mat}^\mathit{feature}$ and the training matrices 
$
Y_\mathit{mat}^{\mathit{feature},\,(1)},\cdots,
Y_\mathit{mat}^{\mathit{feature},\,(J)}
$, as the inputs to a canonical 
classifier, yield a class label, 
i.e., an output in Figure \ref{figure:generalized-classication-with-withhout-feature-pooling} or \ref{figure:generalized-classifier},  for ${X}_\mathit{mat}^\mathit{feature}$.

If the t-matrix ${X}_\mathit{TM}^\mathit{feature}$ is sent to a generalized classifier without 
pooling, a brief interpretation is needed. To be a little concrete, let us take the following  
generalization of the classifier Nearest Neighbor, called TNN (Tensorial Nearest Neighbor), for example.

\subsection*{TNN: Generalized Nearest Neighbor}

The t-matrix ${X}_\mathit{TM}^\mathit{feature} $ 
sent to a generalized classifier 
is representable by $K$ complex matrices 
${X}_{\mathit{mat},\,1}^\mathit{feature},\cdots,{X}_{\mathit{mat},\,K}^\mathit{feature}$ such that 
\begin{equation}
{X}_\mathit{TM}^\mathit{feature} = \scalebox{1}{$\sum\nolimits_{k = 1}^{K}$}  
{X}_{\mathit{mat},\,k}^\mathit{feature} \ltimes \myQ{k} \;\;. 
\label{equation:XTM-decomposed}
\end{equation}

Let the raw training matrices be $Y_\mathit{mat}^{\mathit{raw},\,{(1)}},$ 
$\cdots,$
$Y_\mathit{mat}^{\mathit{raw},\,{(J)}}
$, and let their t-matrix versions be   
$Y_\mathit{TM}^{\mathit{raw},\,{(1)}},$
$\cdots,$ 
$Y_\mathit{TM}^{\mathit{raw},\,{(J)}}
$.  The $j$-th t-matrice $Y_\mathit{TM}^{\mathit{raw},\,(j)}$ 
is sent to a generalized feature extractor, yielding 
an output t-matrix 
$Y_\mathit{TM}^{\mathit{feature},\,{(j)}}$ which is represented by its 
$K$ constituent matrices $Y_{\mathit{mat},\,1}^{\mathit{feature},\,{(j)}}$   
$,\cdots,$
$Y_{\mathit{mat},\,K}^{\mathit{feature},\,{(j)}}   $
as follows 
\begin{equation} 
Y_\mathit{TM}^{\mathit{feature},\,{(j)}} 
\doteq \scalebox{1}{$
\sum\nolimits_{k = 1}^{K} Y_{\mathit{mat},\,k}^{\mathit{{feature}},\,{(j)} }
$}  \ltimes \myQ{k}   \;,\;\forall j \in [J]   \;\;.
\label{equation:training-t-matrix-decomposed}
\end{equation}

The matrices $Y_{\mathit{mat},\,k}^{\mathit{feature},\,{(j)}}$, 
$\forall\, (k,j) \in [K] \times [J]$,  
computed as in equation (\ref{equation:training-t-matrix-decomposed})
from $K$ training sets as follows 
\begin{equation}
\mathcal{S}_{k}^\mathit{training} \doteq
\Big\{
\scalebox{0.92}{$
Y_{\mathit{mat},\,k}^{\mathit{feature},\,(1)} ,\cdots,
Y_{\mathit{mat},\,k}^{\mathit{feature},\,(J)}
$}
\Big\}\,,\,\,\forall k \in [K] \;\;.
\label{equation:the-k-th-training-set}
\end{equation}

The $k$-th training set $\mathcal{S}_{k}^\mathit{training}$, given by equation 
(\ref{equation:the-k-th-training-set}), and the $k$-th feature matrix 
$X_{\mathit{mat},\,k}^\mathit{feature}$ 
in equation (\ref{equation:XTM-decomposed}), as the inputs to  
the $k$-th constituent classifier of a generalized classifier, yields 
the $k$-th constituent output
for all $k \in [K]$. 
The ensemble of these constituent outputs gives the final output of the 
generalized classifier TNN.

Given $J$ training t-matrices 
$Y_\mathit{TM}^{\mathit{feature},\, (1)}
,\cdots,
Y_\mathit{TM}^{\mathit{feature},\, (J)}
$ and a query t-matrix $X_\mathit{TM}^{\mathit{feature}}$, the generalized classifier TNN, in the form of $K$ constituent canonical classifiers, yields the generalized distances
$\mathit{d}(\scalebox{0.8}{$X_\mathit{TM}^\mathit{feature}, Y_\mathit{TM}^{\mathit{feature},\,(1)}$})$
$\cdots,$
$\mathit{d}(\scalebox{0.8}{$X_\mathit{TM}^\mathit{feature}, Y_\mathit{TM}^{\mathit{feature},\,(J)}$})
\geqslant Z_{T}$ as follows 
\begin{equation}
\mathit{d}(\scalebox{0.8}{$X_\mathit{TM}^\mathit{feature}, Y_\mathit{TM}^{\mathit{feature},\, (j)}$} ) \doteq 
\scalebox{1}{$\sum\nolimits_{k=1}^{K}$} 
\mathit{d}(
\scalebox{0.85}{$X_{\mathit{mat},\,k}^{\mathit{feature}}$}, 
\scalebox{0.85}{$Y_{\mathit{mat},\,k}^{\mathit{feature},\, (j)}$} ) \cdot \myQ{k} 
\in S^\mathit{nonneg}
\;\;,\;\; \forall j \in [J]\;.
\label{equation:generalized-distances}
\end{equation}
where $
\mathit{d}(
\scalebox{0.85}{$X_{\mathit{mat},\,k}^\mathit{feature} $}, 
\scalebox{0.85}{$Y_{\mathit{mat},\,k}^{\mathit{feature},\,(j)} $} ) \geqslant 0$, 
a nonnegative real number,  
denotes a canonical distance between the matrices   
$X_{\mathit{mat},\,k}^\mathit{feature}$ and 
$Y_{\mathit{mat},\,k}^{\mathit{feature},\,(j)} $, and the matrices 
$X_{\mathit{mat},\,k}^\mathit{feature}$ and 
$Y_{\mathit{mat},\,k}^{\mathit{feature},\,(j)} $
are given by
equations (\ref{equation:XTM-decomposed}) and (\ref{equation:training-t-matrix-decomposed}).

For example, the canonical distances $\mathit{d}(
\scalebox{0.85}{$X_{\mathit{mat},\,k}^\mathit{feature} $}, 
\scalebox{0.85}{$Y_{\mathit{mat},\,k}^{\mathit{feature}\,(j)}$} ), \forall (k,j) \in [K]\times [J] $,
in equation (\ref{equation:generalized-distances})
can be given by the following 
Frobenius norm as follows 
\begin{equation}
\mathit{d}(
\scalebox{0.85}{$X_{\mathit{mat},\,k}^\mathit{feature} $}, 
\scalebox{0.85}{$Y_{\mathit{mat},\,k}^{\mathit{feature}\,(j)}$} ) \doteq 
\left\|
\scalebox{0.85}{$X_{\mathit{mat},\,k}^\mathit{feature}$} - 
\scalebox{0.85}{$Y_{\mathit{mat},\,k}^{\mathit{feature},\,(j)}$}  
\right\|_F \geqslant 0, \, 
\scalebox{0.92}{$
 \forall (k, j) \in [K]\times [J] \;\;.
$}
\end{equation}

Let $
\mathcal{P} \doteq 
\left\{
\mathit{d}(\scalebox{0.85}{$X_\mathit{TM}^\mathit{feature}, Y_\mathit{TM}^{\mathit{feature},
\,(1)}$} ),\cdots, 
\mathit{d}(\scalebox{0.85}{$X_\mathit{TM}^\mathit{feature}, Y_\mathit{TM}^{\mathit{feature},
\,(J)}$} )
\right\}
$ be the poset 
formed by the generalized distances given by equation (\ref{equation:generalized-distances}). 
If $\inf \mathcal{P} \in \mathcal{P} $, in other words, 
if there exists the least element 
in the poset $\mathcal{P}$, let the least element be 
$\mathit{d}(\scalebox{0.8}{$X_\mathit{TM}^\mathit{feature}, Y_\mathit{TM}^{\mathit{feature},\, (j^\mathit{*})}$} ) \doteq 
\inf \mathcal{P} = \min \mathcal{P} $
where $j^{*} \in [J]$. Then, the label of the raw query matrix $X_\mathit{mat}$ is identified with the label
of t-matrix $\scalebox{0.9}{$Y_\mathit{TM}^{\mathit{feature},\,(j^{*})}$} $, more precisely, 
\begin{equation} 
\operatorname{class}(X_\mathit{mat}) \equiv 
\operatorname{class}(X_\mathit{TM}^\mathit{feature} )
\doteq \operatorname{class}(
\scalebox{0.85}{$
Y_\mathit{TM}^{\mathit{feature},\, (j^{*})}
$}  
)  \;\;.
\end{equation}

If the poset $
\mathcal{P} 
$ has no least element, 
without loss of generality, let the  
t-matrices 
$\scalebox{0.85}{$
Y_\mathit{TM}^{\mathit{feature},\,(1)},\cdots,Y_\mathit{TM}^{\mathit{feature},\,(J_\mathit{min})}
$}$ be the training samples, each having a minimum generalized distance, 
not the least generalized distance, to the query sample $X_\mathit{TM}^\mathit{feature}$.

Then, 
the label of the raw query matrix $X_\mathit{mat}$, or equivalently, the label 
of t-matrix $X_\mathit{TM}^\mathit{feature}$, can be identified with the label of any t-matrix among 
$\scalebox{0.85}{$Y_\mathit{TM}^{\mathit{feature},\,(1)},\cdots, Y_\mathit{TM}^{  
\mathit{feature},\,
(J_\mathit{min})}$}$. More precisely, the 
following identity makes sense when 
$\inf \mathcal{P} \notin \mathcal{P} 
$.
\begin{equation}
\operatorname{class}(X_\mathit{mat}) \equiv 
\operatorname{class}(\scalebox{0.85}{$X_\mathit{TM}^\mathit{feature}$} )
\doteq \operatorname{class}(
\scalebox{0.85}{$
Y_\mathit{TM}^{\mathit{feature},\,(j)}
$}  
) \,\;\text{for any}\,\; j \in \{1,\cdots,J_\mathit{min}\}  \;\;.
\end{equation}

In this scenario, one can never claim that the classification accuracy is $100\%$. 
On the other hand, if a generalized extractor and a generalized classifier are appropriately tuned, we contend, 
the generalized classifier should yield more favorable results than its canonical counterpart.

In theory, a generalized version's computational cost is only $K$ times that of the canonical 
counterpart. In Section 6.4 of \cite{liao2020generalized}, a comparison of the run time of the experiments of some t-matrix manipulations corroborates the above 
claims on computational cost. Interested readers are referred to as the results therein for more details.

Some well-known algorithms on supervised image classification/segmentation are generalized in our early work \cite{liao2020generalized,Liao2017Hyperspectral} and achieve favorable results compared with their canonical counterparts. Interested readers are referred to the reported experiments for more details in \cite{liao2020generalized,Liao2017Hyperspectral}.

\subsection*{TCNN: Generalized Convolutional Neural Networks}

Besides the generalized algorithms and classifiers in \cite{liao2020generalized,Liao2017Hyperspectral}, it is possible to generalize the popular 
Convolutional Neural Network (CNN). The t-matrix paradigm in Figure \ref{figure:generalized-classication-with-withhout-feature-pooling} applies to generalize the 
canonical CNN model for supervised visual-pattern classification.

Figure \ref{figure:generalized-classication-with-withhout-feature-pooling} shows the diagram of 
a generalized CNN model over the t-algebra $C$. A generalized CNN over the 
t-algebra is represented by $K$ (where $K \doteq \myK$) canonical CNNs trained by a set of 
labeled complex matrices decomposed from a generalized training set, e.g., a t-matrix set. If 
appropriately managed, we 
contend, the generalized CNN model should yield more favorable results than its canonical 
counterpart. We leave the verifications of this claim and the implementation of a generalized CNN 
for interested readers.

\bibliographystyle{acm}
\bibliography{egbib}

\end{document}